%% file: colm2025_conference.tex
\documentclass{article} 
\usepackage[final]{colm2025_conference}

\input{math_commands}
\usepackage{wrapfig}
\usepackage{microtype}
\usepackage{hyperref}
\usepackage{url}
\usepackage{booktabs}
\usepackage{graphicx}
\usepackage{lineno}
\usepackage{todonotes}
\usepackage{latexsym}
\usepackage{xcolor}
\usepackage{colortbl}
\definecolor{Gray}{rgb}{0.5,0.5,0.5}
\usepackage{color-edits}
\usepackage{array}
\usepackage{amssymb}
\usepackage{pifont}
\usepackage{xspace}
\usepackage{multirow}
\usepackage{makecell}
\usepackage{enumitem}
\usepackage{CJKutf8}

\definecolor{darkblue}{rgb}{0, 0, 0.5}
\hypersetup{colorlinks=true, citecolor=darkblue, linkcolor=darkblue, urlcolor=darkblue}
\definecolor{thinking}{rgb}{0.68, 0.87, 0.56}
\definecolor{nonthinking}{rgb}{0.96, 0.91, 0.67}
\definecolor{knowledge}{rgb}{0.78, 0.66, 0.74}
\newcommand{\hlthinking}[1]{\colorbox{thinking!100}{#1}}
\newcommand{\hlnonthinking}[1]{\colorbox{nonthinking!100}{#1}}
\newcommand{\hlknowledge}[1]{\colorbox{knowledge!100}{#1}}

\title{\benchmark: Decoupling Multimodal Reasoning Evaluation from Domain Knowledge}

\author{Yueqi Song\thanks{Equal Contributions.},\; Tianyue Ou\footnotemark[1],\; Yibo Kong\thanks{Equal Contributions.},\; Zecheng Li\footnotemark[2],\; Graham Neubig, Xiang Yue \\
\texttt{\{yueqis,tianyueo,gneubig,xyue2\}@cs.cmu.edu}
\\[1em]
\makebox[\textwidth]{\fontsize{11}{11}\selectfont Carnegie Mellon University}
}

\usepackage{color-edits}

\addauthor{gn}{magenta}

\newcommand{\benchmark}{\textsc{VisualPuzzles}\xspace}
\usepackage{caption}
\usepackage{arydshln} 
\usepackage{graphicx}
\usepackage[most]{tcolorbox}
\usepackage{fvextra}

\newtcolorbox[list inside=promptbox, auto counter, number within=section]{promptbox}[2][Prompt]{
    colback=black!5!white,
    colframe=#2,
    title=#1,
    fontupper=\small,
    breakable,
    enhanced,
    left=0pt,
    right=0pt,
    top=0pt,
    bottom=0pt,
    arc=5pt,
    boxrule=1pt,
}

\newtcolorbox[list inside=prompt,auto counter,number within=section]{prompt}[1][]{
    coltitle=white,
    fontupper=\footnotesize,
    boxsep=5pt,
    breakable,
    enhanced,
    left=0pt,
    right=0pt,
    top=0pt,
    bottom=0pt,
    boxrule=1pt,
    #1   
}

\begin{document}

\ifcolmsubmission
\linenumbers
\fi

\maketitle

\vspace{-1.0cm}
\begin{center}
    \url{https://neulab.github.io/VisualPuzzles/}

\end{center}

\begin{abstract}
Current multimodal benchmarks often conflate reasoning with domain-specific knowledge, making it difficult to isolate and evaluate general reasoning abilities in non-expert settings. 
To address this, we introduce \benchmark, a benchmark that targets visual reasoning while deliberately minimizing reliance on specialized knowledge. 
\benchmark consists of diverse questions spanning five categories: algorithmic, analogical, deductive, inductive, and spatial reasoning. One major source of our questions is manually translated logical reasoning questions from the Chinese Civil Service Examination. Experiments show that \benchmark requires significantly less intensive domain-specific knowledge and more complex reasoning compared to benchmarks like MMMU, enabling us to better evaluate genuine multimodal reasoning.
Evaluations show that state-of-the-art multimodal large language models consistently lag behind human performance on \benchmark, and that strong performance on knowledge-intensive benchmarks does not necessarily translate to success on reasoning-focused, knowledge-light tasks.  
Additionally, reasoning enhancements such as scaling up inference compute (with ``thinking'' modes) yield inconsistent gains across models and task types, and we observe no clear correlation between model size and performance. We also found that models exhibit different reasoning and answering patterns on \benchmark compared to benchmarks with heavier emphasis on knowledge. \benchmark offers a clearer lens through which to evaluate reasoning capabilities beyond factual recall and domain knowledge. 

\end{abstract}

\input{sections/1_Intro}
\input{sections/2_Data}
\input{sections/3_Experiments}
\input{sections/4_Analysis}
\input{sections/5_Related_work}

\input{sections/6_Conclusion}
\input{sections/7_Limitations}
\input{sections/8_Ethical_statement}
\input{sections/Acknowledgements}

\bibliography{colm2025_conference}
\bibliographystyle{colm2025_conference}

\newpage
\appendix  

\DoToC
\clearpage
\input{sections/Appendix}

\end{document}

%% file: math_commands.tex
\usepackage{amsmath,amsfonts,bm}

\def\eqref#1{equation~\ref{#1}}

\def\1{\bm{1}}

\DeclareMathAlphabet{\mathsfit}{\encodingdefault}{\sfdefault}{m}{sl}
\SetMathAlphabet{\mathsfit}{bold}{\encodingdefault}{\sfdefault}{bx}{n}

\usepackage{tabularx} 
\usepackage{adjustbox} 

\usepackage[normalem]{ulem}
\useunder{\uline}{\ul}{}
\usepackage{etoc}
\usepackage{titletoc}
\newcommand\DoToC{%
  \startcontents
  \printcontents{}{1}{\noindent \textbf{\Large{Table of Contents in Appendix}}\vskip3pt\vskip5pt}
  \vskip3pt\vskip5pt
}

%% file: sections/1_Intro.tex
\vspace{-10pt}
\begin{figure}[!h]
    \centering
    \includegraphics[width=\linewidth]{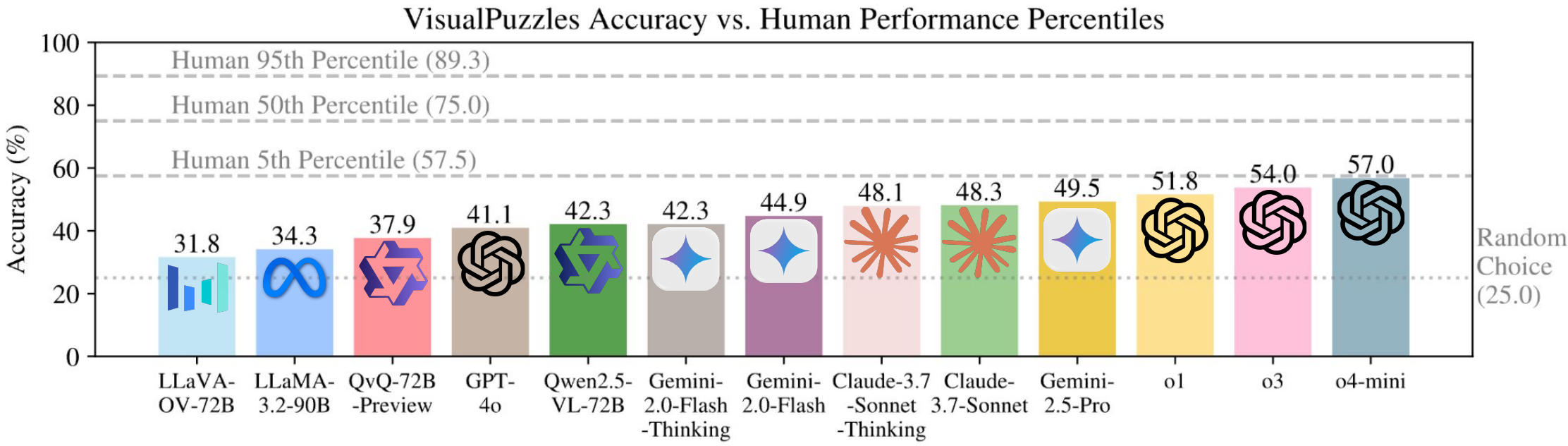}
    \vspace{-20pt}
      \caption{Model accuracy on \benchmark compared to human performance percentiles. All evaluated models fall below the human 5th percentile (57.5\%), highlighting the difficulty of \benchmark. Interestingly, models with explicit "thinking" modes do not consistently outperform their base versions, suggesting that current reasoning strategies do not yet generalize well to \benchmark's scenarios, even though these strategies have proven effective in existing reasoning tasks that often rely heavily on domain-specific knowledge.}
  \label{fig:percentiles}
  \vspace{-10pt}
\end{figure}

\section{Introduction}
\textit{Reasoning} is a cornerstone of both human and artificial intelligence, enabling systems to solve problems, draw inferences, and make decisions from information. Recent advances in multimodal large language models (MLLMs)~\citep{gpt4o,liu2023improvedllava,li2024llavaov,dubey2024llama,Qwen2.5-VL,yue2025pangea} exhibit early signs of reasoning in tackling complex tasks such as answering expert-level visual questions~\citep{yue2023mmmu,yue2024mmmu}, interpreting scientific diagrams~\citep{roberts2024scifibench}, and solving challenging math word problems~\citep{lu2023mathvista}.

Many of the tasks mentioned above are inherently \textit{knowledge-intensive}; large amounts of knowledge in domains such as science or math are necessary to answer questions correctly~\citep{yue2023mmmu}. However, in reality, reasoning does not necessitate knowledge. Even non-expert humans can successfully solve logic puzzles, spatial reasoning problems, and analogical tasks using general inferential skills, without requiring deep domain expertise.
This raises an important question: \textit{Can we measure MLLMs's reasoning ability independently of measuring their acquisition of domain-specific knowledge?}
This question is particularly important with the recent rapid development of reasoning models in the textual domain \citep{o1,deepseekai2025deepseekr1incentivizingreasoningcapability,qwq32b}, and emerging application to the visual domain \citep{qvq}.

\begin{figure*}[!t]
    \centering
    \includegraphics[width=\linewidth]{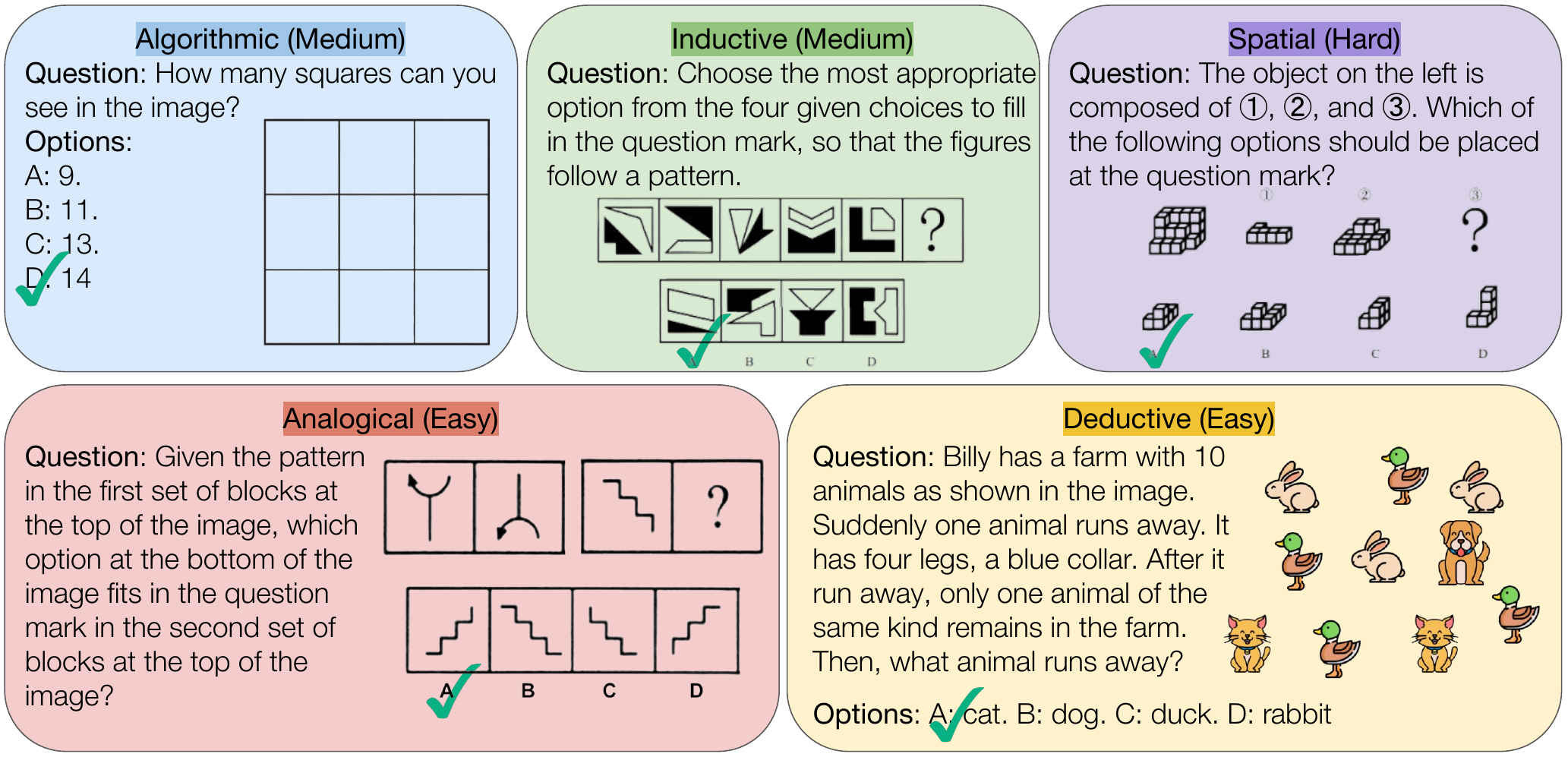}
    \caption{Example \benchmark instances within each reasoning category} 
    \label{fig:example}
    \vspace{-5mm}
\end{figure*}

To address this question, we introduce \benchmark, a multimodal benchmark explicitly crafted to assess reasoning capabilities independent of specialized knowledge.
\benchmark comprises 1,168 carefully curated puzzle-like questions that span five distinct categories of reasoning: algorithmic, analogical, deductive, inductive, and spatial, each annotated with varying difficulty levels.
\benchmark only requires basic common knowledge and the information presented in the question to solve problems, disentangling reasoning from domain-specific knowledge recall.
Our experiments show that \benchmark requires significantly fewer domain-specific knowledge concepts compared to benchmarks like MMMU, and models have sufficient knowledge required to solve \benchmark questions, enabling us to better assess multimodal reasoning versus pretrained factual knowledge.
While \benchmark minimizes reliance on domain expertise, its reasoning complexity exceeds that of existing benchmarks: in \benchmark,  82.1\% of models' solution steps are logical reasoning steps, compared to 71.5\% in MMMU. Additionally, no current MLLM surpasses even the 5th-percentile human performance, highlighting the benchmark’s difficulty and the limitations of today's models in general-purpose visual reasoning.


Our experiments with \benchmark reveal critical limitations in current MLLMs' multimodal reasoning ability by factoring out domain-specific knowledge requirements and only focusing on reasoning. Specifically, we uncover four key findings:


\begin{itemize}[leftmargin=*]
    \item \textbf{Strong performance on knowledge-heavy benchmarks does not transfer well.} Models that rank highly on MathVista and MMMU often experience substantial performance drops on \benchmark, highlighting a disconnect between knowledge-rich and knowledge-light multimodal reasoning tasks. 
    
    \item \textbf{Humans outperform models on easy and medium tasks, while both degrade on harder ones.} Human participants show strong and consistent performance on easy and medium-level questions across reasoning categories. In contrast, models struggle even on simpler tasks.
    
    \item \textbf{Reasoning enhancements (e.g., long CoT and “thinking” mode) yield inconsistent gains.} While explicit reasoning strategies help certain models tackle complex reasoning tasks, these techniques do not consistently improve performance across all model families and task types.
    
    \item \textbf{Scaling model size does not ensure stronger reasoning.} We observe no clear trend indicating that larger models outperform smaller ones on \benchmark, suggesting that scaling up parameters alone is insufficient to improve domain-agnostic multimodal reasoning.
\end{itemize}

%% file: sections/2_Data.tex
\section{\benchmark}
\subsection{Motivation and Design Principles of \benchmark}
Existing benchmarks often conflate multimodal reasoning with domain-specific knowledge, making it difficult to isolate and measure the pure reasoning capabilities of these models.

\benchmark is designed to explicitly address this issue by providing a testbed focused on evaluating multimodal reasoning in isolation from specialized knowledge. Specifically, \benchmark centers on puzzle-like questions that rely solely on the provided image, question text, and basic common-sense reasoning. The core design principle behind \benchmark is to limit the need for external or pretrained domain knowledge.  \autoref{fig:example} shows examples of \benchmark within each reasoning category.

\subsection{Data Collection and Curation}

We curated \benchmark using a multi-stage pipeline. The process involved sourcing, adapting, and validating questions with an emphasis on reasoning quality and minimal reliance on specialized knowledge.

\noindent\textbf{Question Sourcing}. We collected questions from three primary sources: (1) online resources and textbooks focused on logical, visual, and spatial puzzles, (2) synthesized items using images from large-scale vision datasets paired with text prompts, and (3) carefully repurposed items from existing multimodal reasoning benchmarks. Each source was selected to ensure a wide variety of reasoning challenges while avoiding trivial or fact-heavy questions. One major source of our questions is manually translated logical reasoning questions from the Chinese Civil Service Examination\footnote{\begin{CJK}{UTF8}{gbsn}  Chinese Civil Service Examination (Logic Test), 中国国家公务员考试行测（逻辑推理）
\end{CJK}}. Other sources are listed in \autoref{app:stats}.

\noindent\textbf{Format Adaptation}. All collected items were adapted into a consistent multiple-choice format with four options, balancing between text-based and image-based answer choices. This modality balance allows us to better test models' abilities to perform reasoning across diverse formats.

\noindent\textbf{Data Validation}. During curation, we applied strict filtering criteria to eliminate questions requiring advanced mathematical knowledge, specialized domain knowledge and facts. Questions were retained only if they could be solved using information present in the image, the question prompt, and basic common sense. A multi-round validation process was conducted by human annotators, focusing on question clarity, solvability, and reasoning type classification.

\noindent\textbf{Attribute Annotation}. Finally, each question was annotated with two key attributes:
\begin{itemize}[leftmargin=*]
    \item Reasoning Category: Each item was categorized as \emph{algorithmic}, \emph{analogical}, \emph{deductive}, \emph{inductive}, or \emph{spatial} reasoning. These five categories were selected as they represent fundamental forms of reasoning widely discussed in literature~\citep{liu2020logiqa,lu2023mathvista,yue2023mmmu,gao2023lora}. At the same time, we aimed to balance comprehensiveness with conciseness, avoiding an overly fine-grained taxonomy that could dilute the benchmark’s clarity and usability. This categorization ensures that \benchmark covers a broad yet manageable set of reasoning skills relevant to multimodal LLM evaluation.
    \begin{itemize}
        \item Algorithmic Reasoning involves  reasoning over algorithmic rules.
        \item Analogical Reasoning requires analyzing the relationships between a pair of entities. 
        \item Deductive Reasoning involves logically drawing conclusions from known premises.
        \item Inductive Reasoning focuses on generalizing rules from observed patterns. 
        \item Spatial Reasoning  requires interpreting and manipulating spatial relationships. 
    \end{itemize}
    \item Difficulty Level: Labeled as easy, medium, or hard, based on annotators’ estimated cognitive load and time-to-solve metrics.
\end{itemize}

This pipeline ensures that \benchmark presents a diverse set of high-quality questions designed to challenge multimodal LLMs on their reasoning abilities without involving pretrained domain knowledge.

\subsection{Dataset Statistics}

\begin{wraptable}{r}{6.5cm}
\vspace{-18mm}
\centering 
\resizebox{6.5cm}{!}{ 
\begin{tabular}{lc} 
\toprule
Category & Statistics \\
\midrule
Total Questions & 1168 \\
- Algorithmic Reasoning & 262 \\
- Analogical Reasoning & 211 \\ 
- Deductive Reasoning & 200 \\
- Inductive Reasoning & 209 \\
- Spatial Reasoning & 286 \\\midrule
Easy/Medium/Hard & 46\%/39\%/15\%\\
Option Type (Image/Text) & 57\%/43\% \\
AVG. Question Length & 154.9 \\
\% Easy Words & 54\% \\
\bottomrule 
\end{tabular} 
}
\vspace{-3mm}
\caption{Statistics of \benchmark}
\label{tab:stats} 
\vspace{-4mm}
\end{wraptable}
\benchmark comprises 1,168 multimodal reasoning puzzles. It is designed to provide a balanced distribution across different reasoning categories, difficulty levels, and option formats for comprehensive evaluation. The statistics of \benchmark are shown in \autoref{tab:stats}. 

Across the five reasoning types, we maintain a roughly even distribution, ensuring that no single reasoning style dominates the benchmark. Similarly, we balanced the dataset across the three difficulty levels (easy, medium, hard) to capture a wide spectrum of cognitive demands. Approximately half of the answer choices in the dataset are image-based and the other half are text-based, enabling evaluation of models' abilities to reason across diverse query formats.

In terms of language complexity, \benchmark was constructed with an emphasis on accessibility. Most of the question text uses Basic English vocabulary\footnote{\url{https://en.wiktionary.org/wiki/Appendix:Basic_English_word_list}} to minimize the impact of linguistic complexity on reasoning performance, focusing the evaluation strictly on multimodal reasoning. 

Compared to prior benchmarks, \benchmark is unique in that it explicitly minimizes domain-specific knowledge requirements while maintaining high reasoning complexity. We demonstrate these traits of \benchmark in Section \ref{sec:analysis}.

%% file: sections/3_Experiments.tex
\section{Experiments and Results}
\label{sec:experiments}

\subsection{Experimental Setup}

We comprehensively evaluated the reasoning abilities of a variety of MLLMs on \benchmark. Additionally, we performed human evaluations to better understand the gap between human and models' reasoning capabilities.

We selected a diverse set of proprietary and open MLLMs to ensure broad coverage in terms of model architecture, training scale, and intended application domains. This diversity allows us to capture a wide spectrum of current approaches and capabilities in the field. We integrated \benchmark into Lmms-eval \citep{lmms_eval2024}.

\noindent\textbf{Proprietary Models.}
We evaluate several leading proprietary models that represent the current state of the art:
(1) GPT-4o, o1, o3, and o4-mini \citep{gpt4o,o1};
(2) Gemini-1.5-Pro, Gemini-2.0-Flash, Gemini-2.0-Flash-Thinking, and Gemini-2.5-Pro \citep{deepmind_gemini_report};
(3) Claude-3.5-Sonnet and Claude-3.7-Sonnet \citep{Claude}.
Among these, o1, o3, o4-mini are explicitly optimized for reasoning, while Gemini-2.0-Flash-Thinking and Claude-3.7-Sonnet incorporate dedicated modules for extensive step-by-step problem-solving.

\noindent\textbf{Open Models.}
We further evaluate widely used open MLLMs to gauge how open models compare against proprietary models:
(1) LLaVA Series \citep{liu2023improvedllava,liu2024llavanext,li2024llavaov}: LLaVA-1.5 (7B/13B), LLaVA-1.6 (7B/13B/34B), and LLaVA-OV (0.5B/7B/72B);
(2) Llama-3.2-Vision-Instruct (11B/90B) \citep{dubey2024llama};
(3) Qwen-VL Series \citep{bai2024qwenvl,yang2024qwen2,Qwen2.5-VL,qvq}: including Qwen-VL, Qwen2-VL (2B/7B/72B-Instruct), Qwen2.5-VL (3B/7B/72B-Instruct), and QvQ-72B-Preview;
(4) Cambrian (8B/13B) \citep{tong2024cambrian};
(5) Pangea-7B \citep{yue2025pangea}.

We apply both direct multiple-choice prompting and Chain-of-Thought (CoT) prompting to each model, following recent findings that CoT can significantly enhance model reasoning on complex multimodal tasks.
For each model we report the best performance, whether achieved by direct multiple-choice prompting or CoT prompting. 

\noindent\textbf{Human Performance.}
To establish a strong baseline for comparison, we conducted human evaluations with 70 college-level volunteers. Human performance provides a valuable upper-bound reference for assessing the current capabilities and limitations of multimodal reasoning models. While this serves as a benchmark for present-day systems, it is possible that future models could surpass this level of performance. Each participant was randomly assigned a subset of the puzzles and completed them under the same resource-constrained conditions as the models (i.e., without access to external tools or the internet). On average, participants completed each puzzle in 78 seconds, reflecting the typical cognitive load and time demands imposed by \benchmark.

\subsection{Overall Results}

\begin{table*}[!htbp]
\centering
\resizebox{\linewidth}{!}{
\begin{tabular}{@{}lcccccc@{}}
\toprule
\textbf{Model} & 
\textbf{Algorithmic} & \textbf{Analogical} & \textbf{Deductive} & \textbf{Inductive} &
\textbf{Spatial} & \textbf{Overall}\\
\midrule
\color{Gray}\textbf{Random Choice} & \color{Gray}25.0 & \color{Gray}25.0 & \color{Gray}25.0 & \color{Gray}25.0 & \color{Gray}25.0 & \color{Gray}25.0 \\
\textbf{Human (95th Percentile)} & 100.0 & 100.0 & 100.0 & 81.6 & 100.0 & 89.3 \\
\textbf{Human (50th Percentile)} & 88.0 & 66.0 & 80.0 & 50.0 & 90.0 & 75.0 \\
\textbf{Human (5th Percentile)} & 68.1 & 25.0 & 37.0 & 0.0 & 59.1 & 57.5 \\
\midrule
\multicolumn{7}{c}{\textit{Proprietary Models}} \\
\midrule
\textbf{GPT-4o} & 49.2 & 58.3 & 49.0 & 27.3 & 26.2 & 41.3 \\
\rowcolor{thinking}
\textbf{o1} & 63.7 & 68.3 & 67.5 & 29.2 & 34.3 & 51.8 \\
\rowcolor{thinking}
\textbf{o3} & 64.5 & 68.3 & 69.5 & 27.3 & 42.7 & 54.0 \\
\rowcolor{thinking}
\textbf{o4-mini} & 65.3 & 68.7 & 75.5 & 33.0 & 45.5 & 57.0 \\
\noalign{\vspace{2pt}} 
\hdashline
\noalign{\vspace{2pt}} 
\textbf{Gemini-2.0-flash} & 55.3 & 58.8 & 57.0 & 24.4 & 31.8 & 45.0 \\
\rowcolor{thinking}
\textbf{Gemini-2.0-flash-thinking} & 46.6 & 70.1 & 49.0 & 24.9 & 25.5 & 42.2 \\
\noalign{\vspace{2pt}} 
\hdashline
\noalign{\vspace{2pt}}
\rowcolor{thinking}
\textbf{Gemini-2.5-pro} & 60.0 & 64.0 & 60.0 & 29.7 & 36.4 & 49.5 \\
\noalign{\vspace{2pt}} 
\hdashline
\noalign{\vspace{2pt}} 
\textbf{Claude-3.7-Sonnet} & 64.5 & 48.3  & 65.0 & 26.8 & 37.4 & 48.3\\
\rowcolor{thinking}
\textbf{Claude-3.7-Sonnet-Thinking} & 67.2 & 44.1 & 61.5 & 31.1 & 37.1 & 48.2 \\
\midrule
\multicolumn{7}{c}{\textit{Open Models (Qwen-Based)}} \\
\midrule
\textbf{LLaVA-OV-7B} & 27.5 & 28.0 & 40.5 & 24.4 & 28.0 & 29.4 \\
\textbf{Pangea-7B} & 32.4 & 23.7 & 38.5 & 28.7 & 32.5 & 31.3 \\
\textbf{Qwen2.5-VL-7B-Instruct} & 38.2 &23.7 &51.5 &24.9 &31.1 & 33.7 \\
\textbf{LLaVA-OV-72B} & 34.7 & 26.5 & 37.0 & 27.3 & 28.7 & 30.8 \\
\rowcolor{thinking}
\textbf{QvQ-72B-Preview} & 44.8 & 43.6 & 44.0 & 26.8 & 30.8 & 37.8 \\
\textbf{Qwen2.5-VL-72B-Instruct} & 53.4 &46.9 &58.0 &25.8 &29.5 &42.3 \\
\midrule
\multicolumn{7}{c}{\textit{Open Models (Llama-Based)}} \\
\midrule
\textbf{Cambrian-8B} & 31.3 & 24.2 & 36.0 & 24.0 & 29.0 & 28.9 \\
\textbf{Llama-3.2-11B-Vision-Instruct} & 31.0 & 30.8 & 39.0 & 21.1 & 26.2 & 29.4 \\
\textbf{Llama-3.2-90B-Vision-Instruct} & 45.0 & 23.2 & 43.0 & 26.3 & 31.5 & 34.1 \\
\bottomrule
\end{tabular}}
\caption{Performance (\%) comparison of humans and selected models on \benchmark. We report the best performance resulting from direct multiple-choice prompting and CoT prompting for each method. We highlighted all the \colorbox{thinking}{reasoning models}.}
\label{tab:model_performance}
\vspace{-5mm}
\end{table*}


\autoref{tab:model_performance} and \autoref{fig:percentiles} compare the performance of humans and a selected set of models.%
\footnote{
Full results for every model discussed in Section \ref{sec:experiments} are provided in \autoref{app:full}, including separate performance outcomes for both direct multiple-choice and CoT prompting.}
All evaluated models, even the proprietary ones, perform below the 4th percentile of human accuracy, underscoring the significant gap in multimodal reasoning abilities. These results reinforce our finding that, although models have made progress in multimodal understanding, there remains a substantial margin for improvement before they can match or surpass human performance on multimodal reasoning.

This pattern holds across categories as well. 
In \autoref{tab:model_performance}, top human participants (95th percentile) exhibit near-perfect accuracy on multiple reasoning categories, while model performance remains substantially lower, even lower than the worst human performance (5th percentile). These results emphasize the need for continued innovation in model architectures and training paradigms if we aim to close the gap between model and human intelligence on complex multimodal reasoning.


%% file: sections/4_Analysis.tex
\section{Disentangling Reasoning from Domain Knowledge}
\label{sec:knowledge}

\subsection{Knowledge Intensity of \benchmark}

\noindent\textbf{Is \benchmark less knowledge-intensive than existing reasoning benchmarks?}
This question is central to our goal of disentangling reasoning ability from domain-specific knowledge. Many current benchmarks blur this line, making it difficult to assess general reasoning in non-expert settings. \benchmark was designed to target visual reasoning skills while deliberately minimizing reliance on specialized knowledge.

To test whether \benchmark achieves this goal, we prompted GPT-4o to generate ``knowledge concept checklists'' for 50 randomly selected questions from a widely-used knowledge-intensive reasoning dataset MMMU and 50 from \benchmark. 
We manually verified each question as discussed in \autoref{app:knowledge_human}.
Each checklist comprises knowledge-specific questions intended to assess whether a model possesses the background information required to solve the original problem. 
For example, if a question depends on understanding two distinct physics laws, its checklist would include a question to explain each. The number of checklist items per instance serves as a proxy for knowledge intensity.

\begin{wraptable}{r}{6cm}
\centering 
\vspace{-5mm}
\resizebox{6cm}{!}{ \begin{tabular}{lc} 
\toprule
Benchmark & \# Knowledge Qs. \\
\midrule
MMMU & 3.9 \\
\benchmark & 1.1\\
\bottomrule 
\end{tabular} } 
\vspace{-2mm}
\caption{AVG. number of knowledge concept questions generated per instance on MMMU vs.\ \benchmark.} 
\vspace{-2mm}
\label{tab:knowledge} 
\end{wraptable}

We found that MMMU problems resulted in significantly more checklist items on average (3.9) compared to \benchmark (1.1), as shown in \autoref{tab:knowledge}. This supports the hypothesis that \benchmark is substantially less reliant on domain knowledge. As a result, performance on \benchmark more directly reflects a model’s ability to reason over visual and textual content, offering a clearer signal of progress in multimodal reasoning. Full prompt examples and further discussion are provided in \autoref{app:knowledge}.



\begin{figure*}[!h]
    \centering
        \includegraphics[width=\linewidth]{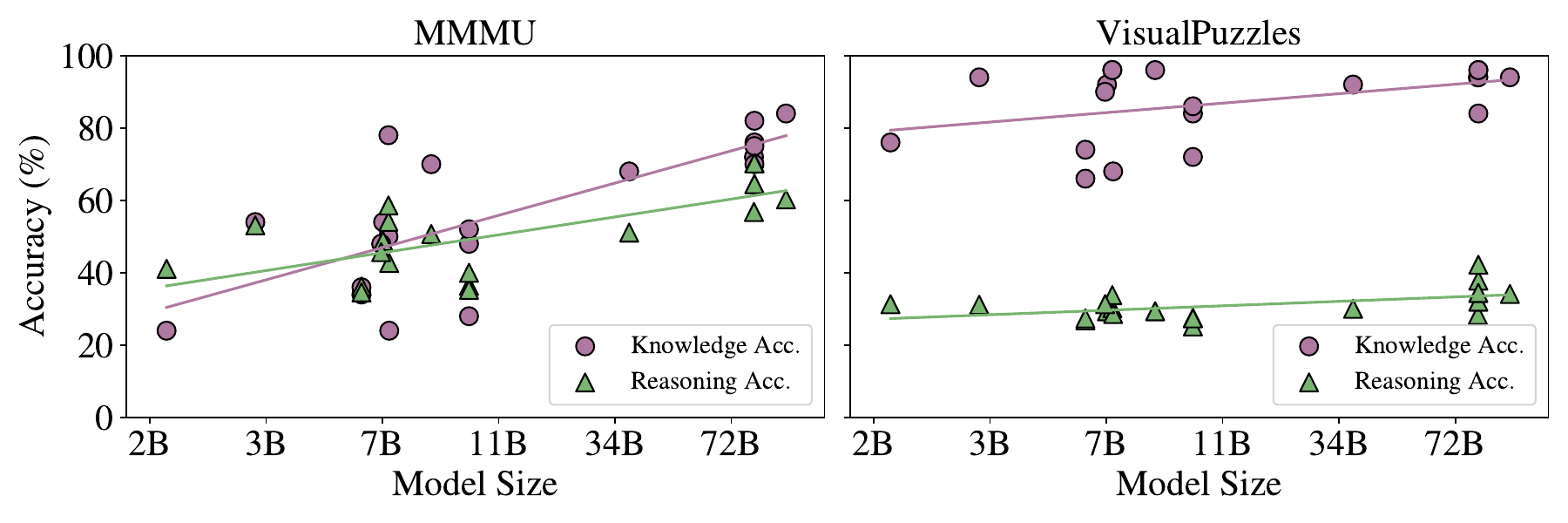}
    \includegraphics[width=\linewidth]{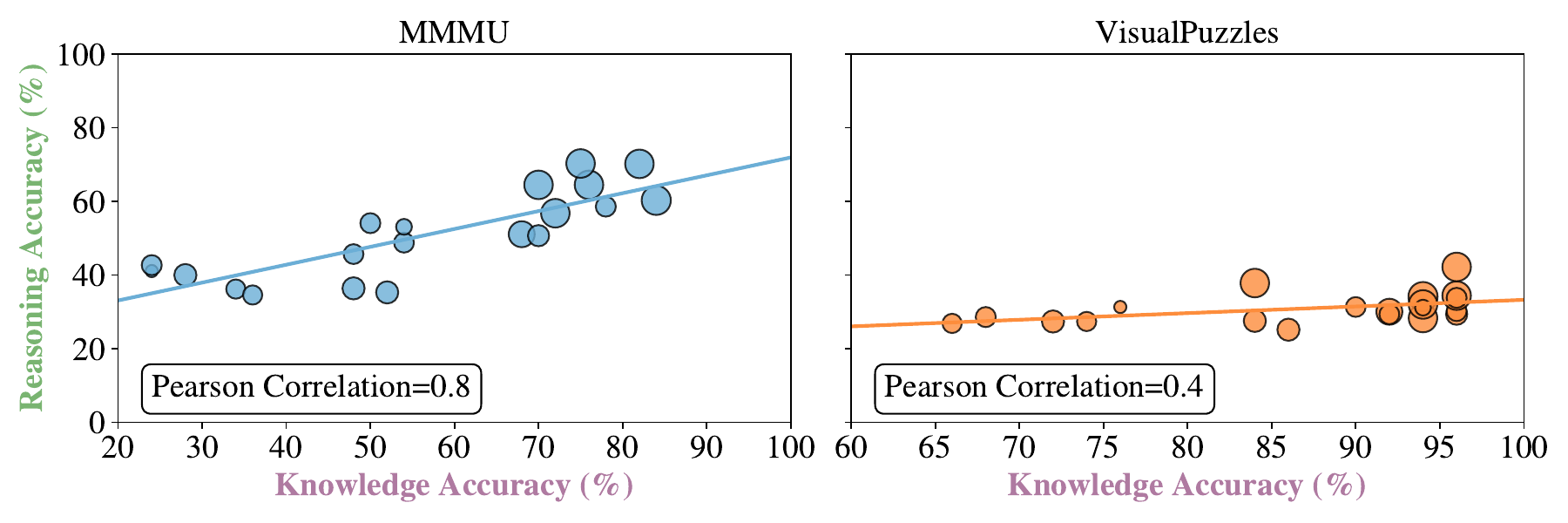}
    \vspace{-6mm}
    \caption{Scatter plots with trend lines of the relationship between accuracy and model size (top) and the relationship between reasoning and knowledge accuracy (bottom) on MMMU and \benchmark. The dots' sizes represent relative model sizes. The correlation between \hlthinking{reasoning accuracy} and \hlknowledge{knowledge accuracy} is higher on MMMU (0.8) than on \benchmark (0.4).} 
    \label{fig:benchmark}
\end{figure*}


\noindent\textbf{Do models already possess the knowledge required to solve \benchmark?}
To explore this, we measured models' knowledge accuracy—their ability to answer the knowledge checklist questions correctly—on both benchmarks. This metric reflects how much of the required knowledge is already known by the model, independent of reasoning. We found a stark contrast: while many models exceed 90\% knowledge accuracy on \benchmark, most score below 60\% on MMMU, with smaller models frequently dropping under 50\%. Only the largest models approach 80\% accuracy on MMMU, underscoring its heavier reliance on domain-specific knowledge. 

\noindent\textbf{Does scaling up model size improve performance?} We also plot reasoning accuracy (i.e., overall performance on the benchmark) in \autoref{fig:benchmark}, revealing some interesting trends: \begin{itemize}[leftmargin=*] 
    \item \textbf{MMMU}. Larger models tend to have higher knowledge accuracy, and this often translates into higher overall benchmark performance. This aligns with MMMU's reliance on domain-specific understanding; models with more parameters and training data are better at recalling relevant factual knowledge, thus improving their overall performance. 
    \item \textbf{\benchmark}. Although many models achieve near-100\% knowledge accuracy on \benchmark, we observe no clear increase in both knowledge and reasoning accuracy as model size grows. In contrast to MMMU, simply scaling number of parameters does not guarantee better performance on \benchmark, implying that further gains on \benchmark must stem from improvements in models' reasoning abilities rather than reliance on extensive knowledge. 
\end{itemize}

\noindent \textbf{What is the relationship between knowledge and reasoning?}
\autoref{fig:benchmark} shows two scatter plots with trend lines that measure how knowledge accuracy correlates with reasoning accuracy across different open models, where the relative sizes of the dots represent the sizes of the models. On MMMU (left), there is a strong positive correlation (0.8), suggesting that a model possessing more knowledge strongly correlates better reasoning performance. In contrast, \benchmark (right) exhibits a more modest correlation (0.4). Although there is still an upward trend, gains in knowledge accuracy lead to smaller improvements in reasoning accuracy. This discrepancy implies that while overcoming knowledge gaps is central to reasoning success on MMMU, \benchmark tasks demand more nuanced inference steps that depends less on domain knowledge.

Overall, these findings reinforce that \benchmark's comparatively lower knowledge requirements are readily met by both proprietary and open models. By contrast, MMMU poses a greater challenge to smaller models in terms of knowledge, for which scaling in size clearly benefits knowledge-intensive tasks. However, on \benchmark, larger model size alone is not a decisive factor, which might imply that genuine multimodal reasoning depends on more than just number of parameters or pre-trained knowledge.


\subsection{Reasoning Complexity of \benchmark}


\noindent\textbf{Do questions in \benchmark require more complex reasoning than those in existing benchmarks like MMMU?}

\begin{wraptable}{r}{6cm}
\centering 
\vspace{-5mm}
\resizebox{6cm}{!}{ \begin{tabular}{lcc} 
\toprule
Model & MMMU & \benchmark \\
\midrule
GPT-4o & 75.1\% & 87.0\% \\
Gemini-2.0-Flash & 67.9\% & 77.3\%\\
\bottomrule 
\end{tabular} } 
\vspace{-2mm}
\caption{Percentage of logical reasoning steps in solving benchmark questions.} 
\vspace{-2mm}
\label{tab:reasoning_complexity} 
\end{wraptable}

Besides observing that models generally achieve lower accuracy on \benchmark compared to MMMU, we further investigated whether this gap stems from increased reasoning complexity. To do so, we measured the proportion of reasoning steps required to solve each question. We began by gathering detailed, step-by-step solutions from the models for each question, which are manually verified for completeness. Then we classified if each step is a logical reasoning step with the help of LLM. We show the result in \autoref{tab:reasoning_complexity}. On average, logical reasoning steps take up 14.8\% more total steps in solving \benchmark questions compared to those of MMMU (82.1\% v.s. 71.5\%). This analysis is based on GPT-4o and Gemini-2.0-Flash across 200 randomly sampled questions per benchmark. These results suggest that \benchmark demand more extensive reasoning, aligning with its goal of evaluating deeper multimodal reasoning beyond factual recall. Prompt example is shown in \autoref{app:reasoning_complexity}.



\subsection{Do Reasoning Models Perform Better than Their Baselines?}

\label{sec:reasoning}

\begin{figure*}[!h]
    \centering
    \includegraphics[width=\linewidth]{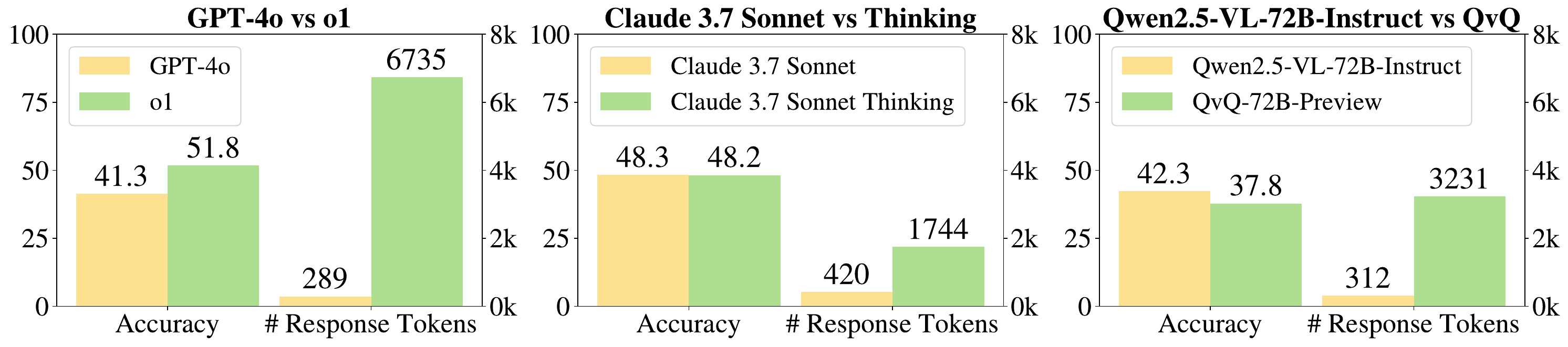}
    \vspace{-6mm}
    \caption{Comparison of accuracy and average number of total completion tokens of \hlthinking{reasoning models} and their \hlnonthinking{general counterparts} on \benchmark. We didn't include Gemini-2.0-Flash models here because Gemini-2.0-Flash-Thinking does not reveal the number of reasoning tokens of responses. The accuracies of Gemini-2.0-Flash and Gemini-2.0-Flash-Thinking is 45.0\% and 42.2\% respectively. Despite much higher number of completion tokens, \hlthinking{reasoning models} do not often achieve better performance on \benchmark.} 
    \label{fig:reasoning_model}
\end{figure*}

Recent reasoning models often scale up inference compute by generating longer chains of thought (CoTs) to enhance reasoning ability. To assess the effectiveness of this strategy on \benchmark, we compare several reasoning models with their non-reasoning counterparts in \autoref{fig:reasoning_model}. The reasoning model o1 outperforms GPT-4o overall. However, structured ``thinking'' modes, despite much higher number of completion tokens, show no consistent benefit. Similarity of output further reveals that the thinking mode primarily increases verbosity without meaningfully altering the underlying reasoning process, as illustrated in \autoref{fig:reasoning_case_study_1}.

\subsection{Are Branching and Revalidation Reasoning Patterns Effective on \benchmark?}


\begin{figure*}[!h]
    \centering
    \includegraphics[width=\linewidth]{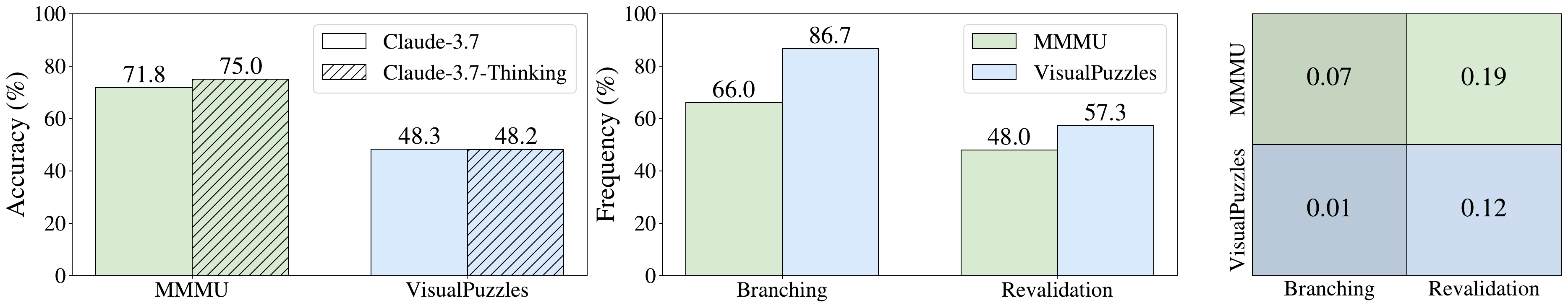}
    \vspace{-6mm}
    \caption{Comparison of Reasoning Pattern of Claude-3.7-Sonnet-Thinking on MMMU and \benchmark. Left figure compares the accuracy of Claude-3.7-Sonnet and Claude-3.7-Sonnet-Thinking on MMMU and \benchmark. Middle figure shows frequency of each pattern. Right figure shows correlation of the patterns with accuracy on the benchmarks.}
  \vspace{-2mm}
  \label{fig:reasoning_types}
\end{figure*}


As discussed in Section~\ref{sec:reasoning}, reasoning-enabled models do not consistently outperform their non-reasoning counterparts on \benchmark. 
To better understand this discrepancy, we examine Claude-3.7-Sonnet-Thinking's reasoning behaviors present in long CoTs, specifically, branching and re-validation, which are known to play important roles in enhancing reasoning performance\footnote{We examined Claude-3.7-Sonnet-Thinking as it explicitly provides thinking output.}.

As shown in \autoref{fig:reasoning_types}, our analysis reveals a striking contrast between benchmarks. On MMMU, both branching and re-validation correlate positively with model accuracy. These strategies help models explore alternative reasoning paths and revisit earlier steps, aiding in the retrieval of relevant factual knowledge,an essential component for solving MMMU's knowledge-intensive questions. An illustrative example is provided in \autoref{app:knowledge}.

\begin{wrapfigure}{r}{0.5\textwidth}
  \vspace{-8mm}
  \begin{center}
    \includegraphics[width=0.5\textwidth]{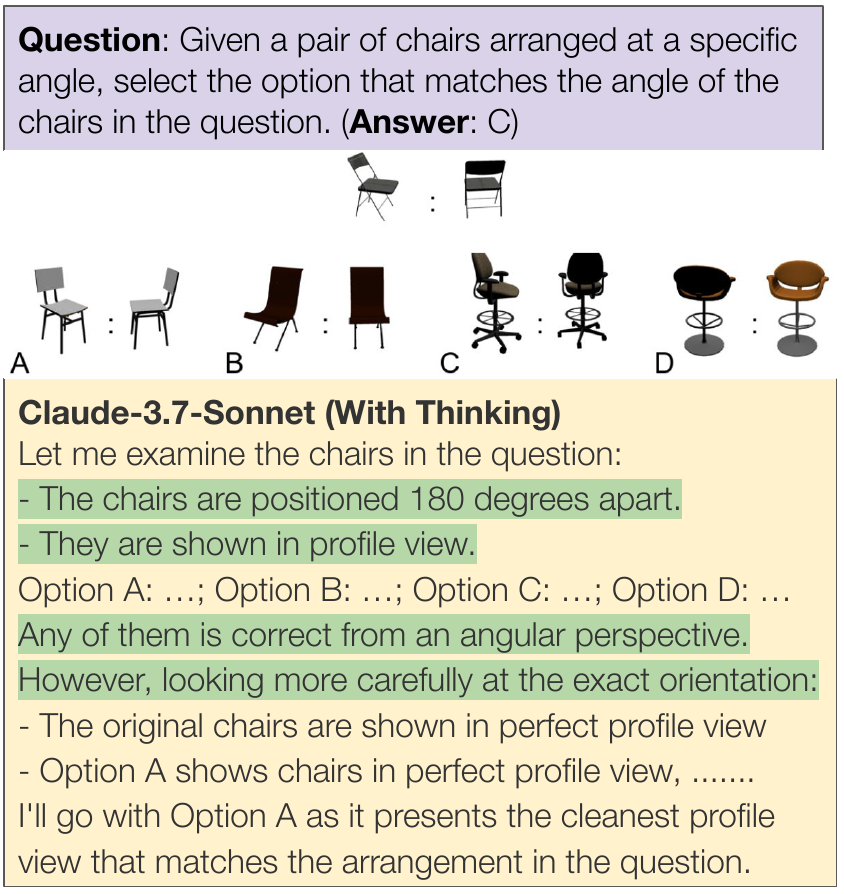}
  \end{center}
  \vspace{-4mm}
  \caption{An example of Claude-3.7-Sonnet-Thinking utilizing \hlthinking{branching} to solve a \benchmark puzzle.}
  \vspace{-18mm}
  \label{fig:reasoning_example}
\end{wrapfigure}
Surprisingly, on \benchmark, these reasoning behaviors are more frequent, yet less predictive of success. Despite their increased presence in long-form responses, we observe no significant correlation between these strategies and task accuracy. This suggests that models may be using branching and re-validation in ways that do not meaningfully contribute to solving the problem.

\autoref{fig:reasoning_example} highlights this with an example from Claude-3.7-Sonnet-Thinking, where the model applies branching on a \benchmark puzzle. However, the additional reasoning paths remain shallow and fail to engage with the core challenge—understanding the spatial arrangement of chairs in the image. The full response is included in \autoref{app:knowledge}.


\section{Analysis}
\label{sec:analysis}
\subsection{Do Models Approach \benchmark Questions Differently?}

\begin{wraptable}{r}{8cm}
\vspace{-5mm}
\centering 
\resizebox{8cm}{!}{ 
\begin{tabular}{lcc}
\toprule
Benchmark & Answer-First & Option-First \\
\midrule
MMMU &  29.3\% & 70.7\%\\
\midrule
\benchmark (Image Options) & 72.5\% & 27.5\%\\
\benchmark (Text Options) & 98.3\% & 1.7\% \\
\bottomrule 
\end{tabular} 
}
\vspace{-3mm}
\caption{Answering Strategy}
\label{tab:answer_strategy} 
\vspace{-4mm}
\end{wraptable}
\autoref{tab:answer_strategy} shows the statistics of Claude-3.7-Sonnet-Thinking's answering strategy. We observe a clear divergence in answering strategies between MMMU and \benchmark. 
On MMMU, the model tend to follow an option-driven approach—using the provided choices early to eliminate unlikely answers and select the most relevant one, often without explicitly solving the problem.
In contrast, models more frequently adopt an answer-first strategy on \benchmark, attempting to solve the question independently before comparing the result to the answer choices. This pattern holds across both textual and image-based options, though the option-first approach appears slightly more often (around 30\%) for image-based tasks—likely due to the added complexity of visual comparison.

\subsection{Does model performance transfer between reasoning categories?}

\begin{wrapfigure}{l}{0.45\textwidth}
  \vspace{-8mm}
  \begin{center}
    \includegraphics[width=0.45\textwidth]{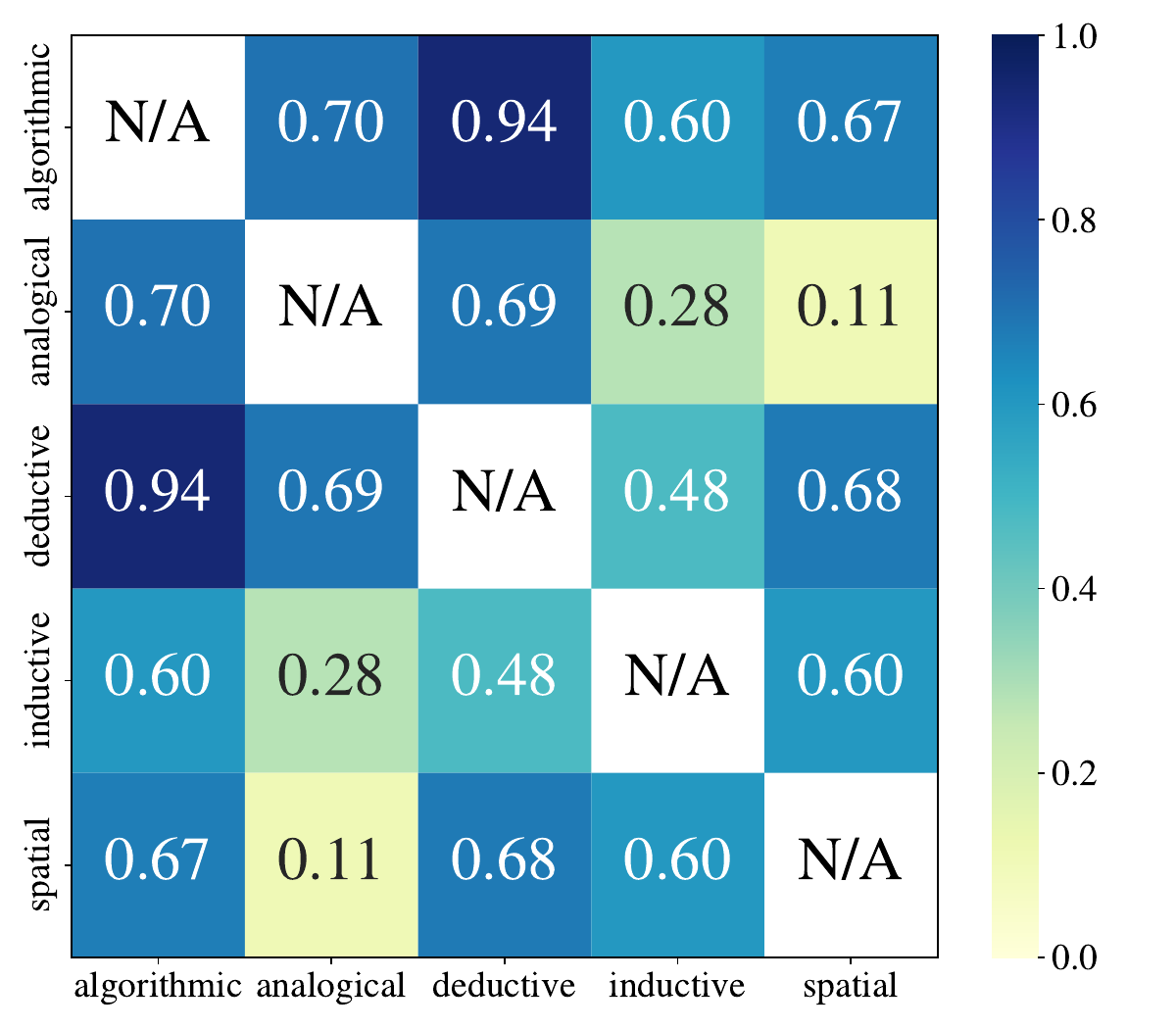}
  \end{center}
  \vspace{-3mm}
  \caption{Correlation Heatmap among reasoning categories for models (averaged across all models we evaluated).}
  \vspace{-5mm}
  \label{fig:reasoning_correlation}
\end{wrapfigure}
\autoref{fig:reasoning_correlation} presents a correlation heatmap illustrating the relationships among the five reasoning categories in \benchmark. We report model correlations averaged across all models in \autoref{tab:model_performance}.
For humans, each reasoning category likely engages different cognitive or mental processes \citep{goel2004differential,green2010connecting,bright2014causal,babcock2015interaction}, so performance in one category might not transfer to performance in another. 
However, the correlation heatmap of the models tells a different story. We observe notably strong correlations across reasoning categories, with values ranging from 0.11 to as high as 0.94. In particular, algorithmic and deductive reasoning show high correlation (0.94), and other pairs such as algorithmic-analogical and deductive-analogical also exhibit strong associations. This suggests that model performance tends to generalize across categories.
However, this generalization may not reflect true reasoning abilities. Instead, the high correlations could indicate that models are leveraging shared surface-level patterns or shortcut strategies that happen to work across multiple structurally different categories, unlike humans, who may rely on distinct cognitive processes. 

\subsection{Error Analysis}

\begin{wrapfigure}{r}{0.35\textwidth}
  \vspace{-15mm}
  \begin{center}
    \includegraphics[width=0.35\textwidth]{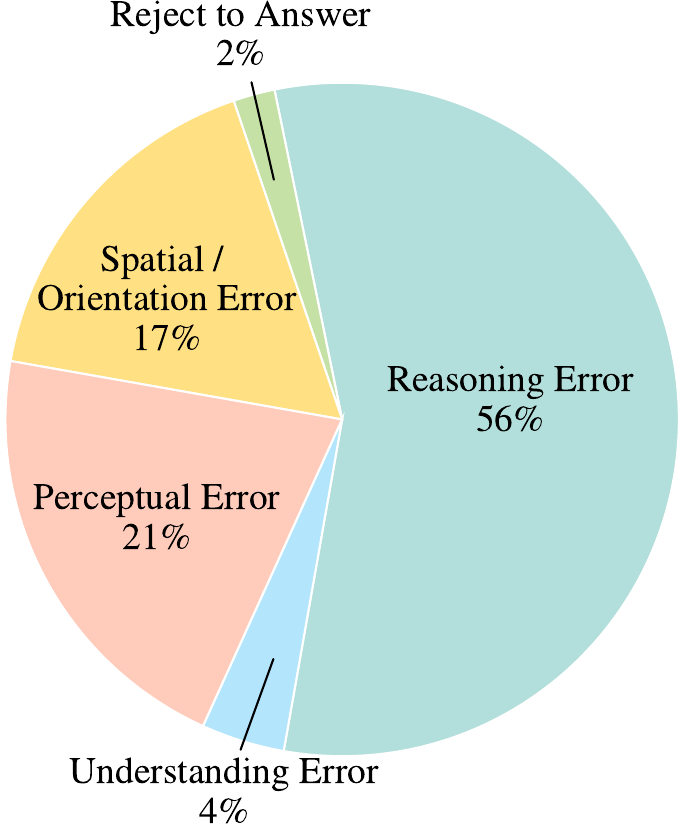}
  \end{center}
  \vspace{-4mm}
  \caption{Error Distribution of Claude-3.7-Sonnet-Thinking}
  \vspace{-8mm}
  \label{fig:error}
\end{wrapfigure}
\autoref{fig:error} shows a pie chart illustrating the distribution of error categories of 100 instances generated by Claude-3.7-Sonnet-Thinking on \benchmark, revealing that reasoning errors dominate at 56\%, reinforcing the fact that reasoning is greatest challenge to models in \benchmark. Perceptual errors (21\%) and spatial / orientation errors (17\%) also constitute substantial portions of failures, reflecting difficulties in interpreting visual elements and understanding spatial relationships. These three categories together account for 94\% of mistakes, emphasizing a need for multimodal models with stronger reasoning capabilities with more robust perception and spatial understanding. Textual and visual understanding errors (4\%) and reject-to-answer cases (2\%) are relatively rare. \autoref{app:case_study} shows samples of error and correct cases of each reasoning and difficulty category.

%% file: sections/5_Related_work.tex
\section{Related Work}

\noindent\textbf{Multimodal Language Models (MLLMs)}, particularly vision language models have experienced significant improvements recently. Large scale vision language models \citep{deepmind_gemini_report}; \citep{gpt4o}; \citep{Claude}; including open weight ones \citep{li2024llavaov}; \citep{yue2025pangea}; \citep{Liu2024HarnessingWU}; \citep{tong2024cambrian}; \citep{dubey2024llama} are capable of utilizing both image and text inputs to solve challenging questions. 

Multimodal reasoning models, models that specialize in complex reasoning, further push the boundary of MLLMs' capabilities. Large scale multimodal reasoning models such as QVQ \citep{qvq}, Claude-3.7-Sonnet-thinking \citep{Claude}, o1 \citep{o1}, Gemini-2.0-flash-thinking \citep{deepmind_gemini_report} excel in reasoning heavy tasks such as coding and solving math problems.

\noindent\textbf{Multimodal Reasoning Benchmarks}. There exists a number of multimodal benchmarks that test on both the models' world knowledge and reasoning abilities. These benchmarks \citep{yue2023mmmu}; \citep{Marino2019OKVQAAV}; \citep{Liu2023MMBenchIY};  \citep{yue2024mmmu}; \citep{Phan2025HumanitysLE} emphasize on the multimodal ability of models as a whole, without further separation of knowledge and reasoning. 

Recently, more multimodal benchmarks have placed emphasis on multimodal logical reasoning abilities. Many of them \citep{lu2023mathvista}; \citep{wang2024measuringmultimodalmathematicalreasoning} focus primarily on mathematic problems, testing on both mathematical knowledge and reasoning. Some others cover on more general logical reasoning problems \citep{Cherian2022AreDN}; \citep{gao2023lora}, testing on both models' knowledge and reasoning in different domains.

%% file: sections/6_Conclusion.tex
\section{Conclusion and Future Work}

We presented \benchmark, a novel multimodal benchmark carefully designed to minimize the impact of domain-specific knowledge and isolate models’ core reasoning capabilities. Our results show that while proprietary and large-scale open models achieve relatively higher performance, they still fall short of human-level reasoning—especially on more complex tasks such as analogical and inductive reasoning. Moreover, we observe that strong performance on knowledge-intensive benchmarks like MathVista and MMMU does not necessarily translate into high accuracy on \benchmark, underscoring the distinct challenge of knowledge-light reasoning tasks.

These findings suggest that purely scaling model size and knowledge resources may not suffice for robust multimodal reasoning skills; rather, methods that promote structured reasoning, such as explicit thinking modes or recursive reasoning steps, can offer substantial improvements, particularly for hard questions. Future research can explore new training strategies, specialized architectures, or model interpretations tailored to reduce reliance on memorized facts and enhance logical inference. Extending \benchmark to include additional types of multi-image reasoning or temporally dynamic visual information may further stress-test models’ core inference abilities. By disentangling domain knowledge from multimodal reasoning, we hope \benchmark will serve as a valuable tool for developing and evaluating next-generation MLLMs that excel at genuinely understanding and reasoning about the world without depending heavily on specialized factual knowledge.

%% file: sections/7_Limitations.tex
\section{Limitations}

\paragraph{Disentangling Knowledge}
Despite our best efforts to isolate domain-specific knowledge from the evaluation of multimodal reasoning, \benchmark is still not entirely free of knowledge dependencies. Basic familiarity with everyday objects or common scenarios is still required; complete knowledge free evaluation remains an ideal rather than a practical reality. 

\paragraph{Real World Application}
\benchmark emphasizes puzzle-like questions that may not reflect the full diversity of real-world scenarios, limiting generalizability to more specialized domains. 

\paragraph{Question Format}
\benchmark focuses on multiple-choice questions, which may not capture the breadth of open-ended reasoning tasks where models must generate complex textual or visual outputs. 

Future work can address these limitations by including more varied question formats, broader domains, and more granular analyses of a model's knowledge versus its multimodal reasoning abilities.

%% file: sections/8_Ethical_statement.tex
\section{Ethical Statement}

This paper uses samples extracted from existing quiz sources for scholarly analysis and testing purposes, in accordance to US fair use law and standard practice. These data are neither intended for, nor capable of, substituting for the original works; thus, we believe their inclusion does not diminish the market value or utility of the source materials. A complete list of references for the data sources is attached in \autoref{app:stats}.

%% file: sections/Acknowledgements.tex
\section*{Acknowledgements}

This project was supported in part by a grant from DSTA Singapore and the Carnegie Bosch Institute. The authors would like to thank CMU NeuLab colleagues for their constructive comments. The authors would also like to thank all volunteers who participated in the human evaluation.

%% file: sections/Appendix.tex
\section{\benchmark Statistics}

\label{app:stats}

\subsection{Breakdown of Statistics of \benchmark}

\autoref{tab:question_stats} shows a breakdown of statistics of \benchmark questions.
\begin{table*}[!h]
    \centering
   
    \begin{tabular}{lccccccc}
        \toprule
        \multirow{2}{*}{Reasoning Category} & 
        \multicolumn{3}{c}{Image Options} & \multicolumn{3}{c}{Text Options} & \multirow{2}{*}{Total}\\
        \cmidrule(lr){2-4} \cmidrule(lr){5-7}
        & Easy & Medium & Hard & Easy & Medium & Hard \\
        \midrule
        Algorithmic & 21 & 8 & 0 & 124 & 100 & 9 & 262 \\
        Analogical & 120 & 81 & 10 & 0 & 0 & 0 & 211 \\
        Deductive & 29 & 24 & 2 & 45 & 79 & 21 & 200 \\
        Inductive & 7 & 70 & 127 & 3 & 2 & 0 & 209 \\
        Spatial & 123 & 41 & 6 & 61 & 52 & 3 & 286 \\\midrule
        Total & 300 & 224 & 145 & 233 & 233 & 33 & 1168\\
        \bottomrule
    \end{tabular}
    \caption{Number of questions in each reasoning category,  option types, and difficulty levels.}
    \label{tab:question_stats}
\end{table*}

\subsection{Data Sources}
\begin{itemize}
    \item Chinese Civil Service Examination \begin{CJK}{UTF8}{gbsn} (中国国家公务员考试) 
\end{CJK}\footnote{\url{https://en.wikipedia.org/wiki/Civil_service_of_the_People\%27s_Republic_of_China\#Examinations}. } (224 puzzles): we manually translated questions from this exam to English from Chinese. 
    \item Textbooks (210 puzzles): we carefully collected and re-purposed questions from online resources and textbooks.
    \item Smart-101 \citep{cherian2022deep} (247 puzzles): we carefully selected images from this benchmark and synthesized new questions.
    \item MATH-Vision \citep{wang2024mathvision} (293 puzzles): we carefully selected and re-purposed questions from this benchmark.
    \item VASR \citep{bitton2023vasr} (194 puzzles): we carefully selected questions from this benchmark.
\end{itemize}
\section{Model Evaluation Setup}

\begin{promptbox}[Model Evaluation Prompt with Chain-of-Thought]{knowledge}
Solve the multiple-choice question and then answer with the option letter from the given choices. The last line of your response should be of the following format: `Answer: \$LETTER' (without quotes) where LETTER is one of options. Think step by step before answering.
\end{promptbox}

\begin{promptbox}[Model Evaluation Prompt w/n Chain-of-Thought]{knowledge}
Answer the question with the option's letter from the given choices directly.
\end{promptbox}

\section{Human Annotation Setup}

\subsection{Difficulty Labeling}
Each question was also carefully assigned a difficulty label from easy, medium, or hard, based on the cognitive load required for reasoning.

\begin{itemize}
    \item \textbf{Easy Level} questions could be solved by the annotator in less than one minute.
    \item \textbf{Medium Level} questions could be solved by the annotator in one to three minutes.
    \item \textbf{Hard Level} questions require the annotator more than five minutes to solve or quit solving.
\end{itemize}

\begin{promptbox}[Annotation Guideline for Puzzle Difficulty]{darkblue}

Try to solve the puzzle first. You need to measure the time you attempted to solve each puzzle. Then, select from Easy, Medium, or Hard based on the time required. \\
- Easy Level: You can solve or answer the question within 1 minute. This level of puzzles should require minimal reasoning. \\
- Medium Level: You can solve or answer the question within 1-3 minutes. This level of puzzles should 
demand moderate reasoning. \\
- Hard Level: You can / cannot solve this question with more than 5 minutes. This level of puzzles should involve significant / multi-step reasoning. 
\end{promptbox}

\subsection{Reasoning Category Labeling}

\begin{promptbox}[Annotation Guideline for Puzzle Reasoning Category]{darkblue}
Assign the category that \emph{best} describes the primary type of reasoning or logic required for each puzzle:

- Algorithmic Reasoning: Involves following or devising a step-by-step procedure or rule-based process.\\
- Analogical Reasoning: Requires identifying relationships by comparison between pairs of entities.\\
- Deductive Reasoning: Involves deriving specific conclusions from general or given premises. \\
- Inductive Reasoning: Focuses on generalizing a rule or pattern from specific instances.\\
- Spatial Reasoning: Involves visualizing and manipulating shapes, distances, or orientations.  
\end{promptbox}

\section{Full Results}

\label{app:full}

\subsection{Full Results w/ CoT}
\begin{table*}[!htbp]
\centering
\resizebox{\linewidth}{!}{
\begin{tabular}{@{}lcccccc@{}}
\toprule
\textbf{Model} & 
\textbf{Algorithmic} & \textbf{Analogical} & \textbf{Deductive} & \textbf{Inductive} &
\textbf{Spatial} & \textbf{Overall}\\
\midrule
\color{Gray}\textbf{Random Choice} & \color{Gray}25.0 & \color{Gray}25.0 & \color{Gray}25.0 & \color{Gray}25.0 & \color{Gray}25.0 & \color{Gray}25.0 \\
\textbf{Human (95th Percentile)} & 100.0 & 100.0 & 100.0 & 81.6 & 100.0 & 89.3 \\
\textbf{Human (50th Percentile)} & 88.0 & 66.0 & 80.0 & 50.0 & 90.0 & 75.0 \\
\textbf{Human (5th Percentile)} & 68.1 & 25.0 & 37.0 & 0.0 & 59.1 & 57.5 \\
\midrule
\multicolumn{7}{c}{\textit{Proprietary Models}} \\
\midrule
\textbf{o4-mini} & 65.3 & 68.7 & 75.5 & 33.0 & 45.5 & 57.0 \\
\textbf{o3} & 64.5 & 68.3 & 69.5 & 27.3 & 42.7 & 54.0 \\
\textbf{o1} & 63.7 & 68.3 & 67.5 & 29.2 & 34.3 & 51.8 \\
\textbf{GPT-4o} & 49.2 & 58.3 & 49.0 & 27.3 & 26.2 & 41.3 \\
\noalign{\vspace{2pt}} 
\hdashline
\noalign{\vspace{2pt}} 
\textbf{Gemini-2.5-pro} & 60.0 & 64.0 & 60.0 & 29.7 & 36.4 & 49.5 \\
\textbf{Gemini-2.0-flash} & 55.3 & 58.8 & 57.0 & 24.4 & 31.8 & 45.0 \\
\textbf{Gemini-2.0-flash-thinking} & 46.6 & 70.1 & 49.0 & 24.9 & 25.5 & 42.2 \\
\textbf{Gemini-1.5-Pro} & 53.4 & 57.4 & 58.5 & 26.3 & 32.5 & 45.0 \\
\noalign{\vspace{2pt}} 
\hdashline
\noalign{\vspace{2pt}} 
\textbf{Claude-3.7-Sonnet} & 64.5 & 48.3 & 65.0 & 26.8 & 37.4 & 48.3 \\
\textbf{Claude-3.7-Sonnet-thinking} & 67.2 & 44.1 & 61.5 & 31.1 & 37.1 & 48.2 \\
\textbf{Claude-3.5-Sonnet} & 53.4 & 47.9 & 51.5 & 25.4 & 34.3 & 42.4 \\
\midrule
\multicolumn{7}{c}{\textit{Open Models}} \\
\midrule
\textbf{LLaVA-1.5-7B} & 23.3 & 21.8 & 36.0 & 20.6 & 19.2 & 23.7 \\
\textbf{LLaVA-1.5-13B} & 24.8 & 21.8 & 23.0 & 25.4 & 25.5 & 24.2 \\
\textbf{LLaVA-1.6-7B} & 27.5 & 23.7 & 30.0 & 22.5 & 21.3 & 24.8\\
\textbf{LLaVA-1.6-13B} & 25.2 & 25.6 & 27.0 & 27.3 & 23.4 & 25.5\\
\textbf{LLaVA-1.6-34B} & 29.4 & 28.0 & 43.0 & 24.9 & 25.5 & 29.7 \\
\textbf{LLaVA-OV-0.5B} & 21.0 & 26.1 & 30.5 & 22.5 & 25.2 & 24.8 \\
\textbf{LLaVA-OV-7B} & 27.9 & 26.1 & 36.5 & 23.4 & 25.5 & 27.7 \\
\textbf{LLaVA-OV-72B} & 34.7 & 26.5 & 37.0 & 27.3 & 28.7 & 30.8 \\
\noalign{\vspace{2pt}} 
\hdashline
\noalign{\vspace{2pt}} 
\textbf{Llama-3.2-11B-Vision-Instruct} & 31.0 & 30.8 & 39.0 & 21.1 & 26.2 & 29.4 \\
\textbf{Llama-3.2-90B-Vision-Instruct} & 45.0 & 23.2 & 43.0 & 26.3 & 31.5 & 34.1 \\
\noalign{\vspace{2pt}} 
\hdashline
\noalign{\vspace{2pt}} 
\textbf{Qwen-VL} & 21.4 & 31.3 & 25.0 & 26.3 & 24.1 & 25.3\\
\textbf{Qwen2-VL-72B} & 41.6 & 28.4 & 39.5 & 22.5 & 29.0 & 32.4 \\
\textbf{QvQ-72B-Preview} & 43.1 & 45.5 & 48.0 & 27.3 & 27.6 & 37.8 \\
\textbf{Qwen2-VL-2B-Instruct} & 26.0 & 26.1 & 24.5 & 27.8 & 25.5 & 26.0 \\
\textbf{Qwen2-VL-7B-Instruct} & 36.3 & 21.8 & 38.5 & 20.6 & 22.7 & 27.9\\
\textbf{Qwen2-VL-72B-Instruct} & 39.9 & 33.5 & 45.2 & 23.5 & 32.4 & 34.9\\
\textbf{Qwen2.5-VL-3B-Instruct}
& 35.1 & 27.5 & 44.5 & 25.8 & 24.8 & 31.2 \\
\textbf{Qwen2.5-VL-7B-Instruct} & 40.5 &26.6 &39.0 &24.0 &29.7 & 32.1 \\
\textbf{Qwen2.5-VL-72B-Instruct} & 53.4 &46.9 &58.0 &25.8 &29.5 &42.3 \\
\noalign{\vspace{2pt}} 
\hdashline
\noalign{\vspace{2pt}} 
\textbf{Cambrian-8B} & 31.3 & 24.2 & 36.0 & 24.0 & 29.0 & 28.9 \\
\textbf{Cambrian-13B} & 24.8 & 25.6 & 39.5 & 24.4 & 21.0 & 26.5 \\
\noalign{\vspace{2pt}} 
\hdashline
\noalign{\vspace{2pt}} 
\textbf{Pangea-7B} & 30.5 & 28.9 & 35.0 & 24.4 & 25.2 & 28.6 \\
\bottomrule
\end{tabular}}
\caption{Performance (\%) of various models with Chain of Thoughts (CoT) on \benchmark. }
\label{tab:model_performance_full_cot}
\end{table*}

\subsection{Full Results w/n CoT}

\begin{table*}[!htbp]
\centering
\resizebox{\linewidth}{!}{
\begin{tabular}{@{}lcccccc@{}}
\toprule
\textbf{Model} & 
\textbf{Algorithmic} & \textbf{Analogical} & \textbf{Deductive} & \textbf{Inductive} &
\textbf{Spatial} & \textbf{Overall}\\
\midrule
\color{Gray}\textbf{Random Choice} & \color{Gray}25.0 & \color{Gray}25.0 & \color{Gray}25.0 & \color{Gray}25.0 & \color{Gray}25.0 & \color{Gray}25.0 \\
\textbf{Human (95th Percentile)} & 100.0 & 100.0 & 100.0 & 81.6 & 100.0 & 89.3 \\
\textbf{Human (50th Percentile)} & 88.0 & 66.0 & 80.0 & 50.0 & 90.0 & 75.0 \\
\textbf{Human (5th Percentile)} & 68.1 & 25.0 & 37.0 & 0.0 & 59.1 & 57.5 \\
\midrule
\multicolumn{7}{c}{\textit{Proprietary Models}} \\
\midrule
\textbf{GPT-4o} & 40.8 & 34.1 & 40.5 & 24.9 & 29.7 & 34.0 \\
\noalign{\vspace{2pt}} 
\hdashline
\noalign{\vspace{2pt}} 
\textbf{Gemini-2.0-flash} & 57.6 & 41.7 & 58.0 & 23.0 & 35.7 & 43.2 \\
\textbf{Gemini-1.5-Pro} & 51.2 & 46.5 & 54.0 & 24.9 & 29.4 & 40.8 \\
\midrule
\multicolumn{7}{c}{\textit{Open Models}} \\
\midrule
\textbf{LLaVA-1.5-7B} & 24.4 & 24.7 & 34.5 & 26.8 & 25.5 & 26.9 \\
\textbf{LLaVA-1.5-13B} & 24.4 & 26.1 & 33.5 & 26.3 & 28.3 & 27.6 \\
\textbf{LLaVA-1.6-7B} & 27.5 & 25.1 & 32.5 & 24.9 & 27.3 & 27.4\\
\textbf{LLaVA-1.6-13B} & 21.4 & 24.7 & 29.5 & 28.2 & 23.1 & 25.0\\
\textbf{LLaVA-1.6-34B} & 31.3 & 27.3 & 43.0 & 24.4 & 27.6 & 29.8 \\
\textbf{LLaVA-OV-0.5B} & 24.4 & 25.6 & 37.5 & 24.9 & 25.5 & 27.2 \\
\textbf{LLaVA-OV-7B} & 27.5 & 28.0 & 40.5 & 24.4 & 28.0 & 29.4 \\
\textbf{LLaVA-OV-72B} & 31.7 & 23.6 & 45.0 & 21.3 & 24.6 & 28.8 \\
\noalign{\vspace{2pt}} 
\hdashline
\noalign{\vspace{2pt}} 
\textbf{Llama-3.2-11B-Vision-Instruct} & 27.5 & 24.2 & 31.0 & 26.3 & 27.6 & 27.3 \\
\textbf{Llama-3.2-90B-Vision-Instruct} & 38.2 & 22.3 & 44.5 & 25.8 & 33.6 & 33.1 \\
\noalign{\vspace{2pt}} 
\hdashline
\noalign{\vspace{2pt}} 
\textbf{Qwen-VL} & 23.7 & 26.5 & 29.5 & 27.8 & 26.6 & 26.6\\
\textbf{Qwen2-VL-72B} & 38.9 & 28.4 & 43.0 & 20.6 & 29.0 & 32.0 \\
\textbf{QvQ-72B-Preview} & 44.8 & 43.6 & 44.0 & 26.8 & 30.8 & 37.8 \\
\textbf{Qwen2-VL-2B-Instruct} & 31.7 & 29.4 & 40.5 & 23.9 & 31.5 & 31.3 \\
\textbf{Qwen2-VL-7B-Instruct} & 33.6 & 24.2 & 46.0 & 22.5 & 26.2 & 30.2\\
\textbf{Qwen2-VL-72B-Instruct} & 40.5 & 30.3 & 46.0 & 25.4 & 29.4 & 34.2\\
\textbf{Qwen2.5-VL-3B-Instruct}
& 36.3 & 26.1 & 47.0 & 25.8 & 22.4 & 31.0 \\
\textbf{Qwen2.5-VL-7B-Instruct} & 38.2 &23.7 &51.5 &24.9 &31.1 & 33.7 \\
\textbf{Qwen2.5-VL-72B-Instruct} & 43.1 &40.3 &51.5 &25.4 &33.7 &38.6 \\
\noalign{\vspace{2pt}} 
\hdashline
\noalign{\vspace{2pt}} 
\textbf{Cambrian-8B} & 25.2 & 20.4 & 35.0 & 23.0 & 20.6 & 24.5 \\
\textbf{Cambrian-13B} & 23.3 & 28.0 & 36.5 & 24.9 & 26.2 & 27.4 \\
\noalign{\vspace{2pt}} 
\hdashline
\noalign{\vspace{2pt}} 
\textbf{Pangea-7B} & 32.4 & 23.7 & 38.5 & 28.7 & 32.5 & 31.3 \\
\bottomrule
\end{tabular}}
\caption{Performance (\%) of various models with Multiple Choice Direct prompting on \benchmark. }
\label{tab:model_performance_full_direct}
\end{table*}

\section{Knowledge Checklist}
\label{app:knowledge}

\subsection{Knowledge Checklist Generation}
\begin{promptbox}[Prompt to Generate Knowledge Checklist Questions]{thinking}
You are an exam writer. You are now writing a knowledge test. You are given a question (Question) regarding an image and its standard solution (Solution), your task is to write free response questions that test on individual knowledge required in answering the question correctly. \\

You should follow these steps to complete the task:\\
1. explicitly analyze the given image, Question, and Solution\\
2. explicitly list out the individual knowledge concepts required to reach Solution. \\
3. write free response questions to test on the definition of each concept listed. Your generated questions should not include details of the given Question. Note that you need to provide answer keys to these questions too. \\
4. format the free response questions in json format.\\

Question: {question}\\
Solution: {answer}
\end{promptbox}

\subsection{Example Knowledge Checklist Question}

\begin{promptbox}[Example Knowledge Checklist Question (MMMU)]{thinking}
- Question: Explain the Arbitrage Pricing Theory (APT) model and its purpose in finance. \\
- Answer: The Arbitrage Pricing Theory (APT) model is a financial theory that estimates the expected return on an asset based on the asset's sensitivity to various macroeconomic factors. It is used to determine the fair price of an asset by considering multiple factors that could affect its return, as opposed to relying on a single market index as in the Capital Asset Pricing Model (CAPM).
\end{promptbox}

\begin{promptbox}[Example Knowledge Checklist Question (\benchmark)]{thinking}
- Question: What is the definition of distance in a geometric context? \\
- Answer: Distance in a geometric context refers to the measurement of space between two points.
\end{promptbox}

\subsection{Knowledge Checklist Human Annotation}
\label{app:knowledge_human}

We asked two human annotators to manually verify and correct the knowledge checklist questions and gave them the following instructions.
The inter-annotator agreement rate is 87.8\%.

\begin{promptbox}[Human Annotation Instructions]{thinking}
You are given a json file, where each item contains the following elements:

- Question: a multiple-choice question.\\
- Answer: the answer to the question with an optional explanation.\\
- Knowledge Concept Checklist: a list of question-answer pairs, where each question in the list is intended to represent a distinct knowledge concept necessary for solving the Question.\\

You task is to annotate the knowledge concept checklists generated by a model. You should carefully evaluate each question-answer pair based on the following criteria:

1. Necessity: Is the question genuinely necessary for solving the problem? If not, then remove the question.
   
2. Repetition: Check if any questions are repetitive or duplicate existing questions within the list. If the question is repetitive or duplicate, then remove the question.

3. Completeness: Ensure no critical knowledge concepts required to solve the problem are missing, and identify if any additional important questions should have been included.

4. Correctness: Verify whether the provided answers are accurate. Revise the checklist in case of incorrect checklist QA pairs.

5. Knowledge v.s. Skills: Ensure each question explicitly evaluates a knowledge concept rather than testing skills or problem-solving techniques. Remove any questions that primarily evaluate skills instead of knowledge.
\end{promptbox}

\section{Reasoning Complexity}
\label{app:reasoning_complexity}

\begin{promptbox}[Instruction Prompt to Solve Questions in Detailed Steps]{Gray}
$<Question> <Image>$

Solve this question with First Order Logic. Write out each thinking step explicitly, do not skip steps.  

In your response, begin each step with \_\_\_STEP\_START\_\_\_

step $<step\_num>$
\end{promptbox}

\section{Comparison with Other Benchmarks}

\begin{table}[h!]
\centering
\resizebox{\columnwidth}{!}
{
\begin{tabular}{lccccccc}
\toprule
\textbf{Dataset} & \textbf{Size} &  \makecell{\textbf{Reasoning} \\ \textbf{Load}}  &  \makecell{\textbf{Knowledge} \\ \textbf{Requirement}}  & \makecell{\textbf{\% Easy}\\ \textbf{Words}} \textbf{Question Type} & \textbf{Answer Type}\\
\midrule
LogiQA & 0.7K & Heavy & Light & 52.0& Text & Text \\ 
GSM8K & 8.5K & Heavy & Heavy & 54.0 & Text & Text \\ 
WikiDiverse & 0.8K & Light & Heavy & 35.8 & Image+Text & Text \\
MathVista & 6.1K & Heavy & Heavy & 51.9 & Image+Text & Text \\
MMMU & 11.5K & Heavy & Heavy & 46.4 & Image+Text & Text \\
MATH-Vision & 3.0K & Heavy & Heavy & 53.8 & Image+Text & Image+Text \\
MathVerse & 2.6K & Heavy & Heavy & 38.2 & Image+Text & Text \\
LogicBench & 1.5K & Heavy & Light & 53.6 & Text & Text \\ 
LogicVista & 0.4K & Heavy & Heavy & 41.2 & Image+Text & Image \\
NaturalBench & 10K & Light & Light & 52.5 & Image+Text & Text \\
\benchmark & 1.2K & Heavy & Light & \textbf{54.1} & Image+Text & Image+Text \\
\bottomrule
\end{tabular}}
\caption{Comparison of other existing benchmarks with \benchmark}
\label{tab:comparison}
\end{table}

\begin{figure*}[!h]
    \centering
    \includegraphics[width=\linewidth/2]{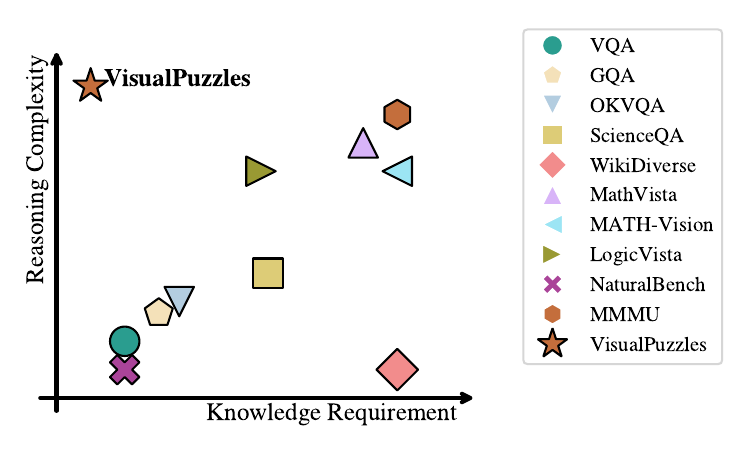}
    \caption{Comparison between \benchmark and several widely-used benchmarks.} 
    \label{fig:comparison}
\end{figure*}

\autoref{fig:comparison} provides a comparative analysis between \benchmark and several widely-used benchmarks for multimodal reasoning, visualizing the knowledge requirement and reasoning complexity of each benchmark. \benchmark has high reasoning complexity and low knowledge requirement, with an aim to disentangle multimodal reasoning from domain-specific knowledge to evaluate general reasoning abilities in non-expert settings.

\begin{table}[h!]
\centering{}
{
\begin{tabular}{lcccccc}
\toprule
\textbf{Model} & \textbf{MathVista} & \textbf{MMMU} & \textbf{\benchmark} \\
\midrule
Human & 60.3 & 88.6 & 80.1 \\
o1 & 73.9 & 78.2 & 51.8 \\
GPT-4o & 63.8 & 69.1 & 41.1 \\
Gemini-2.0-Flash & - & 71.7 & 45.0 \\
Gemini-1.5-Pro & 63.9 & 62.2 & 45.4 \\
Claude-3.5-Sonnet & 67.7 & 68.3 & 42.4 \\
Claude-3.7-Sonnet & - & 71.8 & 48.3 \\
Claude-3.7-Sonnet (Thinking) & - & 75.0 & 48.3 \\
LLaVA-1.5-7B & - & 36.2 & 26.9\\
LLaVA-1.5-13B & 27.6 & 36.4 & 27.6\\
LLaVA-NeXT-7B & 35.8 & 34.6 & 27.4\\
LLaVA-NeXT-13B & 36.2 & 35.3 & 25.3 \\
LLaVA-NeXT-34B & 46.5 & 51.1 & 29.8 \\
LLaVA-OV-0.5B & 34.8 & 31.4 & 27.2 \\
LLaVA-OV-7B & 63.2 & 48.8 & 29.4 \\
LLaVA-OV-72B & 67.5 & 56.8 & 31.8 \\
Llama-3.2-11B-Vision-Instruct & 51.5 & 50.7 & 29.4 \\
Llama-3.2-90B-Vision-Instruct & 57.3 & 60.3 & 34.3 \\
Qwen2-VL-72B & 70.5 & 64.5 & 32.1 \\
QvQ-72B-Preview & 71.4 & 70.3 & 37.9 \\
Qwen2-VL-2B-Instruct & 43.0 & 41.1 & 31.3 \\
Qwen2-VL-7B-Instruct & 58.2 & 54.1 & 30.2 \\
Qwen2-VL-72B-Instruct & 70.5 & 64.5 & 34.9 \\
Qwen2.5-VL-3B-Instruct & 62.3 & 53.1 & 31.2 \\
Qwen2.5-VL-7B-Instruct & 68.2 & 58.6 & 33.7 \\
Qwen2.5-VL-72B-Instruct & 74.8 & 70.2 & 42.3 \\
Cambrian-8B & 49.0 & 42.7 & 28.5 \\
Cambrian-13B & 48.0 & 40.0 & 27.4 \\
\bottomrule
\end{tabular}}
\caption{Comparison of other MathVista and MMMU with \benchmark on human and SOTA models}
\label{tab:comparison_results}
\end{table}

\autoref{tab:comparison_results} compare the performance of various model families across MathVista, MMMU, and \benchmark. Both MathVista and MMMU are benchmarks that have a heavy emphasis on both knowledge and reasoning, whereas \benchmark assess models on domain-disentangled multimodal reasoning alone. We found that success on knowledge-intensive multimodal reasoning benchmarks as MathVista and MMMU does not always carry over to \benchmark that emphasize reasoning rather than extensive pre-trained knowledge.

\section{Additional Analysis}

\subsection{Proprietary V.S. Open Models}

From \autoref{tab:model_performance}, proprietary models (e.g., o4-mini and Claude-3.7-Sonnet) consistently achieve higher overall accuracy than most open-source models on \benchmark. However, some open models also show competitive or even higher performance in both the overall accuracy and specific reasoning categories. 
For instance, Qwen2.5-VL-72B-Instruct demonstrates higher performance than GPT-4o on algorithmic reasoning, deductive reasoning, spatial reasoning, and overall accuracy.
This indicates that while proprietary models currently have leading performance, open models are also rapidly improving on multimodal reasoning capabilities.

\subsection{Reasoning Category and Difficulty Levels}

\begin{figure*}[!h]
    \centering
    \includegraphics[width=\linewidth]{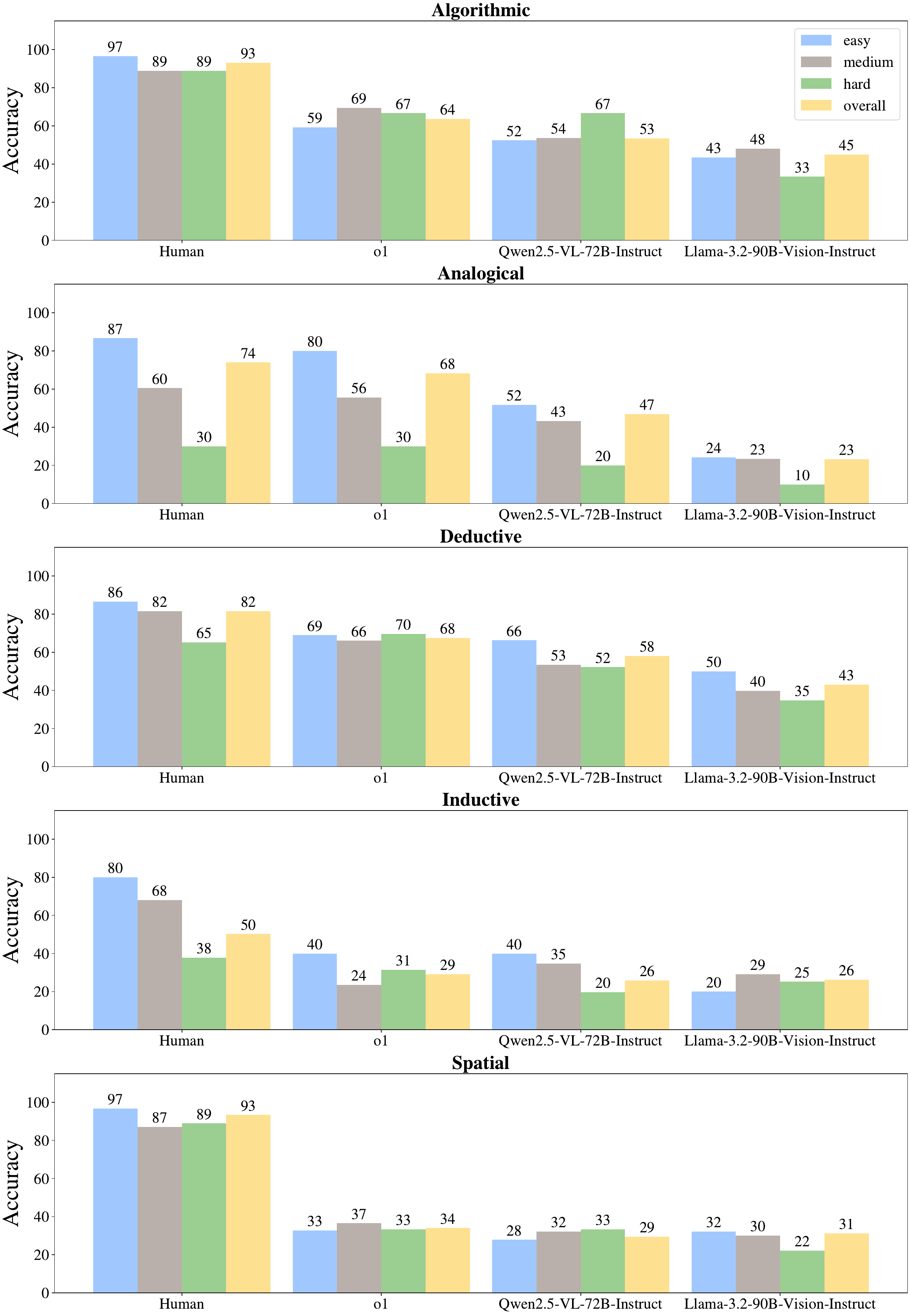}
    \caption{Comparison of accuracy across different reasoning categories for human participants, one of the best performing proprietary models o1, the best performing Qwen-based open model Qwen2.5-VL-72B-Instruct, and the best performing Llama-based open model Llama-3.2-90B-Vision-Instruct, measured on difficulty levels.} \label{fig:cataegory_difficulty}
\end{figure*}

\begin{figure*}[!h]
    \centering
    \includegraphics[width=\linewidth]{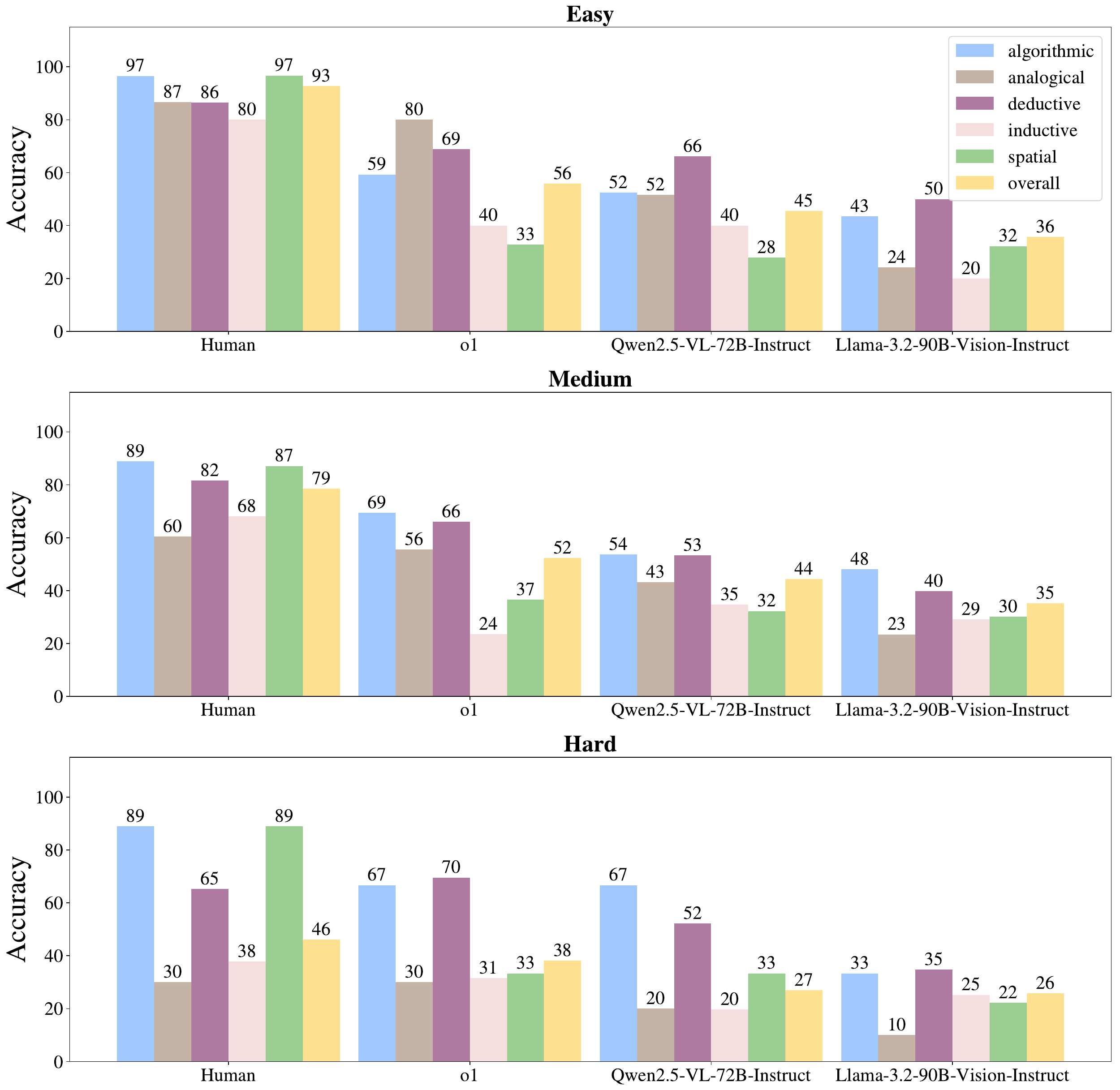}
    \caption{Comparison of accuracy across different difficulty levels for human participants, one of the best performing proprietary models o1, the best performing Qwen-based open model Qwen2.5-VL-72B-Instruct, and the best performing Llama-based open model Llama-3.2-90B-Vision-Instruct, measured across reasoning categories.} \label{fig:reasoning_categories}
\end{figure*}

\autoref{fig:reasoning_categories} and \autoref{fig:cataegory_difficulty} present complementary views of human accuracy against three representative models: o1 (one of the best-performing proprietary models), Qwen2.5-VL-72B-Instruct (the strongest Qwen-based open model), and Llama-3.2-90B-Vision-Instruct (the strongest Llama-based open model). Specifically, \autoref{fig:cataegory_difficulty} compares performance across difficulty levels for each reasoning category, while \autoref{fig:reasoning_categories} compares performance across categories within each difficulty level.

Humans consistently outperform all models across categories and difficulty levels, often by large margins. Notably, human performance remains high and relatively stable in the algorithmic, deductive, and spatial categories, even on hard questions. While accuracy does decline in analogical and inductive reasoning as difficulty increases, humans still maintain a clear advantage over models.

In contrast, model performance declines sharply as difficulty increases, especially for open-source models. Accuracy of Llama-3.2-90B-Vision-Instruct on hard analogical tasks drops to just 10\%. Even one of the strongest proprietary models, o1, while more robust, still lags significantly behind humans, particularly on analogical, inductive, and spatial tasks. On easy tasks, some models perform competitively in certain categories, but this advantage largely disappears on medium and hard questions.

Interestingly, these models maintain a generally stable performance on algorithmic and deductive reasoning. For o1 and Qwen2.5-VL-72B-Instruct, their performances on algorithmic reasoning even go up for more difficult tasks, whereas human performance degraded as the difficulty level increases. However, all models, including o1, perform the worse at analogical, inductive and spatial reasoning in general, especially as the difficulty level increases. This suggests that models are relatively better at tasks requiring structured, rule-based algorithmic processing, while their performance degrades more steeply in tasks requiring relational abstraction (analogical), pattern induction (inductive), and visual understanding (spatial), particularly as the difficulty level increases. In summary, these results indicate that while some models exhibit promising performance on structured and easier reasoning tasks, multimodal models still struggle with abstract and complex reasoning, particularly when difficulty increases. Bridging the gap between model and human reasoning remains a critical challenge.

\subsection{Option Types and Difficulty Levels}

\begin{figure*}[!h]
    \centering
    \includegraphics[width=\linewidth]{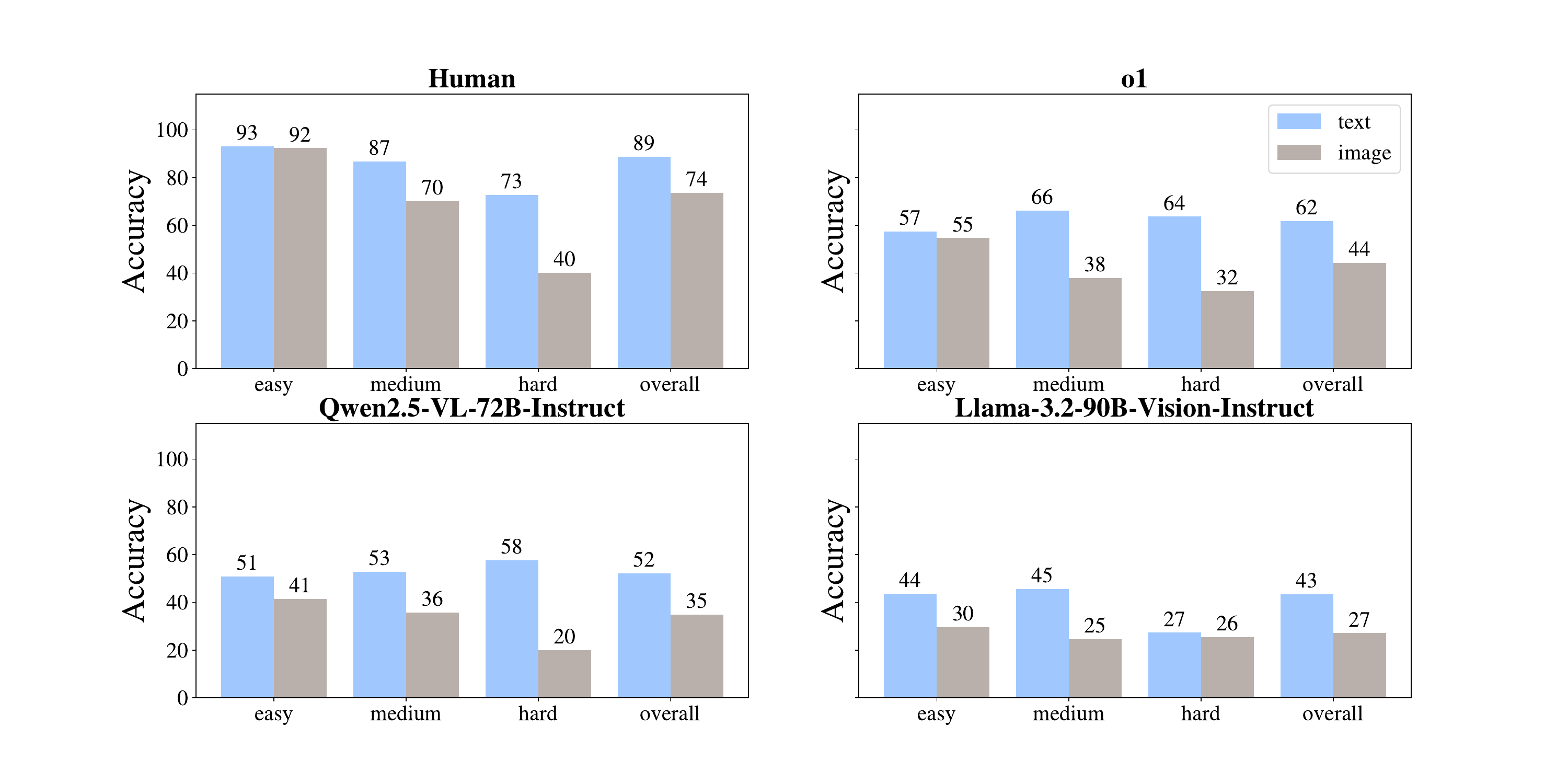}
    \caption{Comparison of accuracy across different difficulty levels for human participants, one of the best performing proprietary model o1, the best performing Qwen-based open model Qwen2.5-VL-72B-Instruct, and the best performing Llama-based open model Llama-3.2-90B-Vision-Instruct, measured on textual v.s. visual option types.} \label{fig:option_difficulty}
\end{figure*}

\autoref{fig:option_difficulty} compares human accuracy against three representative models, o1 (one of the best-performing proprietary models), Qwen2.5-VL-72B-Instruct (the strongest Qwen-based open model), and Llama-3.2-90B-Vision-Instruct (the strongest Llama-based open model), across different difficulty levels, separately for textual and visual answer options.

Across all participants and models, we observe a consistent pattern: text-based options result in higher accuracy than image-based options, with the performance gap widening as task difficulty increases. This trend holds even for human participants, whose accuracy drops from 92\% to 40\% on visual options when moving from easy to hard tasks, compared to a much smaller drop on text-based ones (93\% to 73\%).

For models, the gap is even more pronounced. For instance, Qwen2.5-VL-72B-Instruct achieves 58\% accuracy on hard questions with text options, but only 20\% when image options are used. o1 and Llama-3.2-90B-Vision-Instruct exhibit similar drops, suggesting a broad weakness in multi-image reasoning and visual option discrimination.
These findings suggest that image-based answer options introduce significant additional complexity, requiring models not just to understand the question but to reason over multiple visual cues. This capability is essential for real-world tasks such as product selection, recommendation, and visual planning, where their decision-making process often depends on comparing visual content.

However, most pretraining datasets and benchmarks have traditionally emphasized textual QA formats, with far fewer examples involving visual options or structured visual comparisons. As a result, models may lack the inductive bias or learned attention mechanisms to handle visual alternatives effectively.
These results highlight an important direction for future work: expanding and diversifying training corpora to include multi-choice visual reasoning tasks, and developing architectures that are explicitly designed to process and compare visual candidates, especially under challenging conditions.

\subsection{Case Study of Reasoning}

\begin{figure*}[!h]
    \centering
    \includegraphics[width=\linewidth]{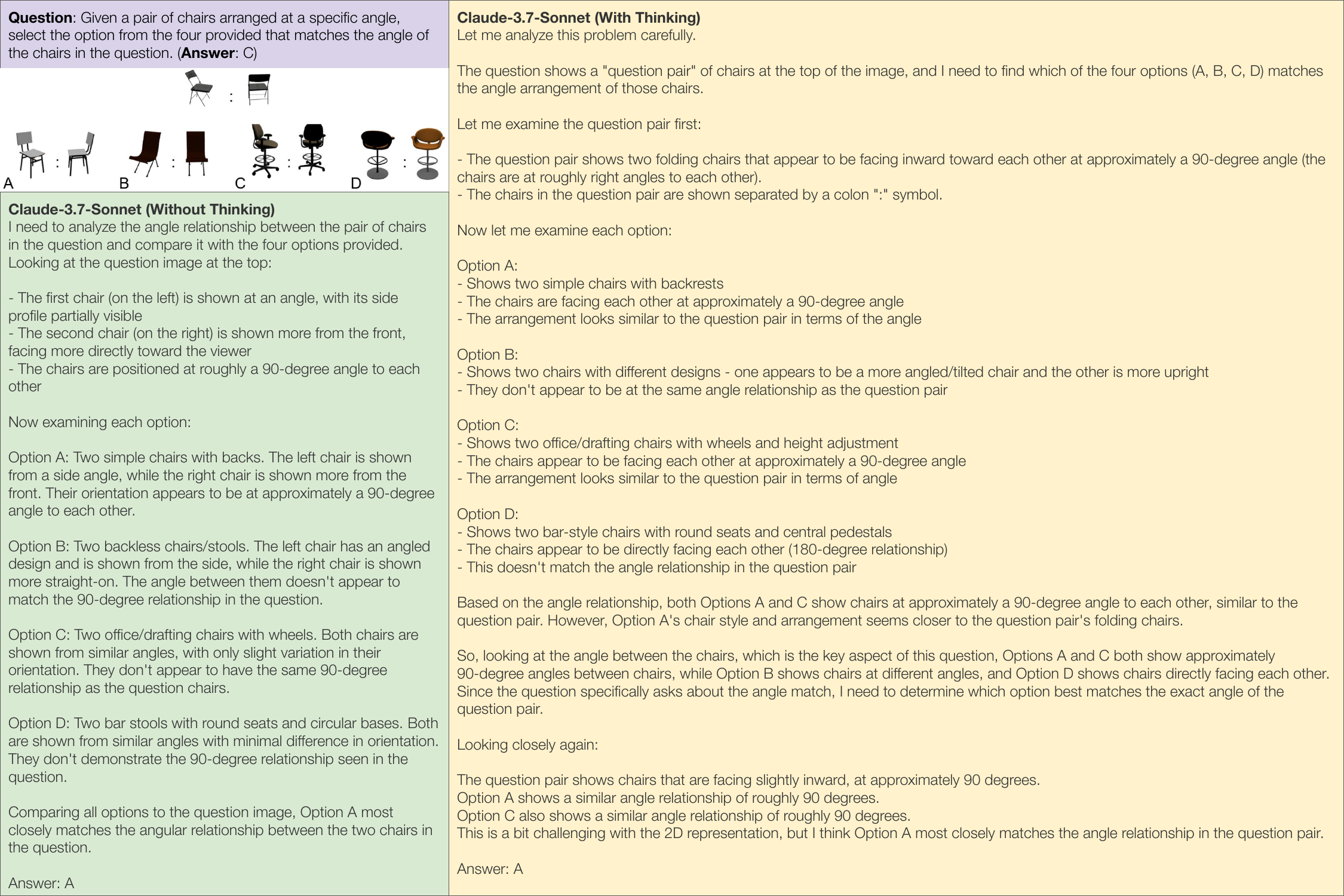}
    \vspace{-6mm}
    \caption{Case Study showing the similarity in structure and reasoning strategy between Claude-3.7-Sonnet-Thinking and Claude-3.7-Sonnet. Similarity between model responses of these two models on \benchmark is 0.9.} 
    \label{fig:reasoning_case_study_1}
\end{figure*}

\autoref{fig:reasoning_case_study_1} shows a case study demonstrating the similarity in structure and reasoning strategy between Claude-3.7-Sonnet and Claude-3.7-Sonnet-Thinking. Average textual similarity between model responses of these two models on \benchmark is 0.9. 

\subsection{Impact of CoT}

\begin{wraptable}{r}{7.5cm}
\vspace{-15mm}
{
\begin{tabular}{l|cc}
\toprule
Model & Direct & CoT\\
\midrule
GPT-4o & 34.0 & 41.6\\ 
Gemini-1.5-Pro & 41.0 & 45.1\\ 
Claude-3.5-Sonnet & 40.0 & 42.5\\ 
Qwen2-VL-2B-Instruct & 31.3 & 26.1\\ 
Qwen2.5-VL-7B-Instruct & 33.7 & 32.0 \\ 
Cambrian-13B & 27.4 & 26.5\\ 
LLaVA-NeXT-34B & 29.8 & 29.6 \\ 
Qwen2.5-VL-72B-Instruct & 38.6 & 42.3 \\ 
LLama-3.2-90B-Vision-Instruct & 33.3 & 33.9 \\
\bottomrule
\end{tabular}}
\vspace{-3mm}
\caption{Comparison of models with Direct Multiple Choice and CoT prompting.}
\label{tab:comparison_prompt}
\vspace{-6mm}
\end{wraptable}

Table \ref{tab:comparison_prompt} compares model performance under two prompting strategies: direct multiple-choice prompt vs. Chain-of-Thought (CoT) prompt. 
We observe that proprietary models and larger open models $(\geq$72B$)$ benefit from CoT, while others show little to no improvement or even a decline in performance with CoT. 
For instance, both GPT-4o and Qwen2.5-VL-72B-Instruct show more than 20\% increases in performance when using CoT. In contrast, several smaller models, such as Qwen2-VL-2B-Instruct and Cambrian-13B, exhibit decreased accuracy with CoT prompting. These results suggest that CoT can indeed enhance the reasoning capability of larger models whereas it may introduce unnecessary complexity or confusion for smaller models and thus decreasing performance.

\newpage
\section{Case Study}
\label{app:case_study}
\begin{figure*}[!h]
    \centering
    \includegraphics[width=\linewidth]{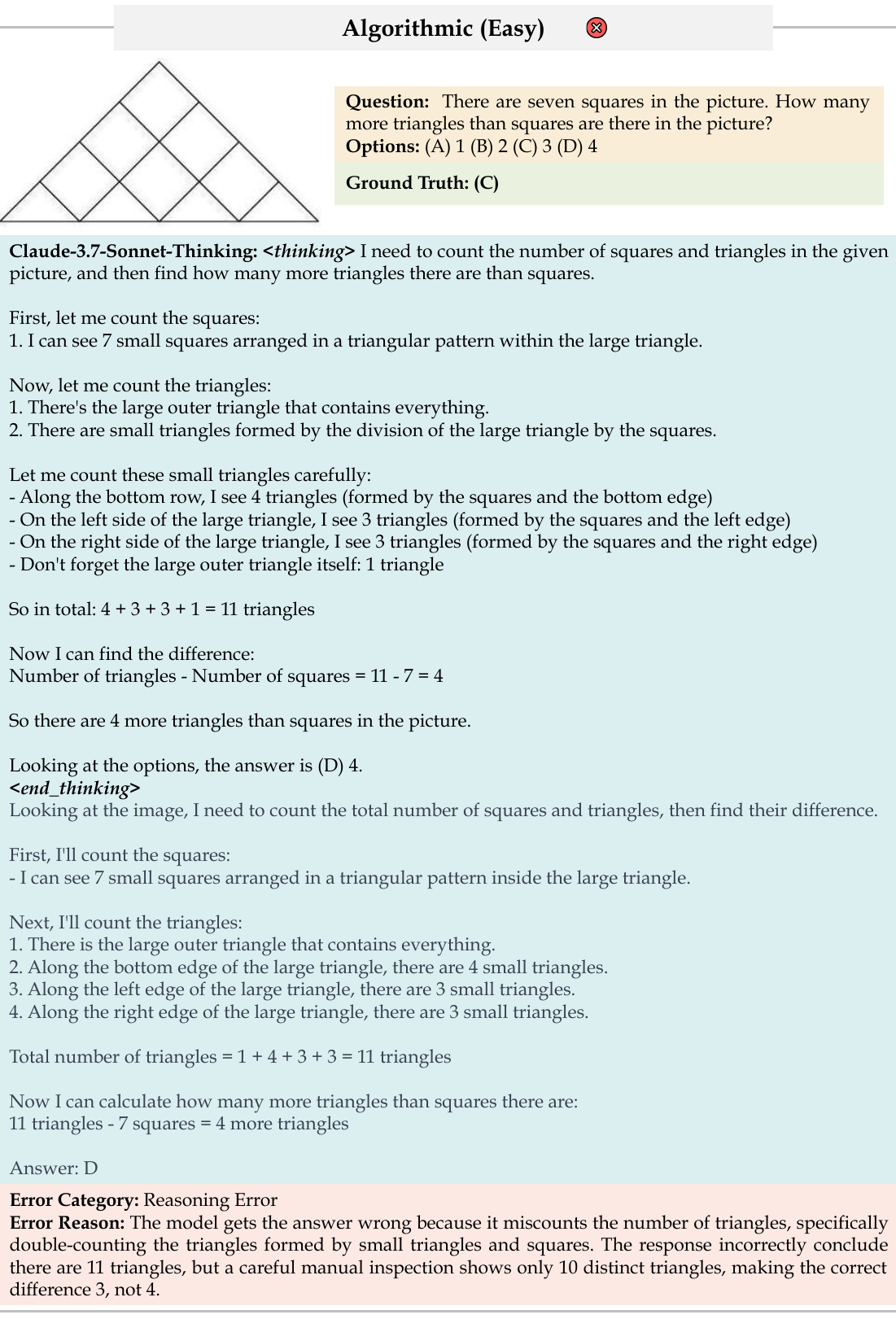}
    \caption{A sample error case of Algorithmic Reasoning (difficulty: Easy).} 
    \label{fig:algorithmic_easy_incorrect}
\end{figure*}

\begin{figure*}[!h]
    \centering
    \includegraphics[width=\linewidth]{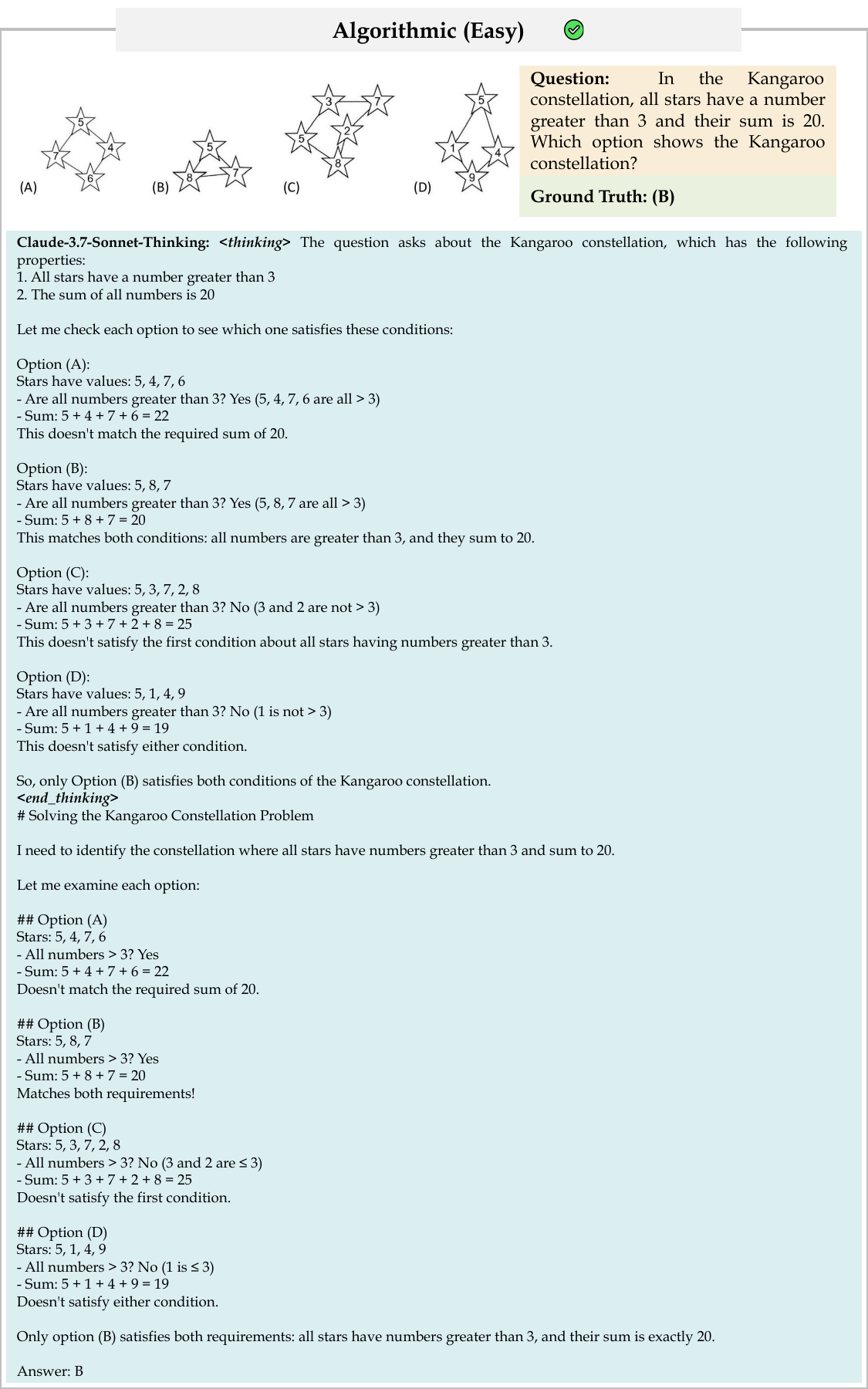}
    \caption{A sample correct case of Algorithmic Reasoning (difficulty: Easy).} 
    \label{fig:algorithmic_easy_correct}
\end{figure*}

\begin{figure*}[!h]
    \centering
    \includegraphics[width=\linewidth]{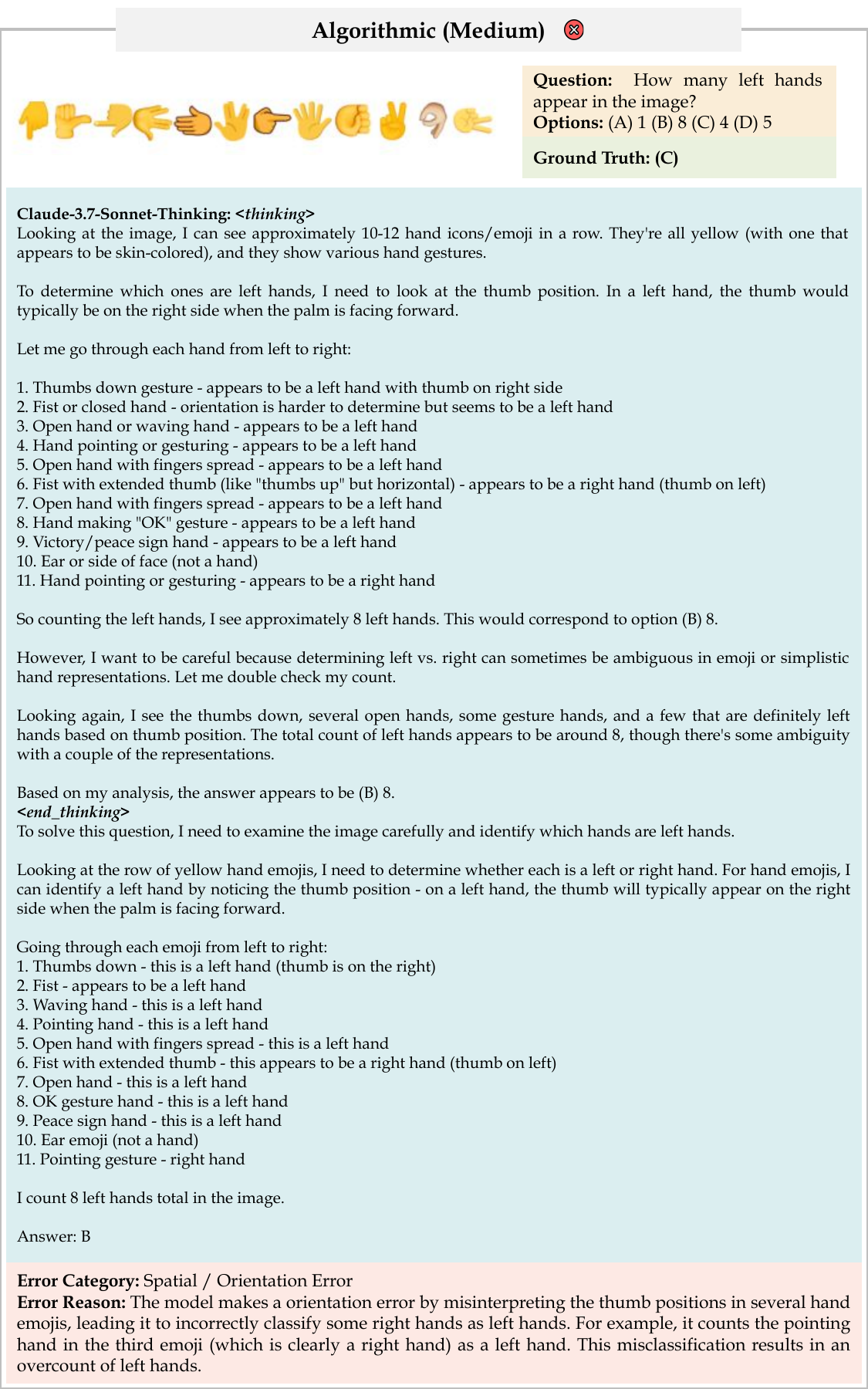}
    \caption{A sample error case of Algorithmic Reasoning (difficulty: Medium).} 
    \label{fig:algorithmic_medium_incorrect}
\end{figure*}

\begin{figure*}[!h]
    \centering
    \includegraphics[width=\linewidth]{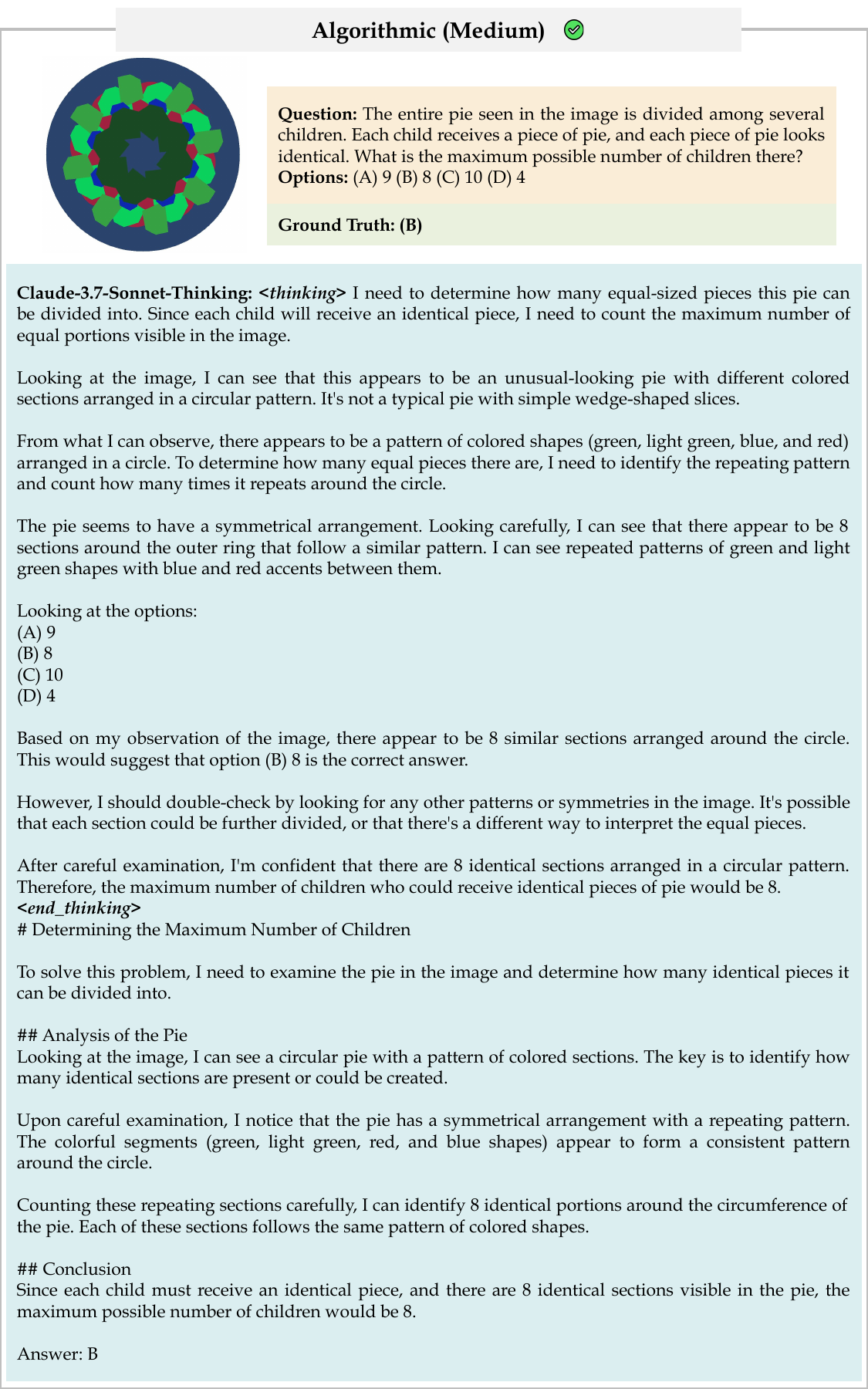}
    \caption{A sample correct case of Algorithmic Reasoning (difficulty: Medium).} 
    \label{fig:algorithmic_medium_correct}
\end{figure*}

\begin{figure*}[!h]
    \centering
    \includegraphics[width=\linewidth]{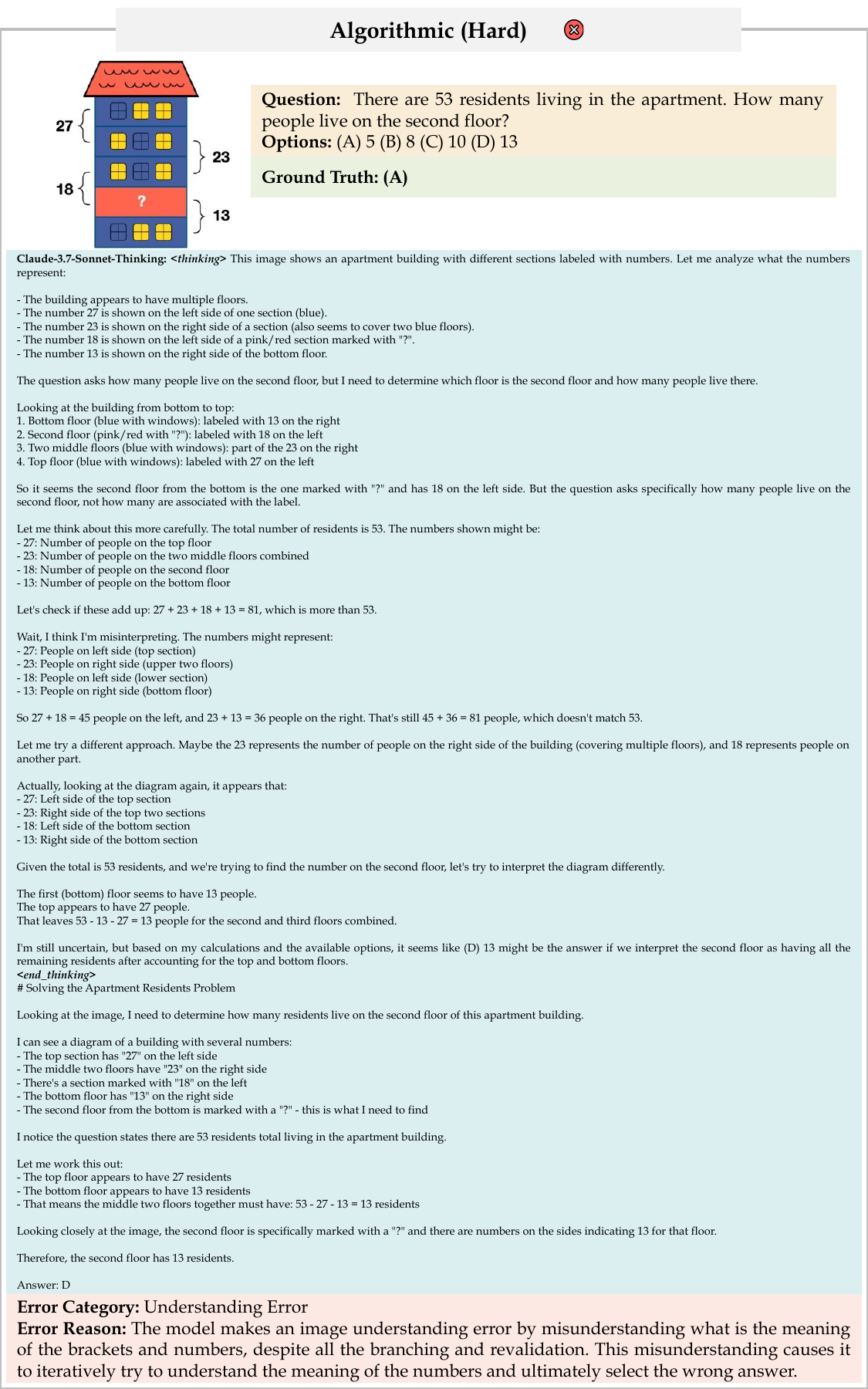}
    \caption{A sample error case of Algorithmic Reasoning (difficulty: Hard).} 
    \label{fig:algorithmic_hard_incorrect}
\end{figure*}

\begin{figure*}[!h]
    \centering
    \includegraphics[width=\linewidth]{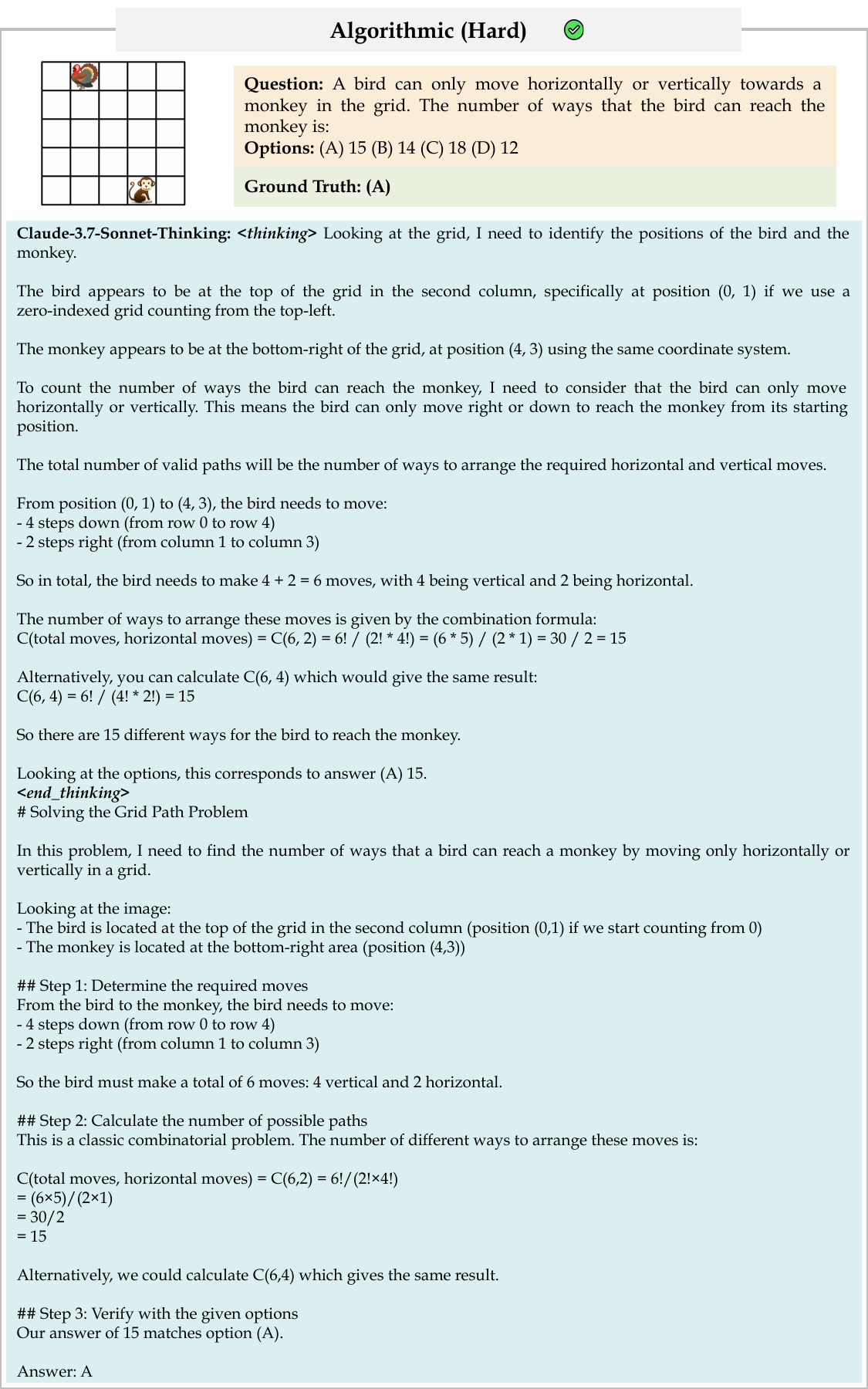}
    \caption{A sample correct case of Algorithmic Reasoning (difficulty: Hard).} 
    \label{fig:algorithmic_hard_correct}
\end{figure*}

\begin{figure*}[!h]
    \centering
    \includegraphics[width=\linewidth]{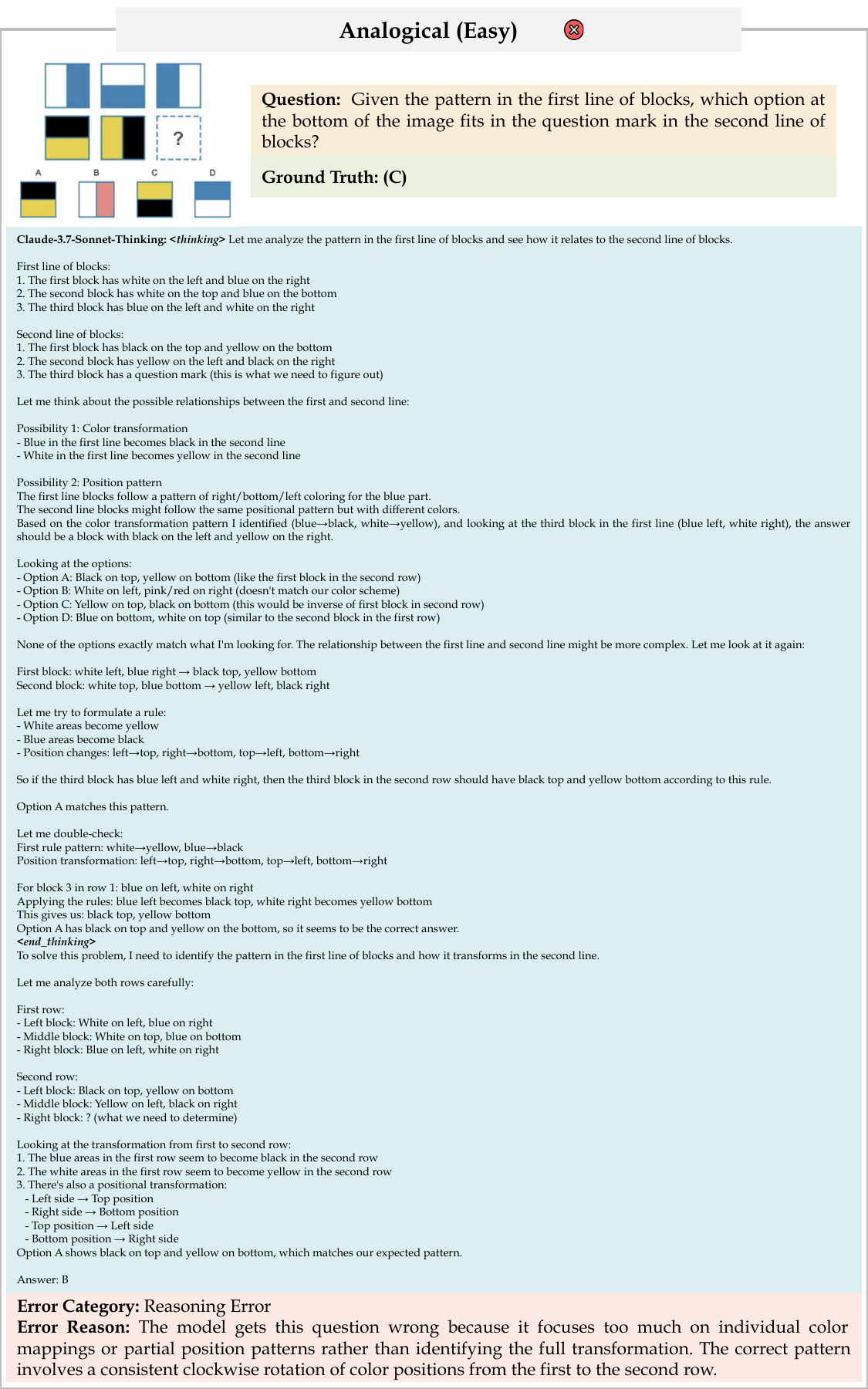}
    \caption{A sample error case of Analogical Reasoning (difficulty: Easy).} 
    \label{fig:analogical_easy_incorrect}
\end{figure*}

\begin{figure*}[!h]
    \centering
    \includegraphics[width=\linewidth]{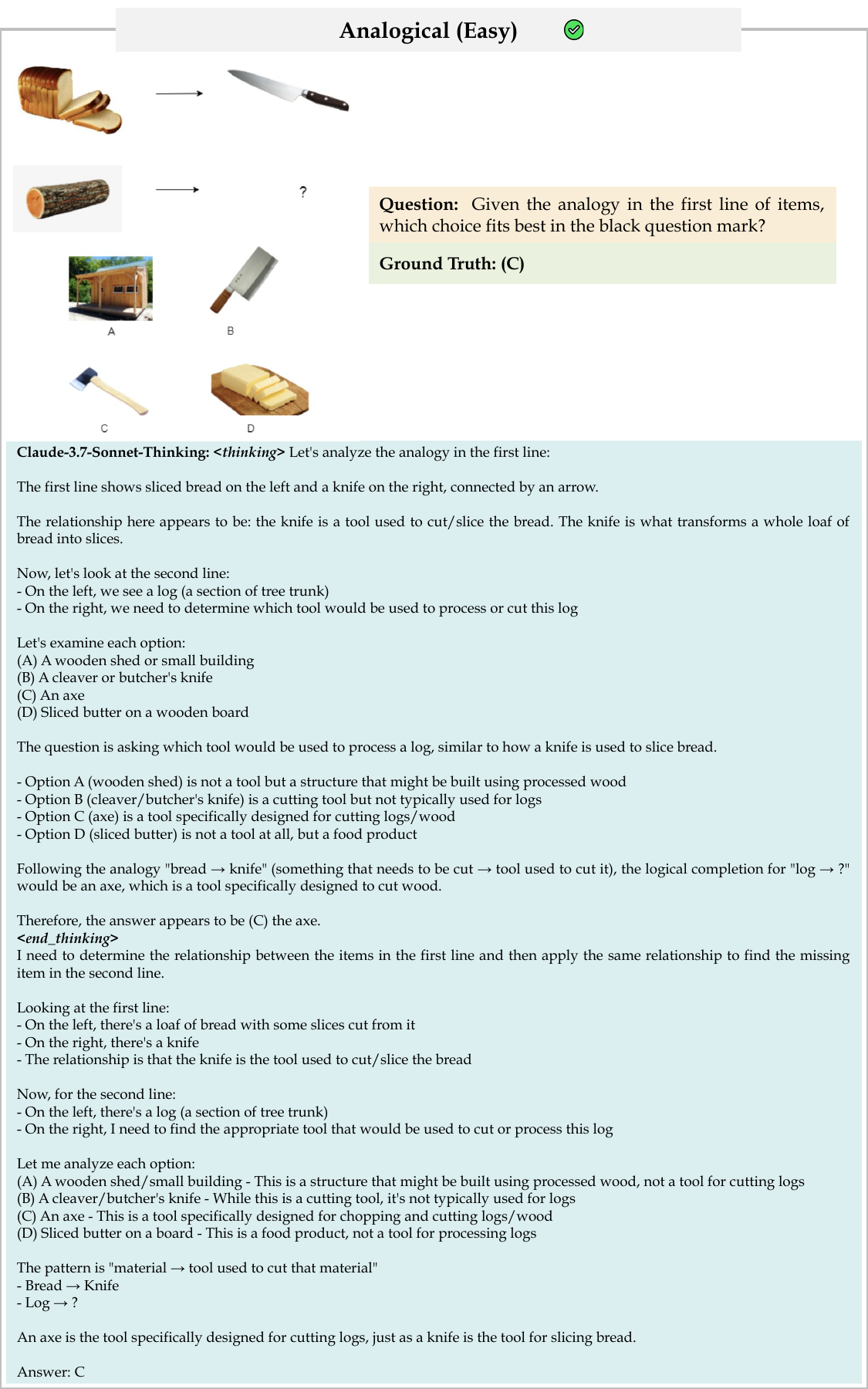}
    \caption{A sample correct case of Analogical Reasoning (difficulty: Easy).} 
    \label{fig:analogical_easy_correct}
\end{figure*}

\begin{figure*}[!h]
    \centering
    \includegraphics[width=\linewidth]{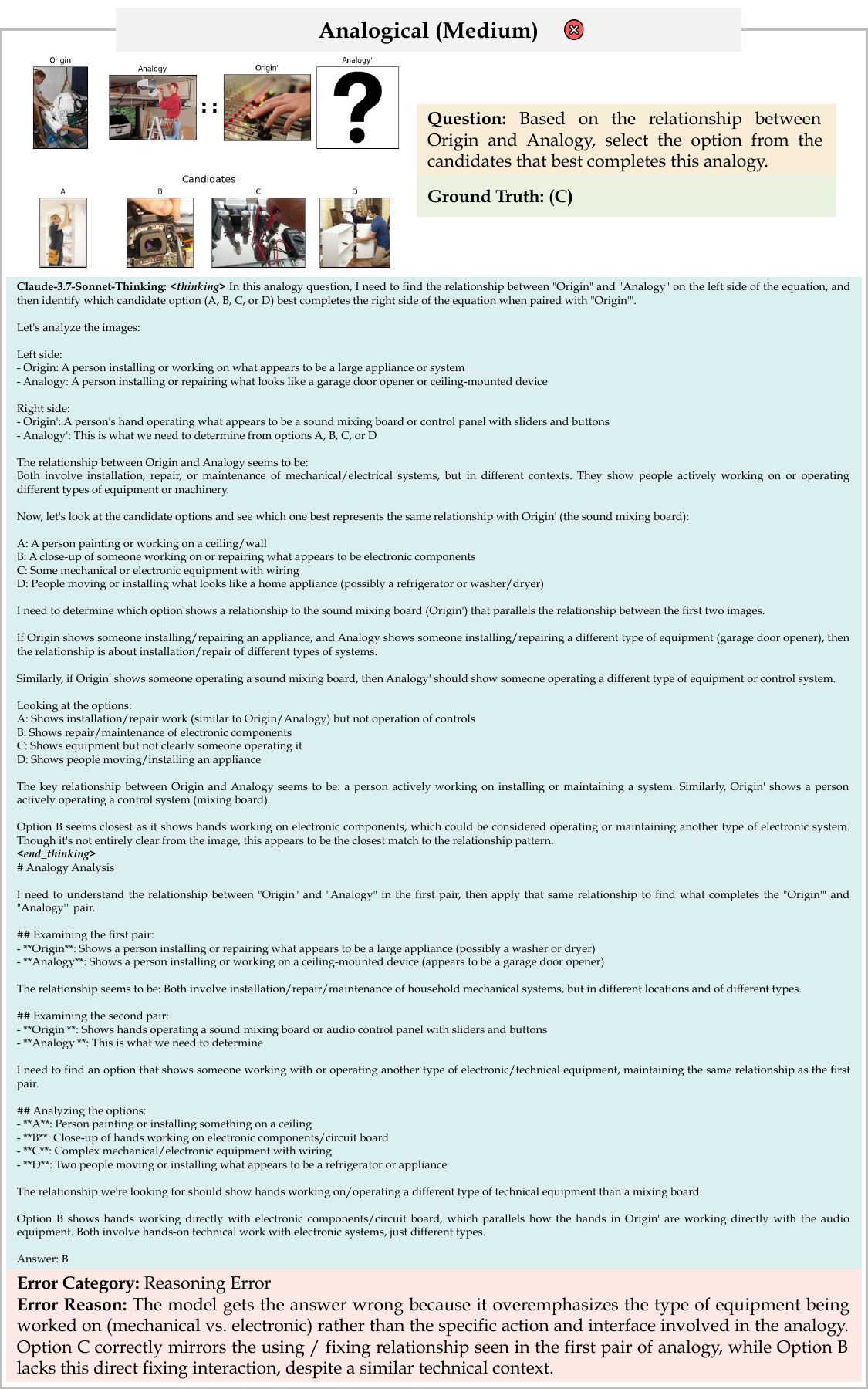}
    \caption{A sample error case of Analogical Reasoning (difficulty: Medium).} 
    \label{fig:analogical_medium_incorrect}
\end{figure*}

\begin{figure*}[!h]
    \centering
    \includegraphics[width=\linewidth]{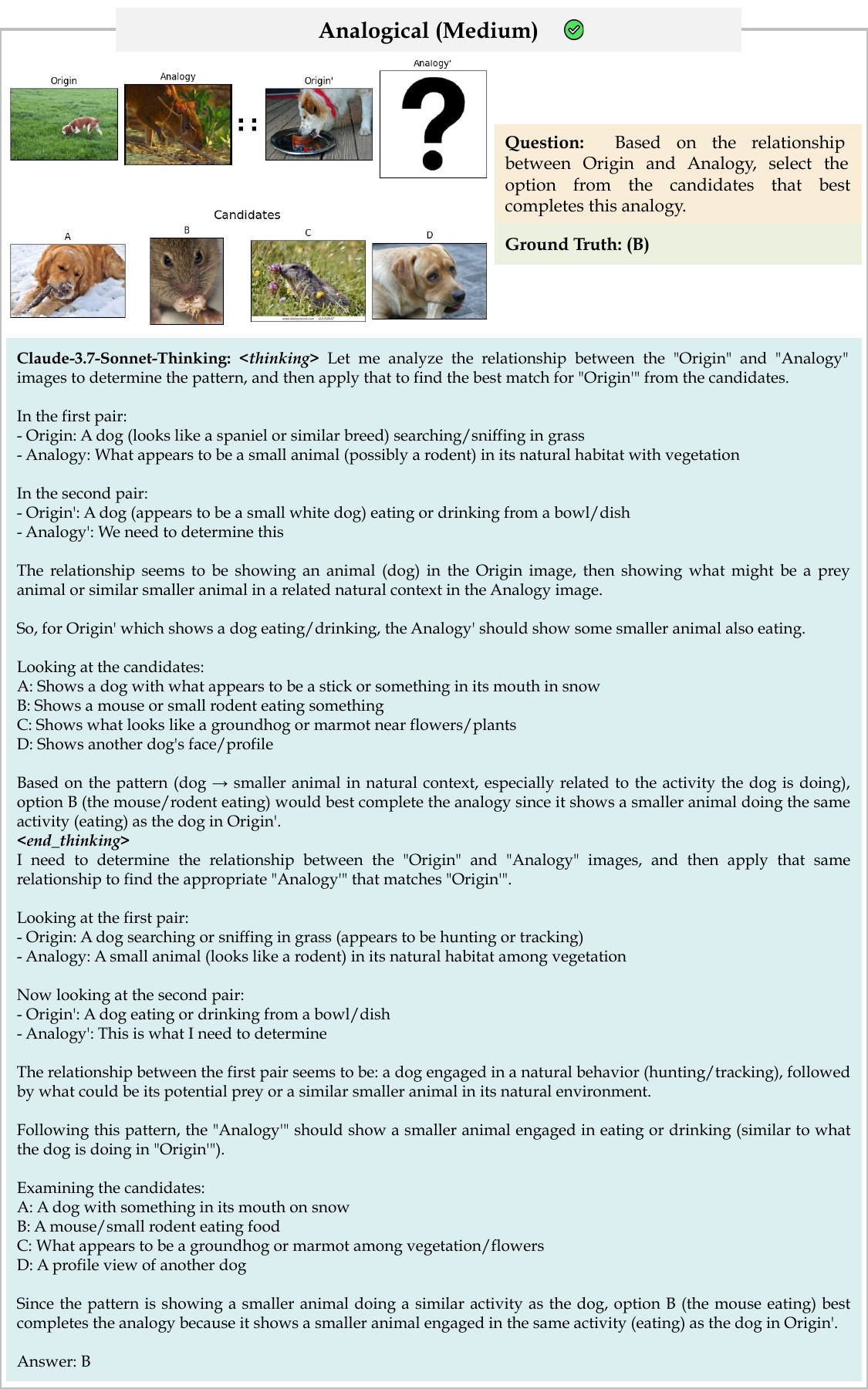}
    \caption{A sample correct case of Analogical Reasoning (difficulty: Medium).} 
    \label{fig:analogical_medium_correct}
\end{figure*}

\begin{figure*}[!h]
    \centering
    \includegraphics[width=\linewidth]{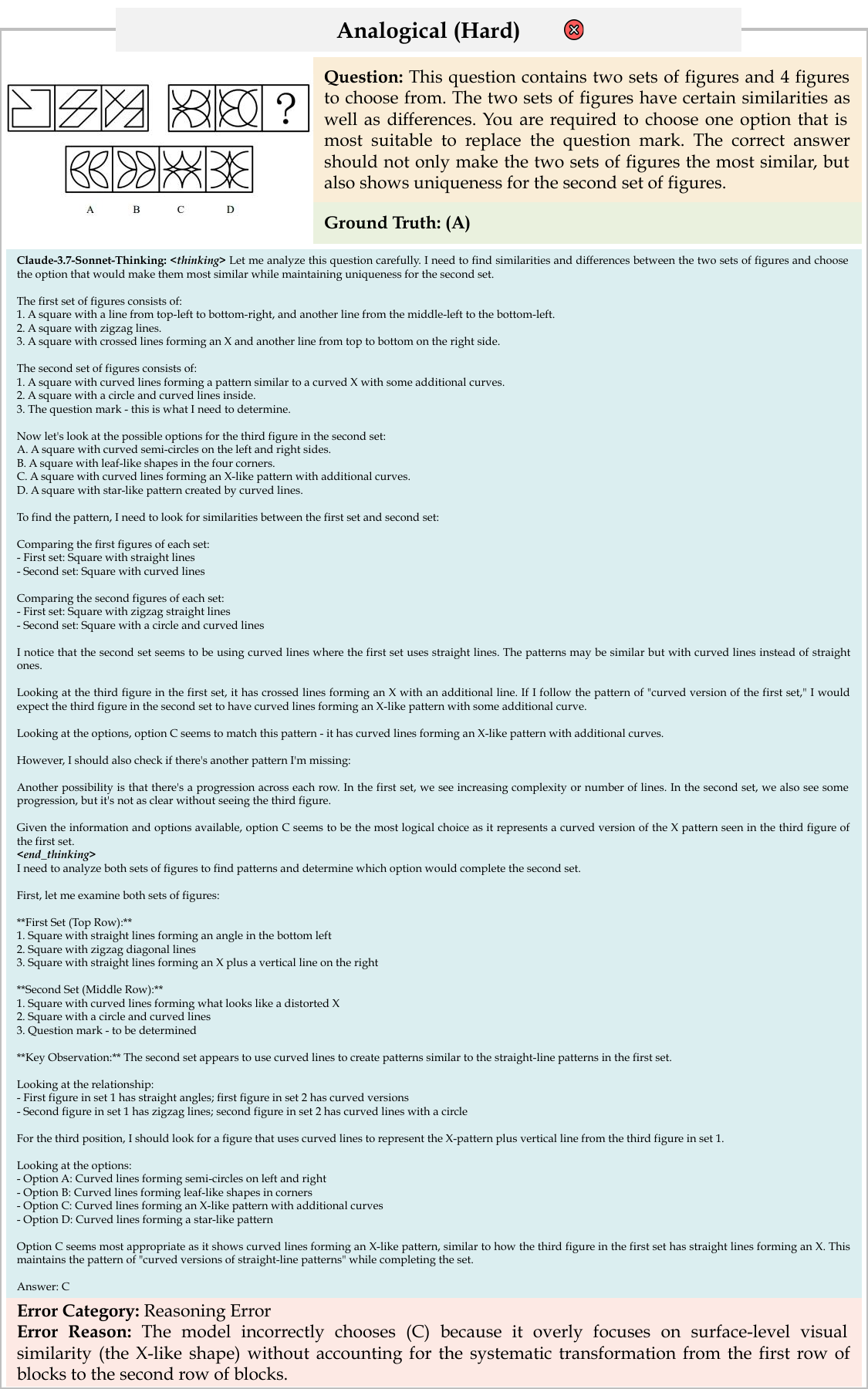}
    \caption{A sample error case of Analogical Reasoning (difficulty: Hard).} 
    \label{fig:analogical_hard_incorrect}
\end{figure*}

\begin{figure*}[!h]
    \centering
    \includegraphics[width=\linewidth]{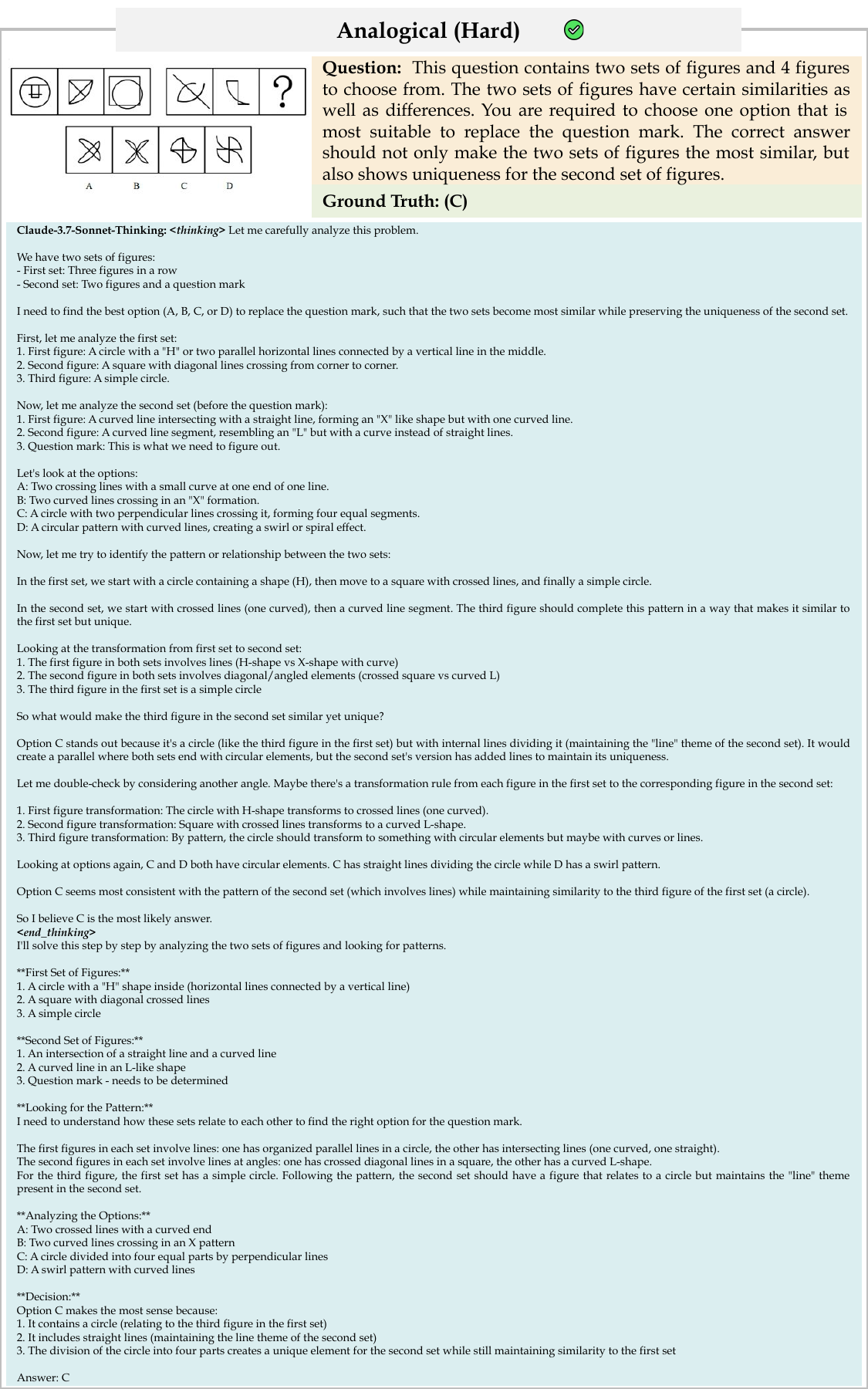}
    \caption{A sample correct case of Analogical Reasoning (difficulty: Hard).} 
    \label{fig:analogical_hard_correct}
\end{figure*}

\begin{figure*}[!h]
    \centering
    \includegraphics[width=\linewidth]{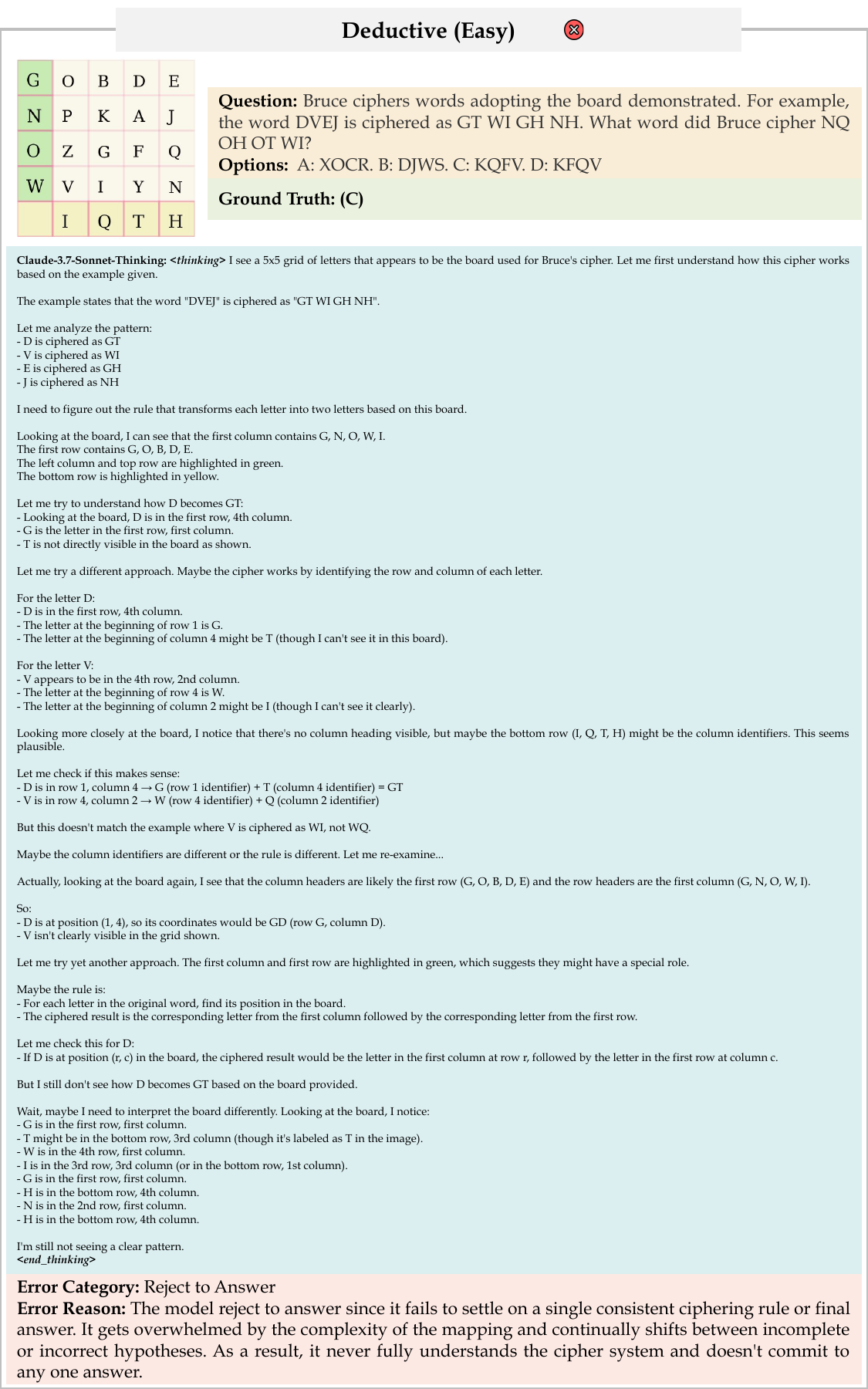}
    \caption{A sample error case of Deductive Reasoning (difficulty: Easy).} 
    \label{fig:deductive_easy_incorrect}
\end{figure*}

\begin{figure*}[!h]
    \centering
    \includegraphics[width=\linewidth]{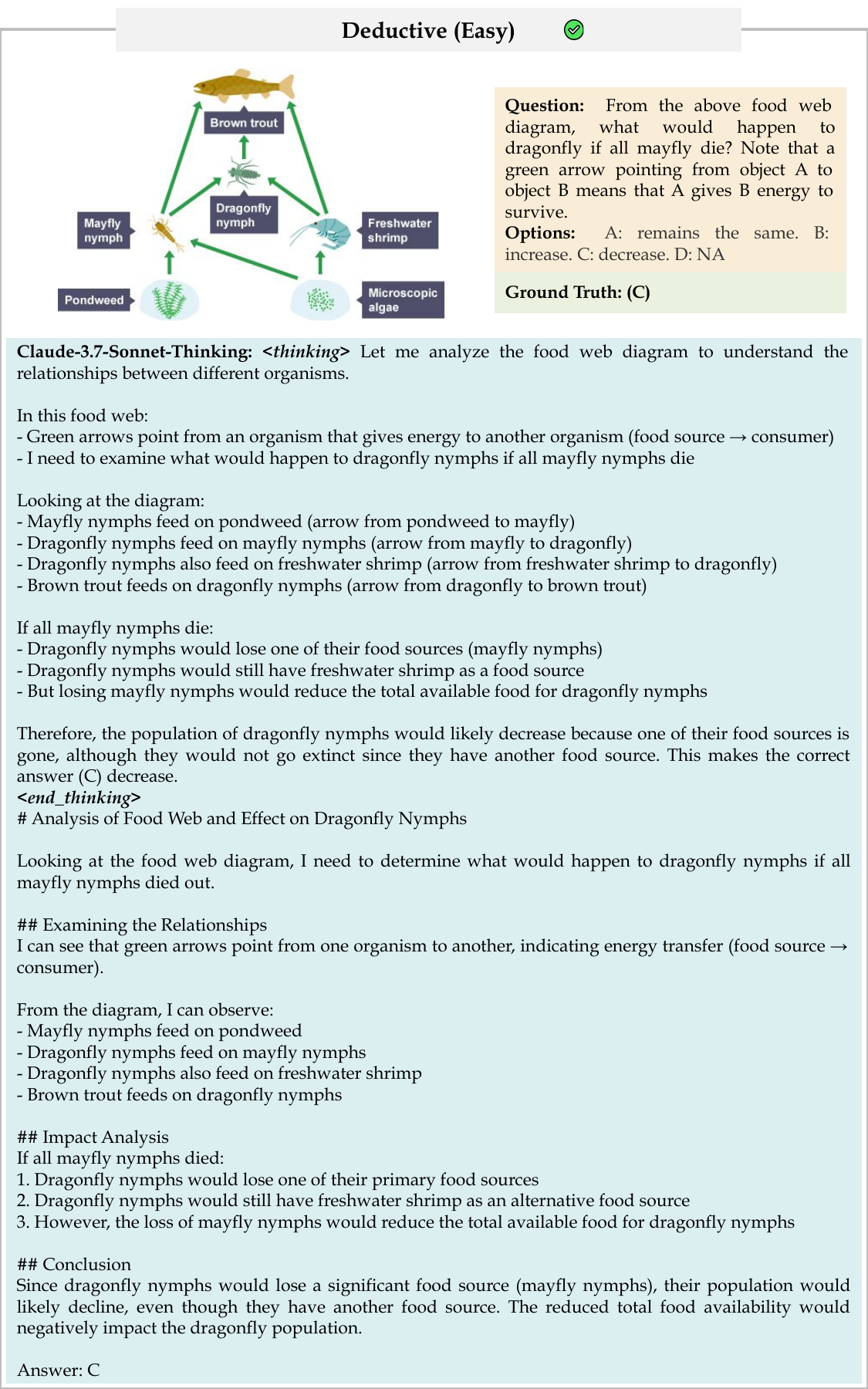}
    \caption{A sample correct case of Deductive Reasoning (difficulty: Easy).} 
    \label{fig:deductive_easy_correct}
\end{figure*}

\begin{figure*}[!h]
    \centering
    \includegraphics[width=\linewidth]{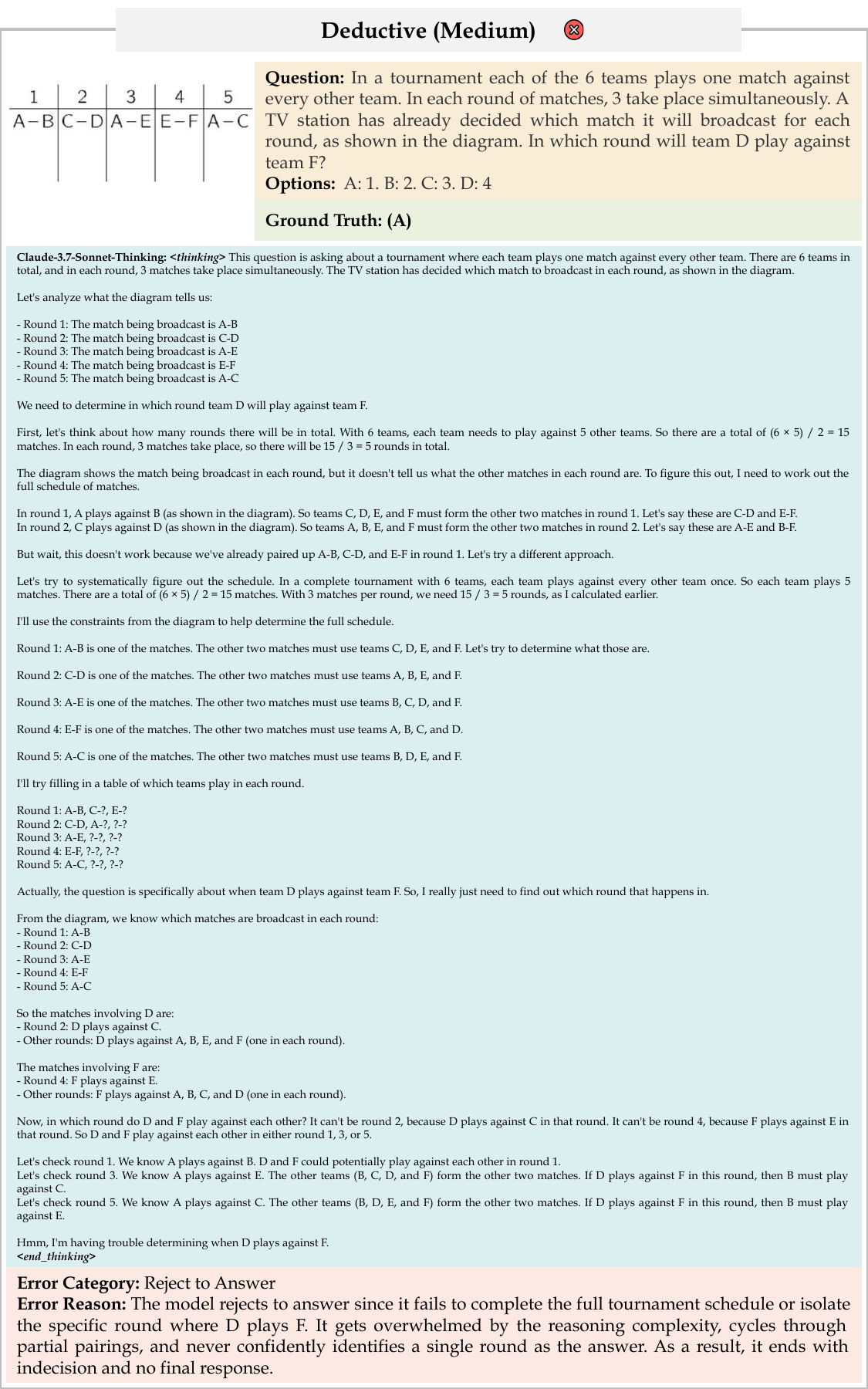}
    \caption{A sample error case of Deductive Reasoning (difficulty: Medium).} 
    \label{fig:deductive_medium_incorrect}
\end{figure*}

\begin{figure*}[!h]
    \centering
    \includegraphics[width=\linewidth]{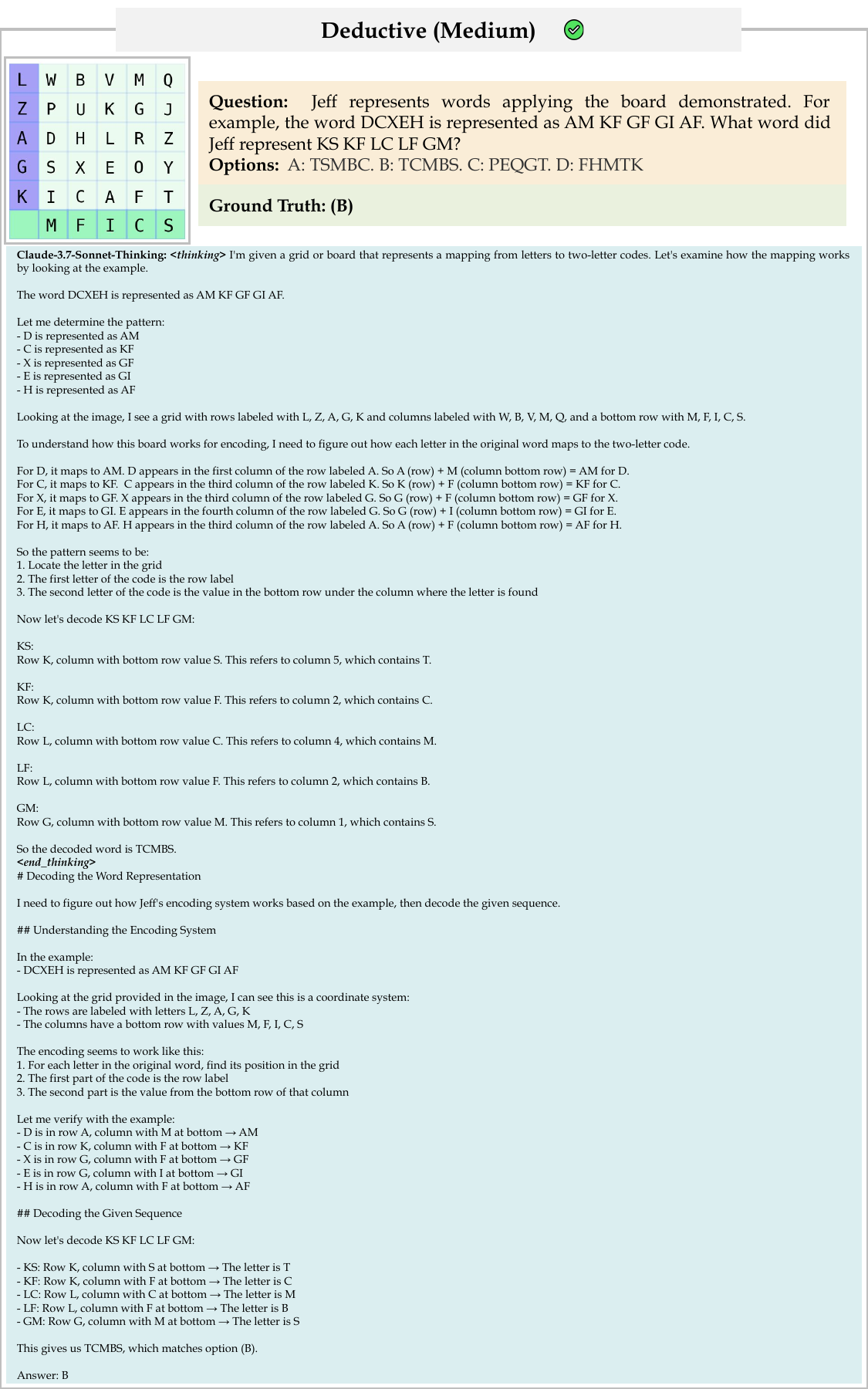}
    \caption{A sample correct case of Deductive Reasoning (difficulty: Medium).} 
    \label{fig:deductive_medium_correct}
\end{figure*}

\begin{figure*}[!h]
    \centering
    \includegraphics[width=\linewidth]{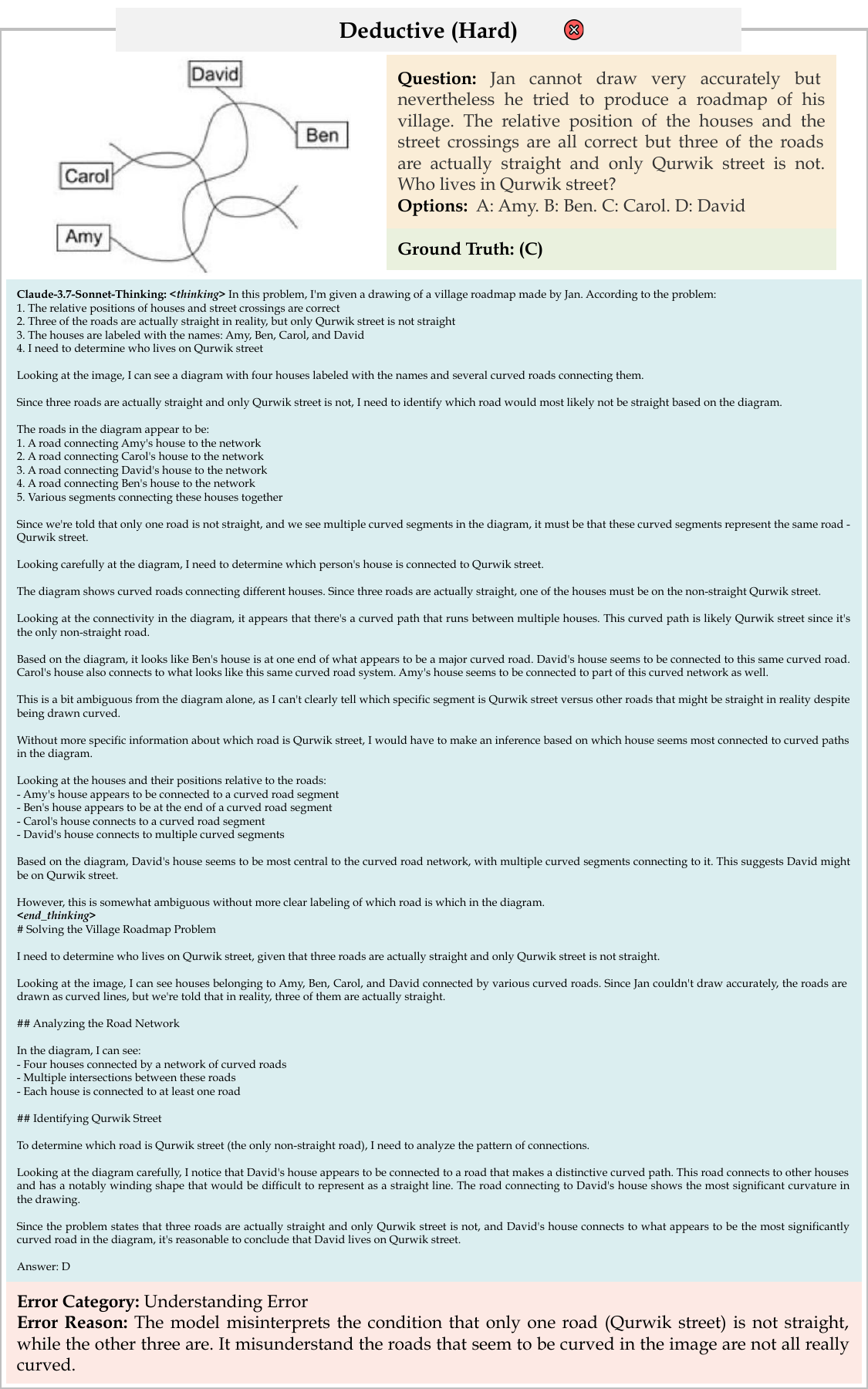}
    \caption{A sample error case of Deductive Reasoning (difficulty: Hard).} 
    \label{fig:deductive_hard_incorrect}
\end{figure*}

\begin{figure*}[!h]
    \centering
    \includegraphics[width=\linewidth]{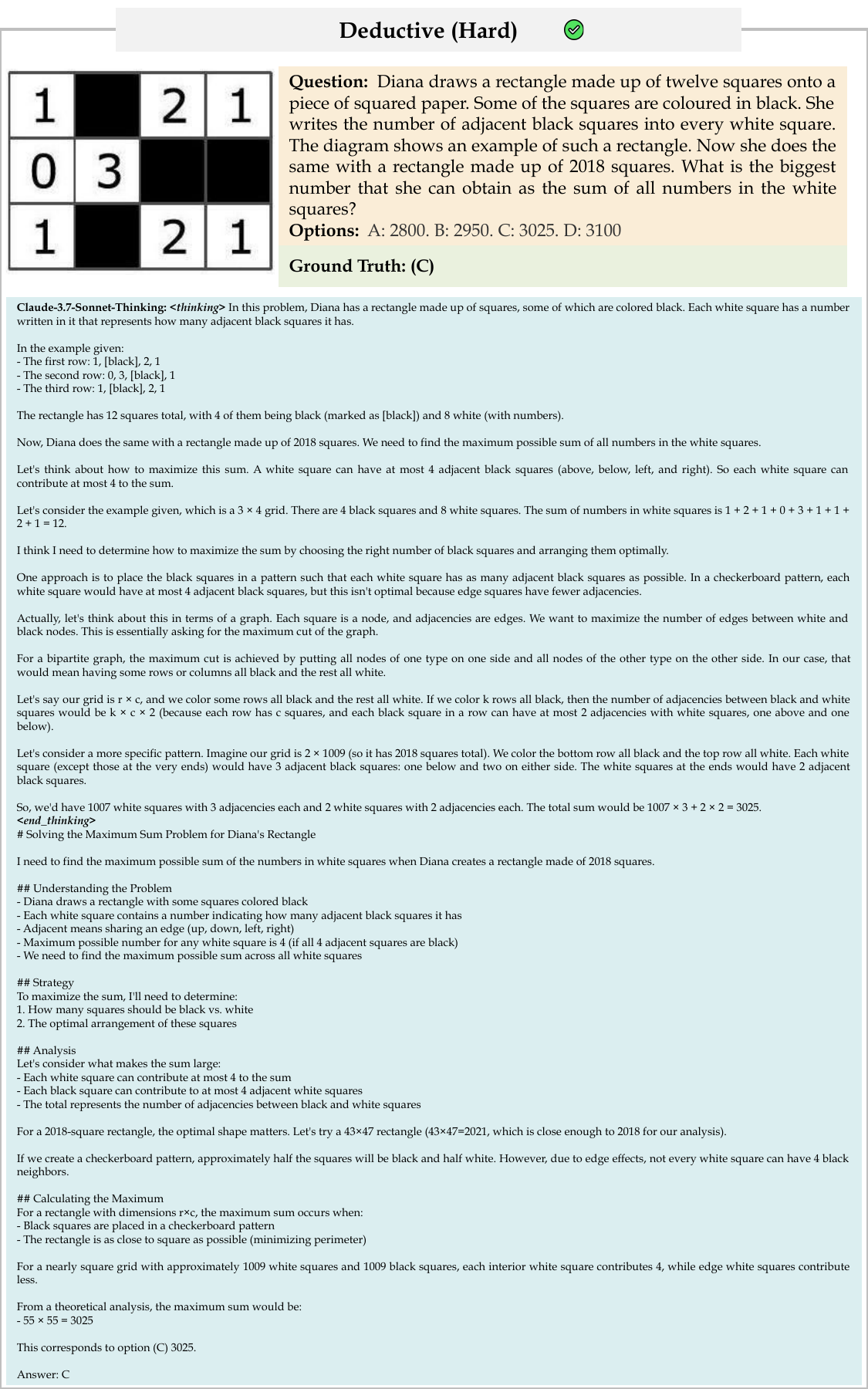}
    \caption{A sample correct case of Deductive Reasoning (difficulty: Hard).} 
    \label{fig:deductive_hard_correct}
\end{figure*}

\begin{figure*}[!h]
    \centering
    \includegraphics[width=\linewidth]{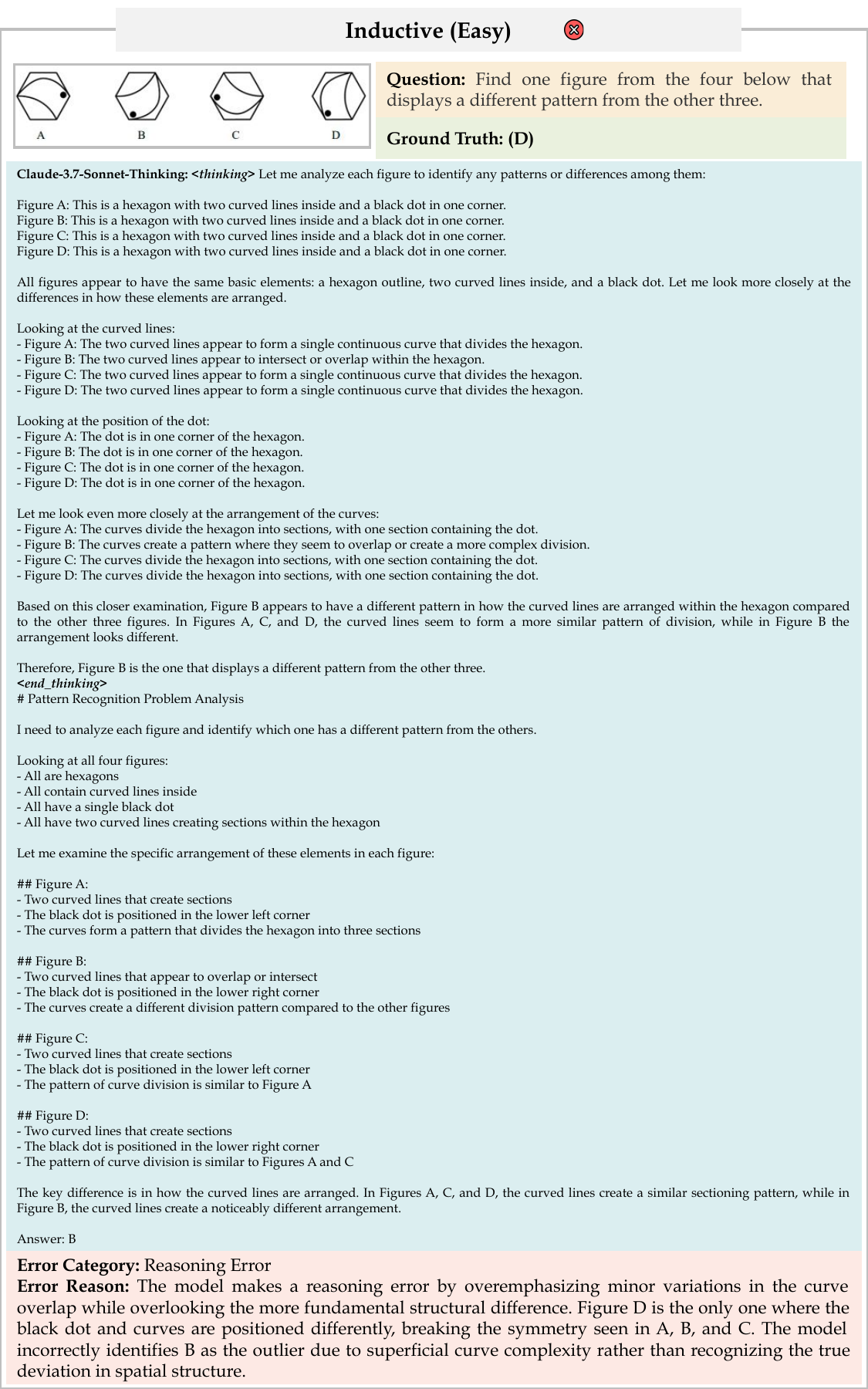}
    \caption{A sample error case of Inductive Reasoning (difficulty: Easy).} 
    \label{fig:inductive_easy_incorrect}
\end{figure*}

\begin{figure*}[!h]
    \centering
    \includegraphics[width=\linewidth]{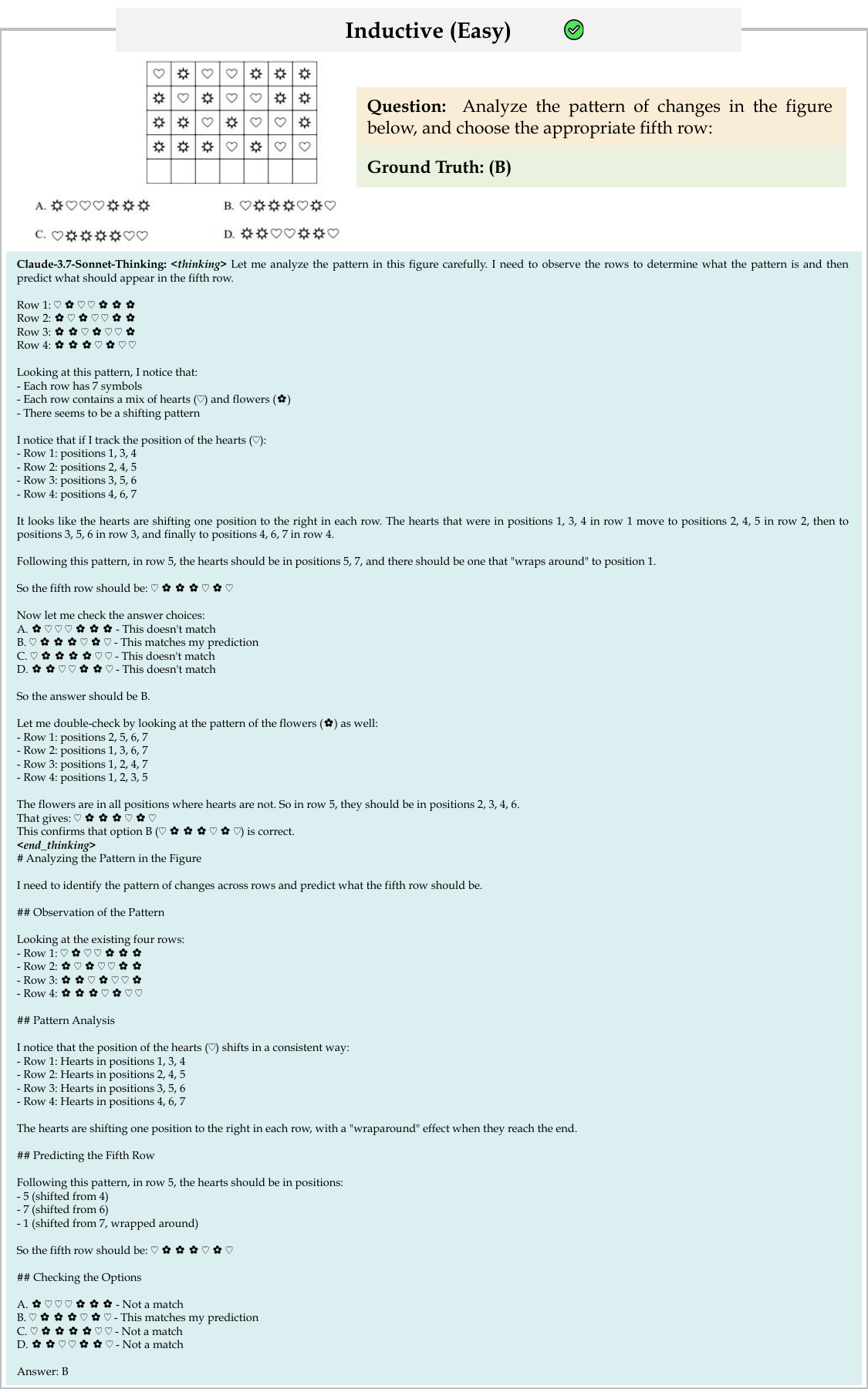}
    \caption{A sample correct case of Inductive Reasoning (difficulty: Easy).} 
    \label{fig:inductive_easy_correct}
\end{figure*}

\begin{figure*}[!h]
    \centering
    \includegraphics[width=\linewidth]{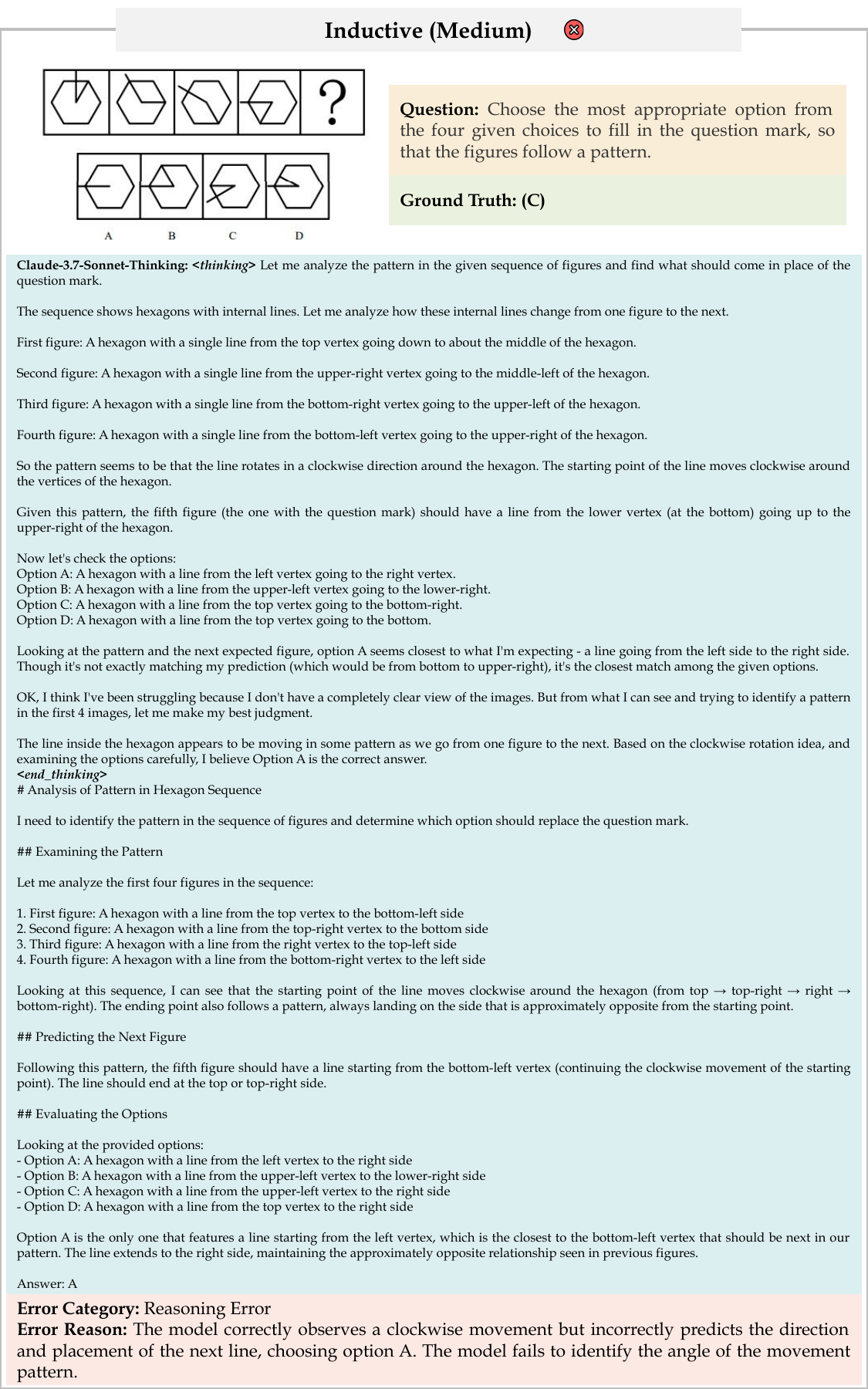}
    \caption{A sample error case of Inductive Reasoning (difficulty: Medium).} 
    \label{fig:inductive_medium_incorrect}
\end{figure*}

\begin{figure*}[!h]
    \centering
    \includegraphics[width=\linewidth]{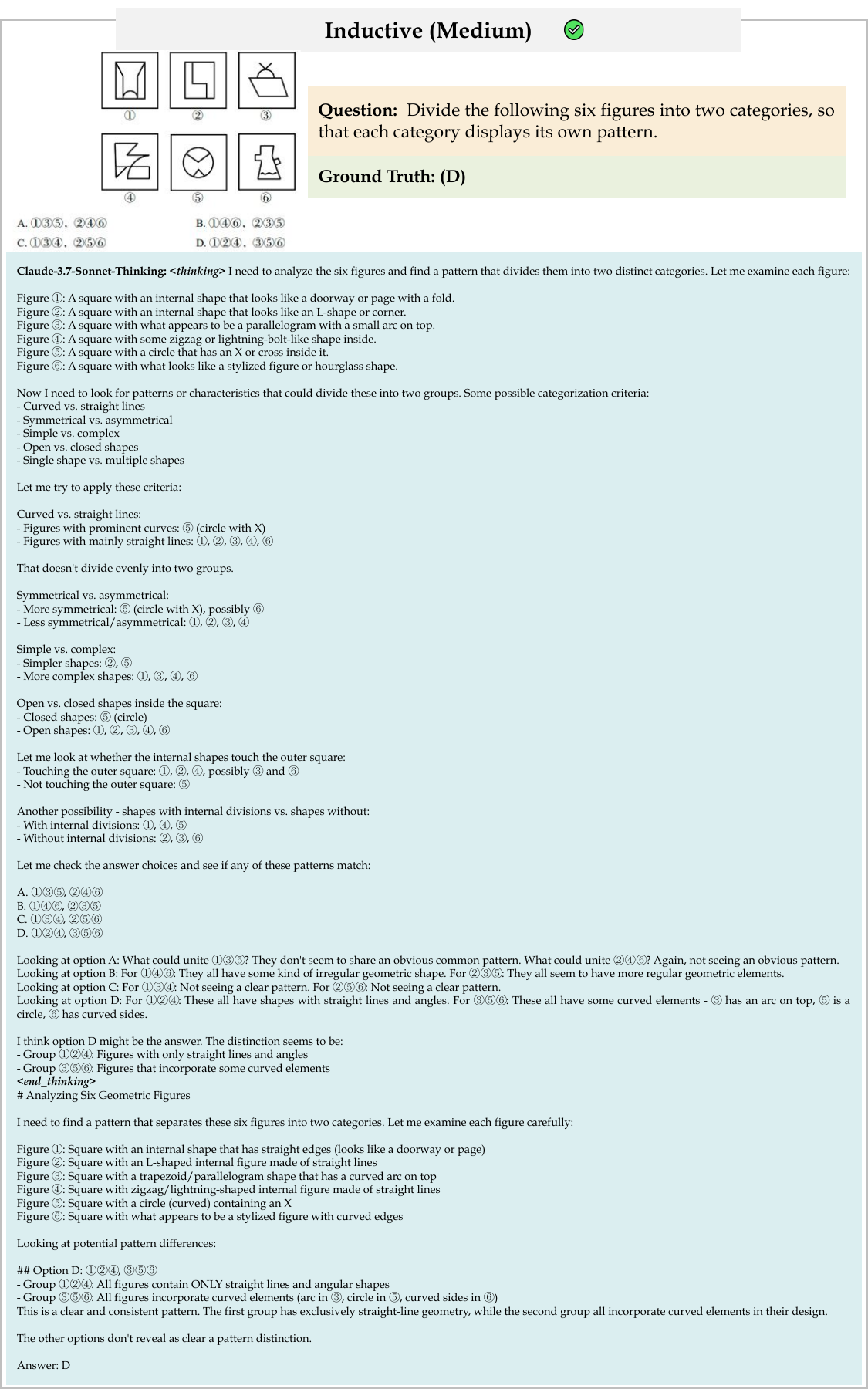}
    \caption{A sample correct case of Inductive Reasoning (difficulty: Medium).} 
    \label{fig:inductive_medium_correct}
\end{figure*}

\begin{figure*}[!h]
    \centering
    \includegraphics[width=\linewidth]{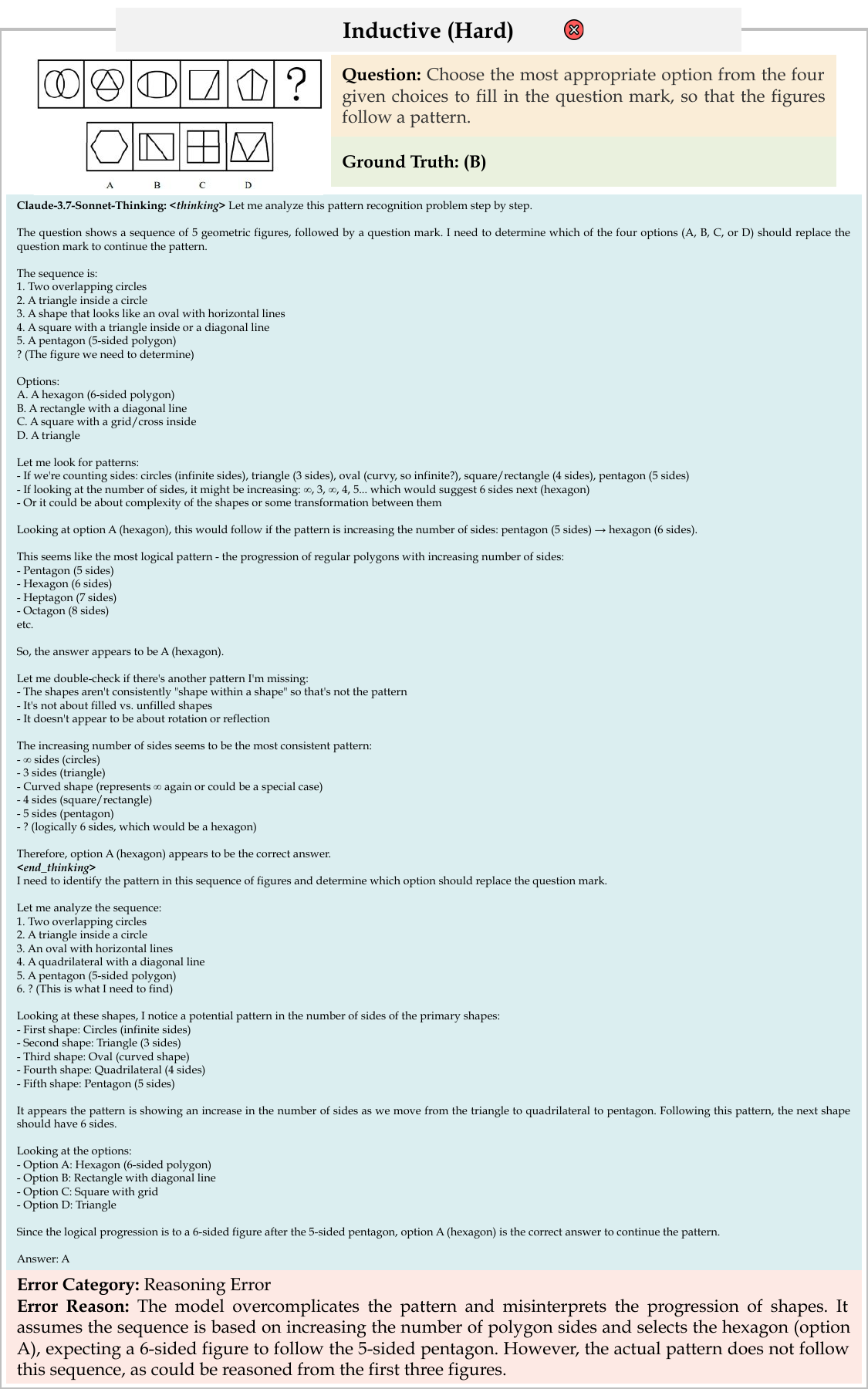}
    \caption{A sample error case of Inductive Reasoning (difficulty: Hard).} 
    \label{fig:inductive_hard_incorrect}
\end{figure*}

\begin{figure*}[!h]
    \centering
    \includegraphics[width=\linewidth]{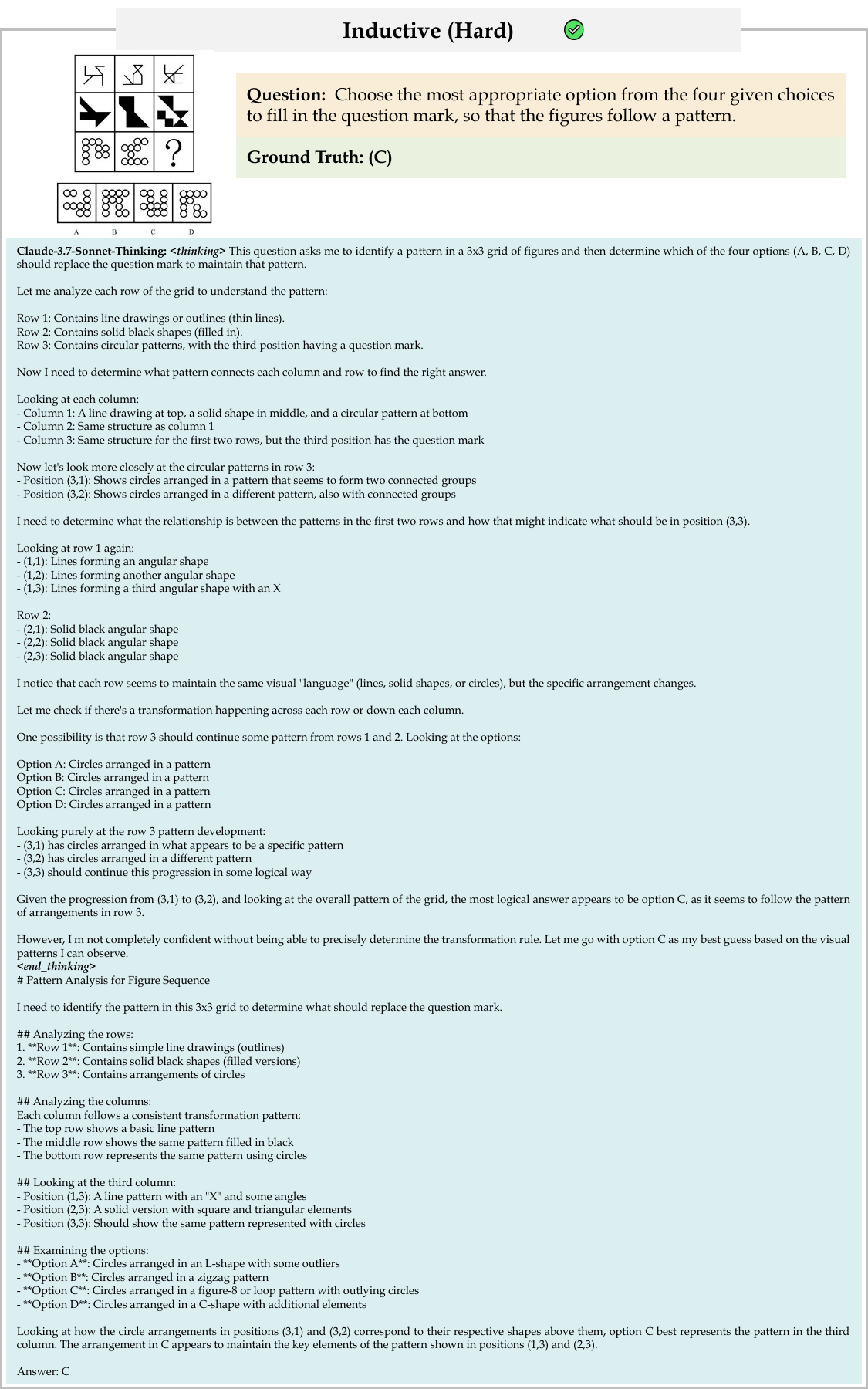}
    \caption{A sample correct case of Inductive Reasoning (difficulty: Hard).} 
    \label{fig:inductive_hard_correct}
\end{figure*}

\begin{figure*}[!h]
    \centering
    \includegraphics[width=\linewidth]{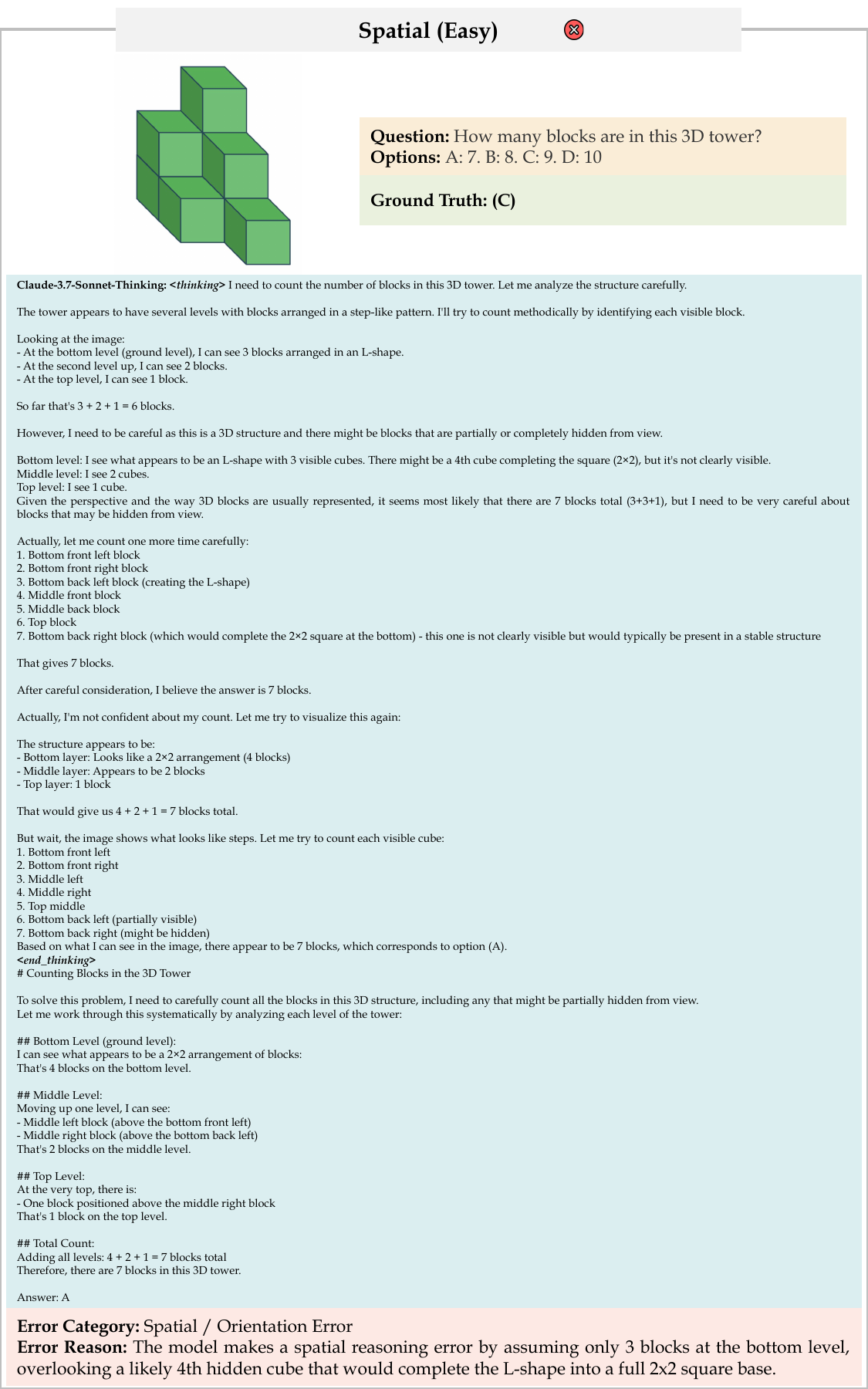}
    \caption{A sample error case of Spatial Reasoning (difficulty: Easy).} 
    \label{fig:spatial_easy_incorrect}
\end{figure*}

\begin{figure*}[!h]
    \centering
    \includegraphics[width=\linewidth]{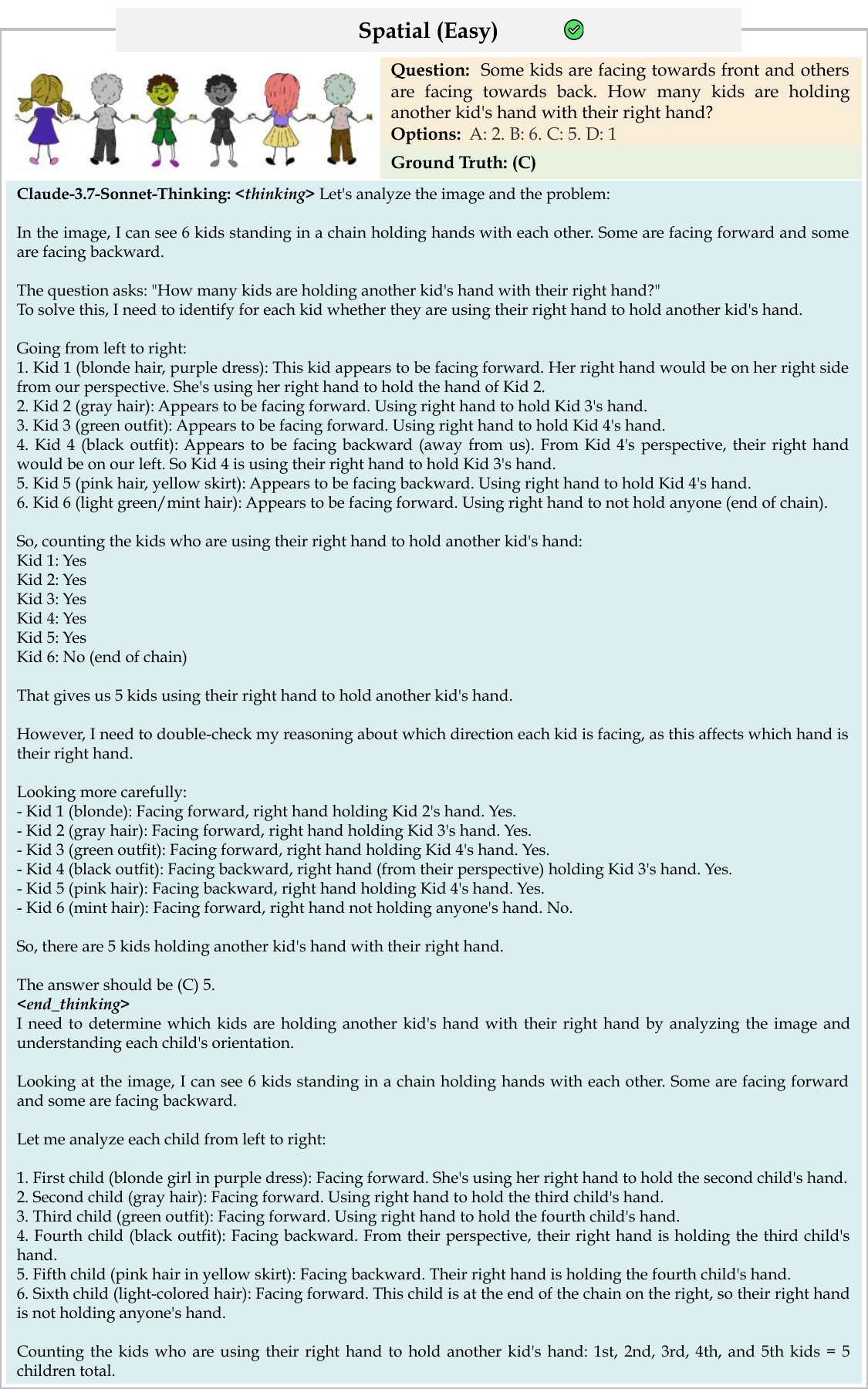}
    \caption{A sample correct case of Spatial Reasoning (difficulty: Easy).} 
    \label{fig:spatial_easy_correct}
\end{figure*}

\begin{figure*}[!h]
    \centering
    \includegraphics[width=\linewidth]{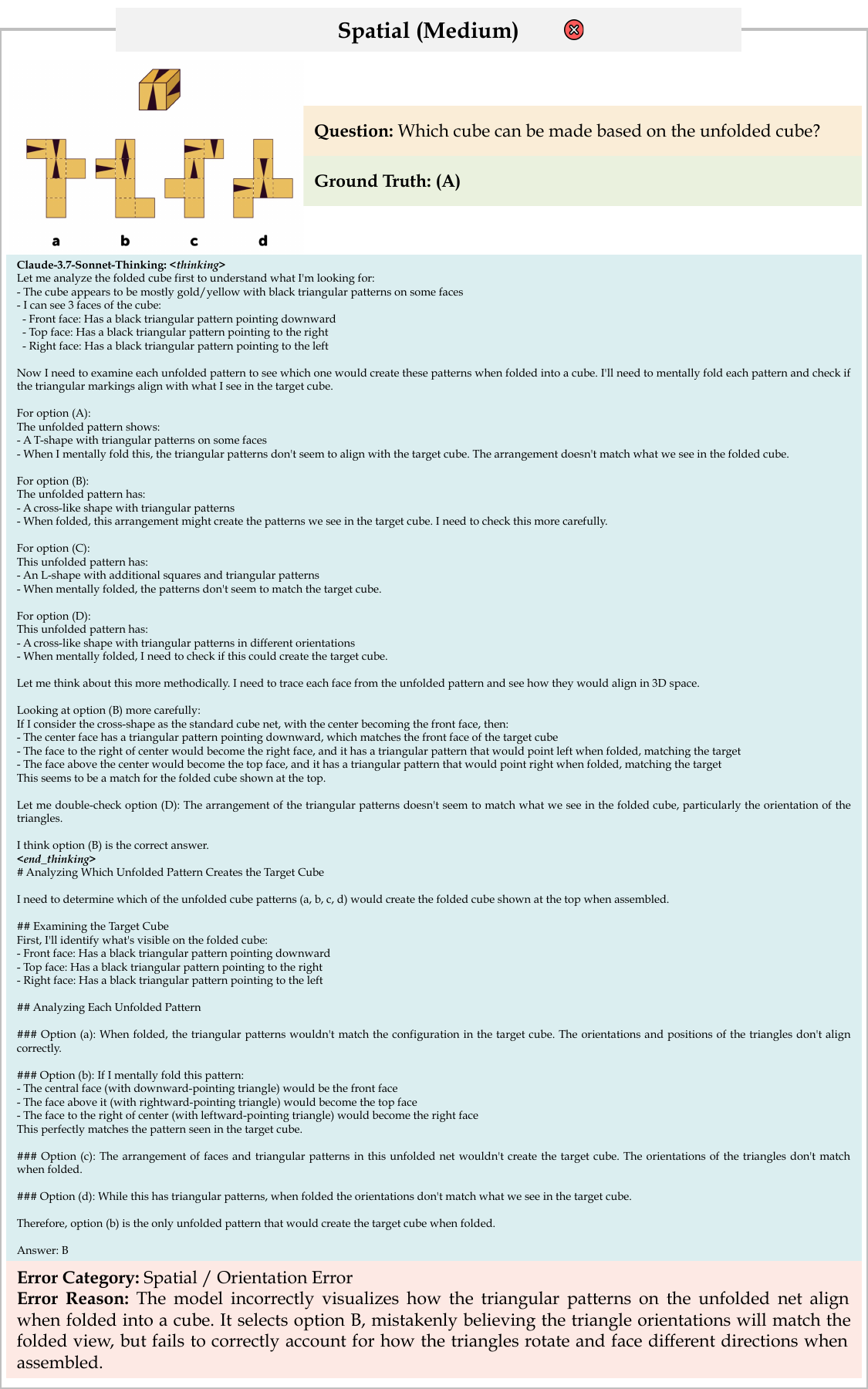}
    \caption{A sample error case of Spatial Reasoning (difficulty: Medium).} 
    \label{fig:spatial_medium_incorrect}
\end{figure*}

\begin{figure*}[!h]
    \centering
    \includegraphics[width=\linewidth]{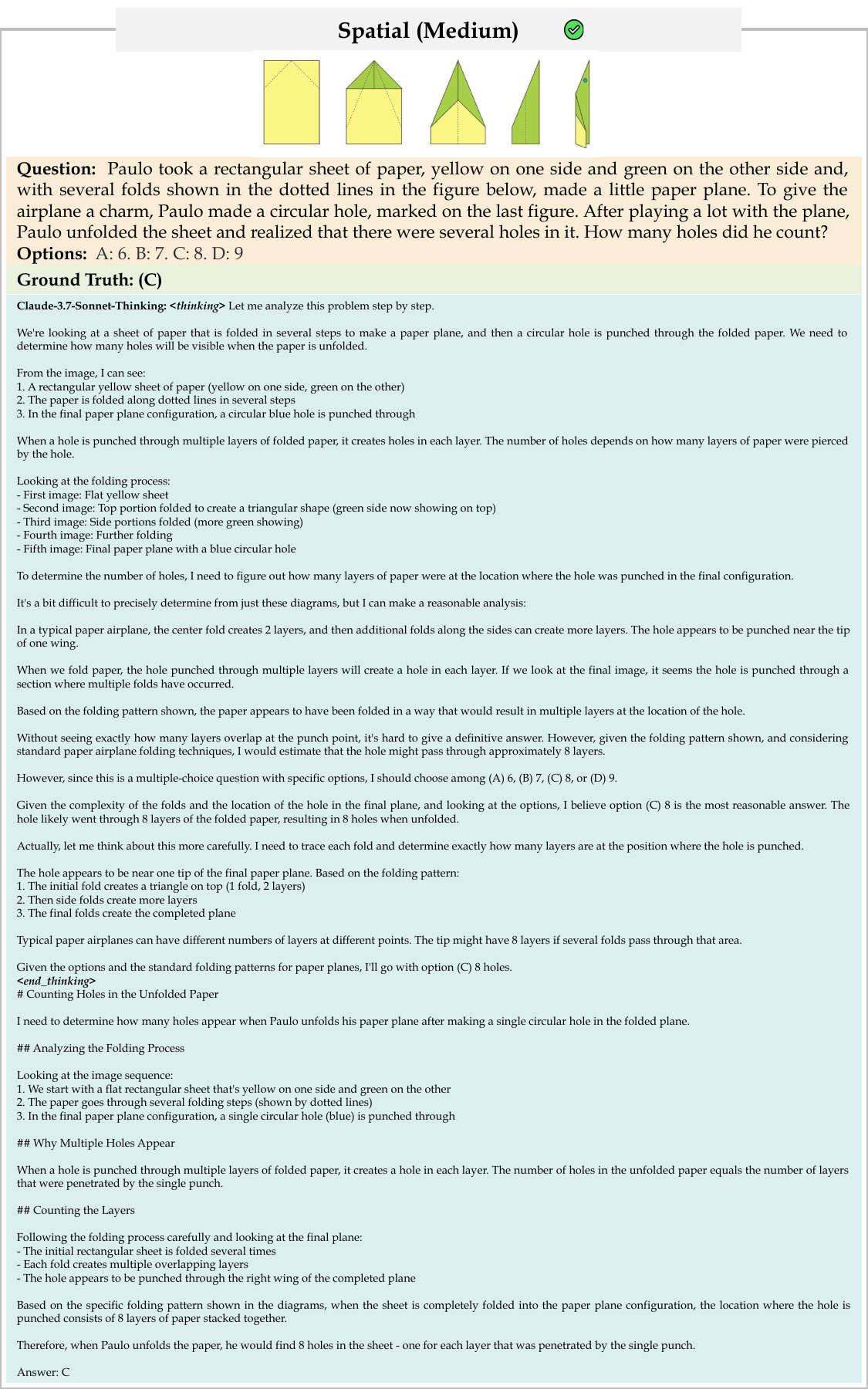}
    \caption{A sample correct case of Spatial Reasoning (difficulty: Medium).} 
    \label{fig:spatial_medium_correct}
\end{figure*}

\begin{figure*}[!h]
    \centering
    \includegraphics[width=\linewidth]{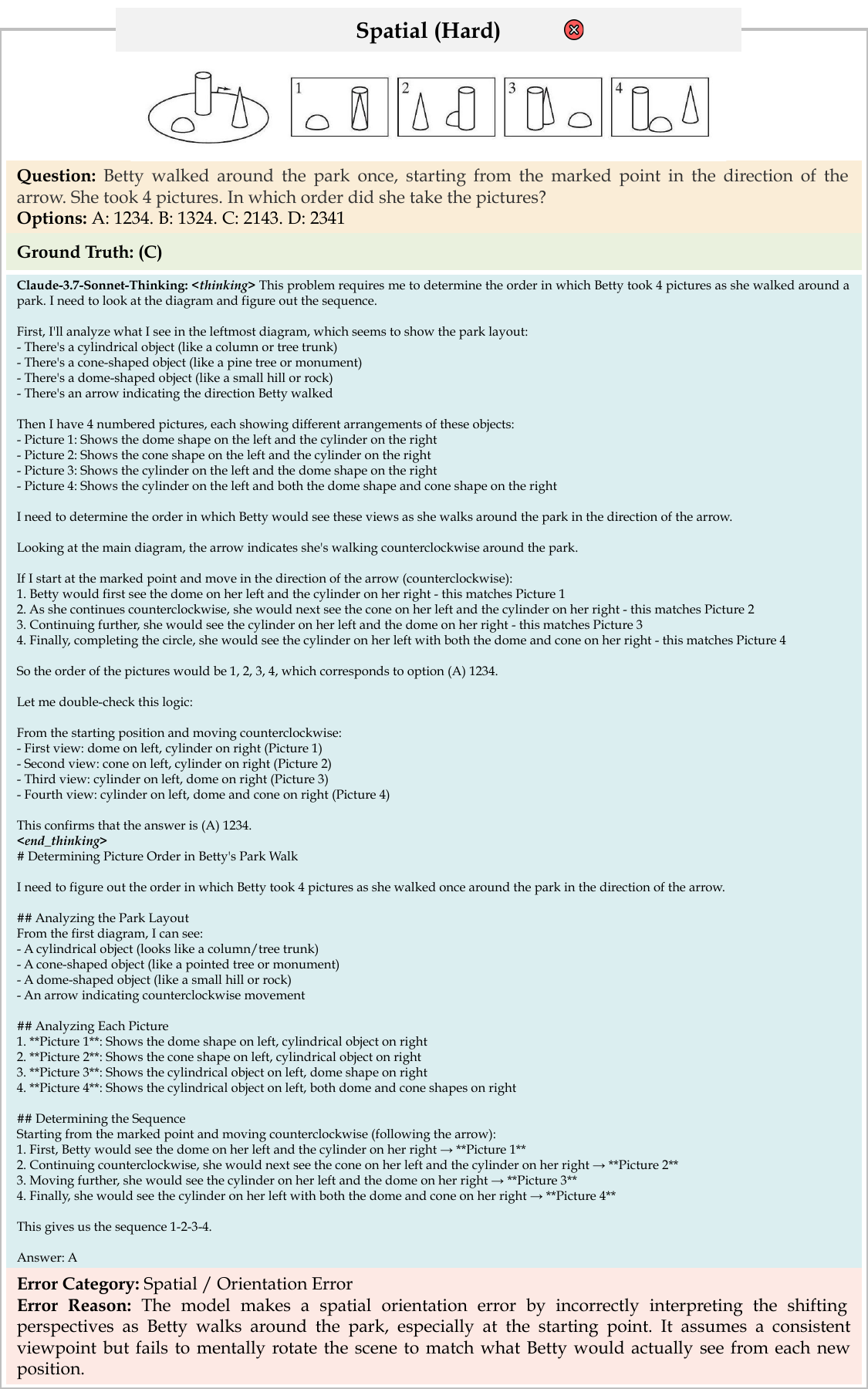}
    \caption{A sample error case of Spatial Reasoning (difficulty: Hard).} 
    \label{fig:spatial_hard_incorrect}
\end{figure*}

\begin{figure*}[!h]
    \centering
    \includegraphics[width=\linewidth]{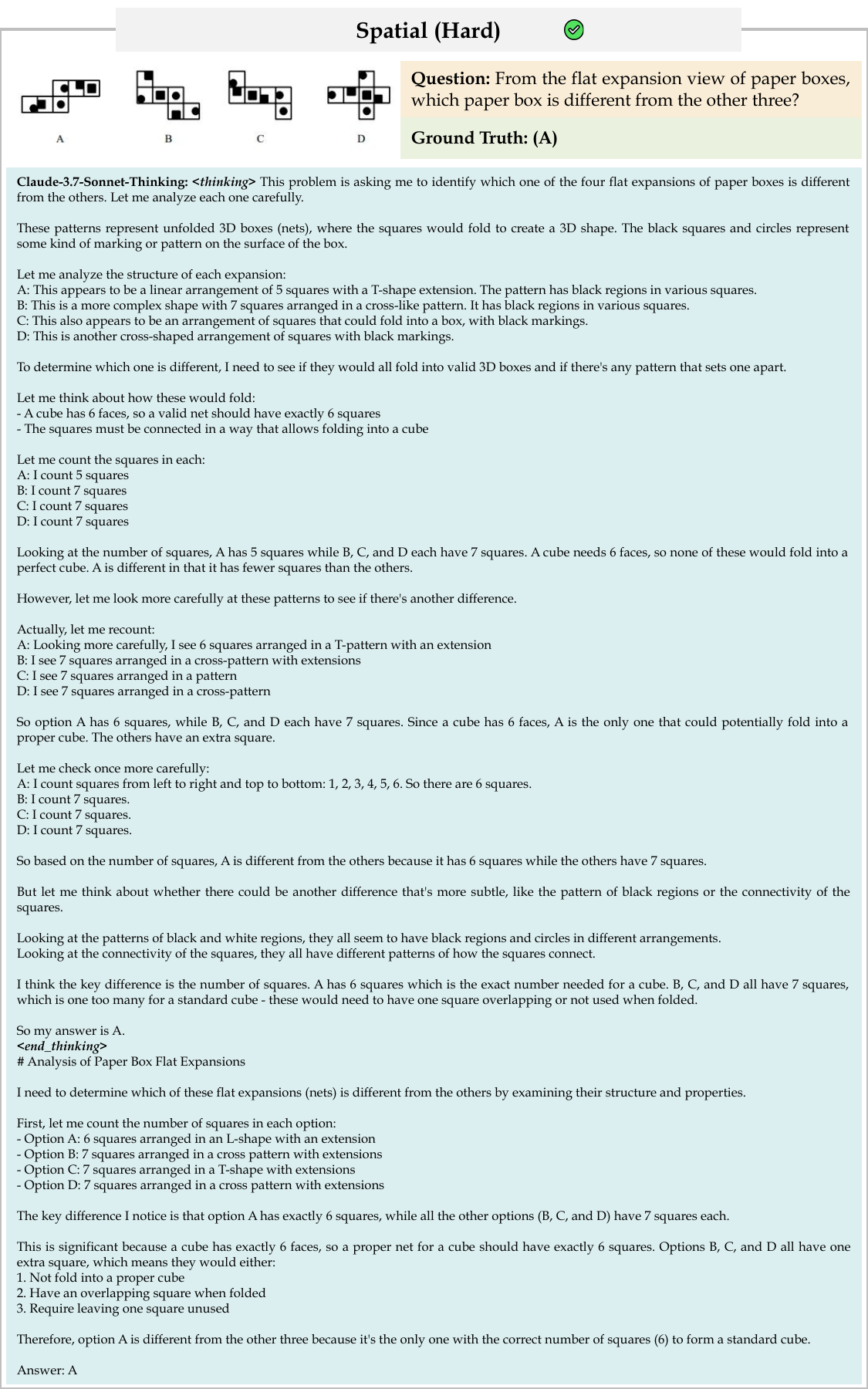}
    \caption{A sample correct case of Spatial Reasoning (difficulty: Hard).} 
    \label{fig:spatial_hard_correct}
\end{figure*}